\documentclass[letterpaper]{article} % DO NOT CHANGE THIS
\usepackage[preprint]{aaai2027}  % DO NOT CHANGE THIS
\usepackage[hyphens]{url}  % DO NOT CHANGE THIS
\usepackage{graphicx} % DO NOT CHANGE THIS
\usepackage{natbib}  % DO NOT CHANGE THIS AND DO NOT ADD ANY OPTIONS TO IT
\usepackage{caption} % DO NOT CHANGE THIS AND DO NOT ADD ANY OPTIONS TO IT
\usepackage{algorithm}
\usepackage{algorithmic}
\usepackage{amsmath, amsfonts, amssymb}
\usepackage{comment}
\usepackage{cuted}
\usepackage{url}
\usepackage{newfloat}
\usepackage{listings}
\DeclareCaptionStyle{ruled}{labelfont=normalfont,labelsep=colon,strut=off} % DO NOT CHANGE THIS
\floatstyle{ruled}
\newfloat{listing}{tb}{lst}{}
\floatname{listing}{Listing}

\usepackage{booktabs}
\usepackage{subcaption}
\title{Accelerating Wildfire Hazard Mapping Across Canada with an AI Surrogate Modelling Framework}
\author {
    Hager Radi Abdelwahed\textsuperscript{\rm 1},
    Toluwaleke Olutayo \textsuperscript{\rm 1},
    Chirag Goel\textsuperscript{\rm 1}\equalcontrib,
    Jana Osea\textsuperscript{\rm 1} \equalcontrib,
    Aldo Zaimi\textsuperscript{\rm 1} \equalcontrib,
    Francisco Emiliano Lopez \textsuperscript{\rm 1},
    Gaétan Marceau Caron \textsuperscript{\rm 1} \\
    Tom Swystun \textsuperscript{\rm 2}, Sandy Erni \textsuperscript{\rm 2}, Yan Boulanger \textsuperscript{\rm 2} 
}
\affiliations {
    \textsuperscript{\rm 1}Mila: Quebec AI Institute\\
    \textsuperscript{\rm 2}Natural Resources Canada - Canadian Forest Services\\
    hager.radi@mila.quebec, toluwaleke.olutayo@mila.quebec, gaetan.marceau.caron@mila.quebec, tom.swystun@nrcan-rncan.gc.ca, yan.boulanger@nrcan-rncan.gc.ca 
}
\usepackage{eso-pic}
\usepackage{xcolor}

\begin{document}

\maketitle

\begin{abstract}
Canada’s increasing wildfire activity requires frequent national hazard mapping to guide proactive mitigation, including fuel management, land-use planning, infrastructure protection, and evacuation routing.
However, process-based wildfire simulation frameworks (e.g., BurnP3 and BurnP3+) rely on computationally intensive Monte Carlo simulations to model burn probability, fire intensity, and rate of spread. These workflows can take \textit{days} to \textit{weeks} to process a single regional modelling unit.
To overcome these computational bottlenecks, we present the following contributions, developed through collaboration between AI researchers and wildfire modelling experts: (a) a standardized framework for training, fine-tuning, and evaluating AI surrogate models against process-based wildfire simulators, (b) a multi-task deep learning surrogate model trained on national-scale data to emulate burn probability, mean fire intensity and mean rate of spread, for wildfire hazard assessment at a fraction of the computational cost (\textit{seconds}), and  (c) a suite of domain-specific metrics and counterfactual evaluation protocols designed to probe out-of-distribution generalization.
By integrating multi-source geospatial data (fuel, terrain, ignition likelihood) and regional weather, our model approximates high-fidelity BurnP3+ simulation outputs, achieving strong performance across agreement, rank-correlation, error, and spatial-overlap metrics. It also preserves downstream hazard classifications and produces physically plausible responses to expert-informed counterfactual data interventions. 
Furthermore, we demonstrate zero-shot transfer to a separate regional BurnP3+ study and show that lightweight fine-tuning can substantially improve regional calibration and predictive agreement. 
The surrogate achieves $\sim806\times$ end-to-end CPU acceleration and uses 13\% of BurnP3+'s host-memory on a representative region, facilitating rapid scenario analysis.

\end{abstract}
 
% Uncomment the following to link to your code, datasets, an extended version or similar.
% You must keep this block between (not within) the abstract and the main body of the paper.
% Make sure that you do not de-anonymize yourself with these links.
\begin{links}
    \scriptsize
    \link{Tool and Code}{https://github.com/milatechtransfer/AISurrogate_wildfireriskmapping}
    % \link{Datasets}{https://aaai.org/example/datasets}
    % \link{Extended version}{https://aaai.org/example/extended-version}
\end{links}

\section{Introduction}
% Hager - in review [Gaetan, Jana]
\begin{figure*}[t]
    \centering
    \includegraphics[width=0.85\textwidth]{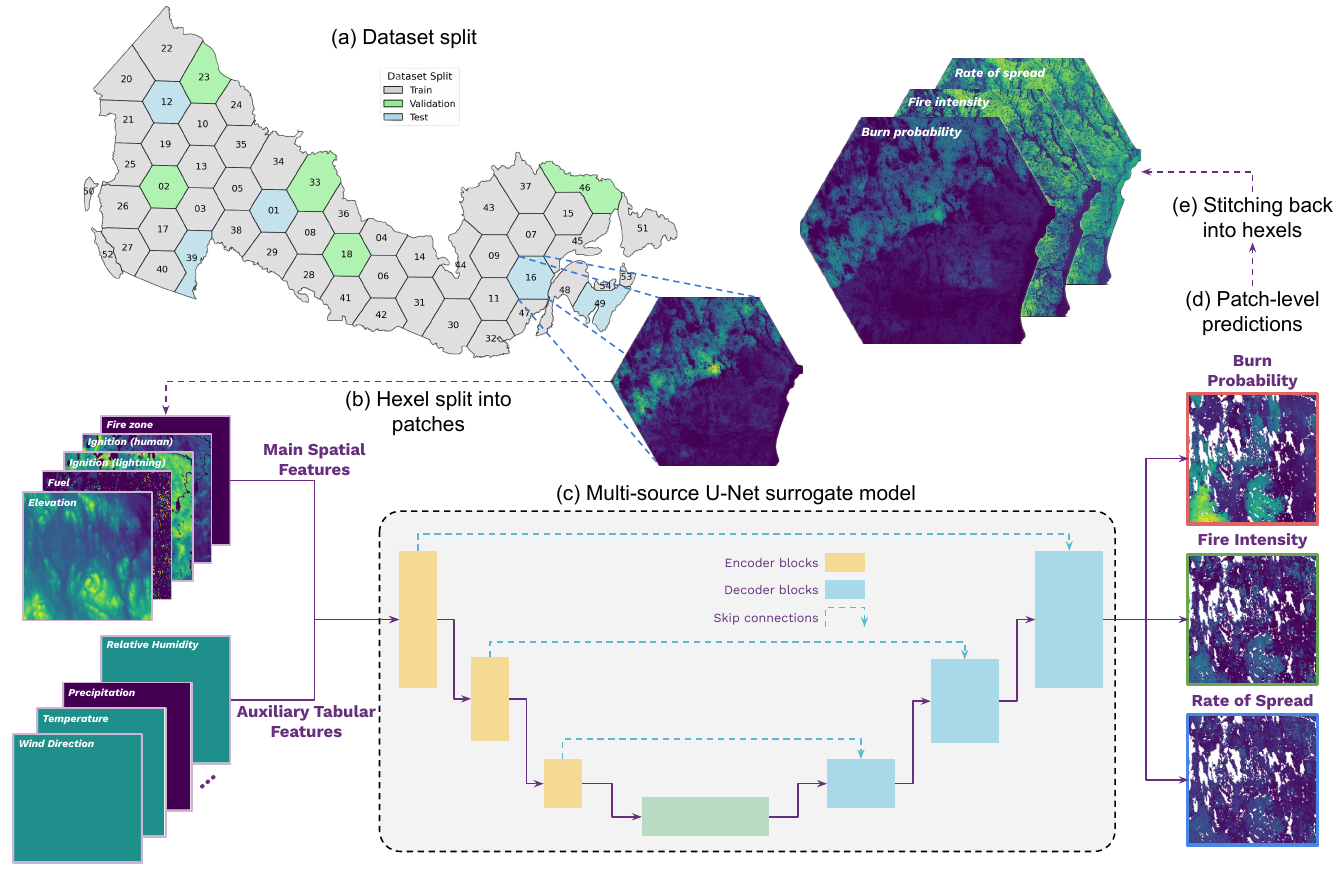}
    \caption{\textbf{Overview of the proposed AI surrogate framework}. (a) Geographic regions (hexels) across Canada are split into training, validation, and test sets (44, 5, and 5 regions, respectively). A standard region (hexel) covers an area of 130,000 $\mathrm{km}^2$. (b) Hexels are divided into square patches for training. (c) A multi-source U-Net model combines spatial and tabular data to predict wildfire hazard in a multi-task fashion. (d) Three spatial outputs are predicted: burn probability, fire intensity, and rate of spread. (e) During inference, predicted patches are reassembled into complete regions using a sliding window approach.}
    \label{fig:model_design}
\end{figure*}

Canada has experienced a marked increase in wildfire activity in recent decades, reflecting multiple interacting drivers, including increasingly fire-conducive conditions intensified by climate change \citep{wri2024canadawildfires, jain2024drivers, kirchmeier2024human}.
Large-scale fire events may cause significant ecological impacts, displace communities, degrade air quality across the continent, and impose billions of dollars in suppression and recovery costs \citep{ibc2024recordlosses}.
Therefore, accurate, high-resolution risk assessments are essential for resource managers planning fuel treatment, utility safety, zoning, critical infrastructure, and emergency evacuation.
Wildfire risk experts in Canada use a sophisticated process-based modelling workflow to characterize and map wildfire hazard across the country. 
This process integrates spatial fire growth models like Prometheus~\citep{prometheus} or FireSTARR \citep{cwfmf2024firestarr}, used to simulate the spread of individual fires, within the BurnP3 ~\citep{burnp3} or BurnP3+ ~\citep{burnp3plus} simulation framework. 
BurnP3/BurnP3+ is then used to produce national or regional probabilistic maps that characterize three key fire-behaviour metrics: (1) burn probability (BP), (2) mean fire intensity (FI), and (3) mean fire spread rate (ROS), which are important metrics for long-term hazard mapping in Canada ~\citep{erni2024mapping}.
Assessments must be regularly updated to reflect shifting climate and fuel conditions, but simulation models take days or weeks even on high-performance computers \citep{parisien2019applications}. This computational burden restricts the frequency of national updates and limits the evaluation of alternative scenarios.
Building on recent advances that have used AI to reduce mechanistic fire model run times by over 99\% \citep{radford2024efficient}, this work presents a multi-source, multi-task U-Net-style \citep{ronneberger2015unet} AI surrogate model for wildfire hazard mapping. 
We train our model directly on outputs (burn probability, mean fire intensity, and mean fire rate of spread) produced by BurnP3+, the latest and most computationally efficient iteration of the simulator \citep{burnp3plus, apexrms2024burnp3plus}. 
To comprehensively characterize wildfire hazard, we evaluate these spatial outputs and benchmark our model's predictions directly against the BurnP3+ reference simulator.
This project represents an interdisciplinary collaboration between wildland fire research scientists at Natural Resources Canada and AI researchers at Mila; consequently, all modelling decisions, feature engineering strategies, and evaluation metrics were developed in consultation with wildfire-domain experts.
Beyond national-scale wildfire mapping, this framework establishes a scalable paradigm for AI-driven emulation, offering a transferable template to bypass computational bottlenecks across environmental and physical sciences. Our contributions are summarized as follows:
\begin{itemize}
\item A framework that allows wildland fire scientists to systematically train, evaluate, fine-tune, and deploy deep learning surrogate models across new geographical regions in Canada. \footnote{While our code and models are publicly accessible, the underlying dataset is subject to controlled access to prevent misinterpretation and improper use by non-experts in high-stakes public safety and risk planning.}
\item A deep learning multi-source AI surrogate model, trained at the national scale, that acts as a high-speed emulator for BurnP3+, accurately approximating burn probability, fire intensity, and rate of spread with strong cross-metric agreement at a fraction of the computational cost (from \textit{days} to \textit{seconds}).
\item A suite of domain-specific metrics alongside counterfactual evaluation protocols designed to probe spatial generalization and responses to out-of-dataset interventions.
% \item Evaluation on an independent BurnP3+ regional study, demonstrating the transferability of the nationally trained AI surrogate and showing that lightweight fine-tuning can effectively adapt to divergent experimental design.
\item Evaluation of the national AI surrogate on an independent BurnP3+ regional study, demonstrating strong cross-study transfer and showing that lightweight fine-tuning effectively corrects out-of-distribution regional miscalibration.
% \item An evaluation framework for assessing process-based wildfire simulators against real-world observations, enabling systematic analysis of simulator fidelity and potential model--reality gaps.

\end{itemize}

\section{Related Work}
% Hager - in review [Gaetan, Jana, Aldo, Chirag]
Wildfire behaviour is driven by weather, topography, and fuel, alongside spatial patterns of natural and anthropogenic ignitions \citep{moritz2005, vanwagner1987development}. While process-based simulators explicitly model fire growth using these variables to achieve high physical accuracy and interpretability, this mechanistic fidelity comes at extreme computational cost. AI surrogate models can resolve this speed–accuracy trade-off, enabling rapid approximation of process-based wildfire simulation outputs. AI surrogates used to generate burn maps have achieved more than a 99\% reduction in computation time relative to JULES-INFERNO, a process-based wildfire simulation model used in the UK \citep{julesinferno}. Similar advances in emulator technology have shown that spatio-temporal neural networks can reproduce complex fire-front simulations at a fraction of the computational cost \citep{australia2022spatio}. In South Australia, dimensionality-reduction techniques combined with an artificial neural network predicted burn-probability maps while requiring only 0.6\% of the original simulation time \citep{radford2024predicting}. Other work has demonstrated that deep-learning-based wildfire forecasts can be produced within seconds, making them $\mathrm{10^2} \text{ to } \mathrm{10^4}$ times faster than direct simulations \citep{realtimeforecast}. This speedup is particularly important for uncertainty estimation: neural emulators can efficiently generate large ensembles and probabilistic fire-arrival maps, providing actionable information within the narrow decision window before a fire spreads \citep{australia2023neural}. While AI surrogates accelerate predictions of fire perimeters and burn probabilities, models trained on narrow geographic regions often suffer from limited generalization and inherited simulation biases. 
By leveraging simulation outputs across Canada’s diverse fuel types, terrain, and fire regimes, our framework is aimed at reducing these regional constraints, positioning AI surrogates to complement traditional simulators for rapid scenario analysis and decision support.

\section{Dataset}
\label{sec:dataset}

% 1970 - 2023 (25 ha) - except for Prince Edward Island (1 ha) - it is a fire weather zone by its own (it should be 2025 but we didn't have the data)

% retaining only simulated fires that reached a minimum size threshold of $25$ ha
% Hager 
% reviewed by Tom, Aldo, Gaetan, Jana, chirag
\subsection{BurnP3+ Simulations}
BurnP3+ is a fire simulation framework combining static and stochastic inputs with fire growth engines such as FireSTARR \citep{cwfmf2024firestarr}, which is based on the Canadian Forest Fire Behaviour Prediction System \citep{fbpSystem1992}.
Each iteration simulates one realization of a fire season by stochastically sampling ignition locations, fire durations, and daily weather conditions to represent spatial and temporal variability. 
Facilitated by our close collaboration with wildfire modelling experts, this work leverages a national-scale BurnP3+ simulation study completed between 2024 and 2026, representing national fire hazard for 2026 onwards. 
Due to their high computational demands, process-based burn probability assessments at this spatial extent remain rare in Canada, the only prior example to our knowledge being \citet{erni2024mapping}. 

\paragraph{Inputs.}The simulation environment utilizes a baseline 2020 Fire Behaviour Prediction (FBP) System fuel type map derived from the Spatialized Canadian Forest Inventory \citep{guindon2024new}, dynamically updated with observed wildfire disturbances spanning 2020 to 2025 \cite{skakun2024mapping, pelletier2024inter}. 
Historical conditioning inputs incorporate meteorological records from 1980 to 2024, a Digital Elevation Model (DEM), and distinct human- and lightning-caused ignition probability surfaces partitioned across distinct fire seasons, derived from the Canadian National Fire Database \cite{cfs2021canadian}. 
Due to the computational demands of processing Canada’s massive landmass simultaneously, the national map is partitioned into 54 regional modelling units, or “hexels”, with a standard size of 130,000 $\mathrm{km}^2$. To achieve simulation stability, between $20,000$ and $150,000$ Monte Carlo iterations were run for each region within the BurnP3+ framework, resulting in approximately $22$ million simulated fires across the study area (Canada-wide). 
We provide the AI surrogate with the same spatial drivers used in the Monte Carlo simulations, including weather, topography, fuel, and ignition variables.

\paragraph{Outputs.} Given the inputs mentioned above, the simulations produce three nationwide rasters at $100$-meter resolution, aggregated across Monte Carlo iterations, serving as simulator-derived training targets: a) BP: the number of times a cell burned divided by the number of iterations, b) FI: fire intensity in $\mathrm{kW \cdot m^{-1}}$, combining head, flank, and back fire intensities and averaged over iterations in which the cell burned, and c) ROS: fire spread rate in $\mathrm{m \cdot min^{-1}}$, combining head, flank, and back rates of spread and averaged over iterations in which the cell burned.

\subsection{Machine Learning Dataset Preparation}
% Jana: In review []
% reviewed by: Hager, Gaetan, Aldo, Chirag
Dataset preparation for the AI surrogate model defines spatial and tabular inputs alongside three spatial reference outputs (burn probability, mean fire intensity, and mean rate of spread). Input sources and processing procedures are detailed below, with further details in Appendix \ref{app:data_prep}.

% Jana: In review []
% reviewed by: Hager, Gaetan, Aldo, chirag
\subsubsection{Spatial Inputs.} 
Spatial features constitute the primary spatial inputs to BurnP3+ and are provided at a $100$-m resolution. Each feature is processed independently as a 2D grid and subsequently stacked to form a multi-channel input tensor for the AI surrogate model.

\begin{itemize}
    \item \textbf{Fuel:} 
    Fuel data is derived from the Spatialized Canadian Forest Inventory following the Fire Behaviour Prediction (FBP) system \citep{guindon2024new, wotton2009updates}, at a spatial resolution of $100$-m, which sets the native resolution of all inputs, targets, and model predictions. While the map originally includes $\approx 30$ distinct fuel classes, we avoid using these categorical classes directly by computing domain-specific features using the Canadian FBP system \citep{fbpSystem1992, cffdrPackage}. Specifically, we calculate 18-dimensional initial Rate of Spread (iROS) curves representing a fire's initial rate of spread under fixed environmental conditions using the package's expert-verified equations. A small neural network compresses this 18-value fuel vector into a 4-dimensional spatial embedding, potentially supporting transfer across unseen but related fuel types.
    \item \textbf{Elevation:} Elevation data is sourced from a Digital Elevation Model (DEM) of Canada \citep{nrcan_mrdem}. The raw elevation values are resampled from their native $30$-m spatial resolution to $100$-m and subsequently min--max normalized. Additionally, we compute slope and aspect directly from the DEM to serve as supplementary elevation features (two extra channels).
    \item \textbf{Ignition:} The raw ignition grids are provided separately by season and cause. For each ignition cause, the seasonal rasters are combined using a weighted sum, with weights given by the corresponding seasonal ignition likelihoods used in the simulation framework. This aggregation produces one ignition frequency map per cause for each region, corresponding to natural lightning and human-caused ignitions.
\end{itemize}

\subsubsection{Tabular Inputs.}
These features consist of non-spatial weather data processed into tabular formats, which are part of the BurnP3+ framework input.
\begin{itemize}
\item \textbf{Weather Data}: Daily weather records are aggregated from Environment and Climate Change Canada~\citep{eccc_canada}, provincial and territorial fire management agencies~\citep{nrcan_cwfis} and supplemented by ERA5-Land reanalysis data~\citep{munoz2021era5land}. These channels include Temperature, Relative Humidity, Precipitation, $\text{wind}_x$, $\text{wind}_y$, Fine Fuel Moisture Code (FFMC), Duff Moisture Code (DMC), Drought Code (DC), Initial Spread Index (ISI), Buildup Index (BUI), and the Fire Weather Index (FWI). Features are normalized with either min-max normalization or z-score normalization depending on their distribution; further details can be found in Appendix~\ref{app:data_prep}.

\item \textbf{Fire Size Distribution:} Historical fire sizes are sourced from the National Fire Database (NFDB) \cite{cfs2021canadian}, with a minimum fire size of 25 ha applied for the national model. The values are then processed using a $\log(x+1)$ transformation followed by min-max normalization.
\end{itemize}
% Daily weather are compiled from ECCC  and CWFIS (2000–2024) \citep{nrcan_cwfis}

% Jana: In review
% reviewed by: Hager, Gaetan, Aldo, Chirag
\subsubsection{Reference Targets.}
The reference target labels for training the AI surrogate model are three spatial output maps produced by BurnP3+. Each target variable undergoes specific preprocessing using training-split statistics to ensure numerical stability during optimization: a) \textbf{BP:} values are min-max normalized to the range $[0, 1]$, b) \textbf{FI} and \textbf{ROS:} values are first transformed using a $\log(1+x)$ operation and are subsequently $z$-score standardized using the training-derived log mean and standard deviation, to stabilize variance and manage severe skewness.
% because their distributions were highly skewed and covered a wide range. The log transform prevents extreme values from dominating the loss and makes the targets easier to model
% Jana: In review []
% reviewed by: Hager, Gaetan, Aldo, Chirag
\subsubsection{Data Sample Generation.}
We partition zero-padded regional grids into $256 \times 256$ patches (with 0.2 overlap) using a sliding window, discarding patches with $<1\%$ valid data. The processed spatial variables within each valid patch are then stacked into a unified multi-channel input tensor. The patching and reconstruction process is illustrated in Figure \ref{fig:model_design}(b) and (e). Finally, the dataset is split into training, validation, and test sets using an approximate 80/10/10 distribution. Designed in collaboration with wildfire research domain experts, this splitting is strictly performed at the region level (see Figure \ref{fig:model_design}(a)) rather than the patch level to prevent spatial data leakage. This results in the following number of samples for each split: \textbf{15,422} training, \textbf{2,146} validation, and \textbf{1,736} test samples after filtering, where each sample (patch) covers an area of $25.6$ km $\times$ $25.6$ km.
% [Before filtering]: Total number of samples in train, val, eval are: 22099 3166 2397

\section{Methodology}
% Hager: in review [GMC, Jana, Aldo, Chirag]
\subsection{Problem Formulation}
Let $\Omega \subset \mathbb{R}^2$ represent the spatial grid domain. The input to our surrogate model is a multi-channel geospatial raster $\mathbf{X} \in \mathbb{R}^{H \times W \times C}$, where $H \times W$ denotes the spatial grid dimensions and $C$ represents the total number of input channels. These channels combine geospatial attributes (fuel models, terrain elevation, slope, aspect, and ignition likelihood maps) alongside tabular features (regionally representative weather data).

We define three distinct hazard outputs, which are commonly used as primary sources for wildfire hazard mapping:
\begin{itemize}
    \item $\mathbf{Y}^{(1)} \in [0, 1]^{H \times W}$: \textbf{burn probability (BP)} map.
    \item $\mathbf{Y}^{(2)} \in \mathbb{R}_{\ge 0}^{H \times W}$: Mean \textbf{fire intensity (FI)} map.
    \item $\mathbf{Y}^{(3)} \in \mathbb{R}_{\ge 0}^{H \times W}$: Mean \textbf{rate of spread (ROS)} map.
\end{itemize}

We train a single, unified, multi-task deep learning surrogate model $f_{\theta}$ (parameterized by weights $\theta$) that acts as a high-speed emulator, framed as a dense, pixel-wise regression mapping that simultaneously predicts all three hazard outputs, such that $f_{\theta}(\mathbf{X}) = \hat{\mathbf{Y}}$. For conciseness, we hereafter refer to \textit{mean fire intensity} and \textit{mean rate of spread} simply as \textit{fire intensity} and \textit{rate of spread}.

% Chirag - in review [Jana, Aldo]
\subsection{Model Design}
Our surrogate model, shown in Figure~\ref{fig:model_design}(c), is based on a U-Net backbone \citep{ronneberger2015unet} adapted to jointly incorporate spatial and tabular inputs for wildfire prediction. While modern backbones such as ConvNeXt ~\citep{liu2022convnet2020s} and Vision Transformers~\citep{liu2021swintransformerhierarchicalvision} offer higher capacity, a standard U-Net provides a lightweight, data-efficient architecture that captures local spatial patterns with minimal computational latency. Crucially, it is designed to capture local landscape characteristics (e.g., fuel, elevation) while conditioning predictions on the broader regional context provided by our tabular weather features. Given a multi-channel spatial input patch together with its associated tabular features, the model predicts three aligned spatial output maps at the same resolution: burn probability, fire intensity, and rate of spread.

\paragraph{Input representation.} The main input to the model is a stacked spatial tensor constructed from the patch-based features described in Section \ref{sec:dataset}. In addition, the tabular weather-related and fire-size distribution features are incorporated directly into the input representation. We aggregate the tabular features at the regional level, and then project the resulting mean to the spatial dimensions of the input patch. For fire-size distribution, we use 3 channels representing the 10th, 50th and 90th percentiles of the fire size distribution. These projected tabular features are stacked with the original spatial channels, yielding a unified multi-channel tensor that is passed directly to our model. We opted for this simple channel stacking approach after empirical tests showed that more complex bottleneck fusion techniques offered minimal performance gains at the cost of significant architectural overhead (detailed in Appendix~\ref{app:additional_experiments}).

\paragraph{Training objective.} Our model is optimized using a joint loss function that aggregates task-specific objectives for each output channel. For the burn probability channel, the loss component is a composite objective designed to capture distributional agreement and spatial ranking. For the fire intensity and rate of spread channels, we apply a weighted regression objective combining error-based and correlation-based terms. Full loss definitions, weighting schemes, and validation-based tuning details are provided in Appendix~\ref{app:experimental_details}.

% Aldo - in review
% Reviewed by: GMC, Hager, Jana, Chirag
\subsection{Evaluation Metrics}
To comprehensively evaluate our model across different dimensions of predictive accuracy, we use a diverse set of metrics mapped into the following three categories:
\begin{itemize}
    \item \textbf{Magnitude-Based Error Metrics:} We use the Normalized Mean Absolute Error (\textit{Normalized MAE}) and Bias (\textit{Normalized Bias}) to quantify the absolute numerical differences and directional offsets between predicted probabilities and reference values.
    % \item \textbf{Distributional Shift Metrics:} We use the Kullback-Leibler (KL) Divergence to evaluate how well the overall predicted probability mass aligns with the true target distribution.
    \item \textbf{Rank and Agreement Metrics:} We calculate the \textit{Spearman} Correlation to assess monotonic relationships independent of magnitude offsets, and the Concordance Correlation Coefficient (\textit{CCC}) to assess absolute agreement, which additionally penalizes systematic bias and scale differences.
    \item \textbf{Spatial Overlap Metrics:} To isolate and evaluate the classification boundaries of predicted versus actual hotspots, we use the Area Under the Curve of top-K Intersection over Union (AUC-IoU). We report the \textit{Top-10\% AUC-IoU} to isolate extreme-risk areas and the \textit{AUC-IoU Full} for overall spatial accuracy.
\end{itemize}
The formal mathematical definitions for all evaluation metrics are provided in Appendix \ref{app:eval_metrics}.

\section{Experiments}
% Hager - In review
% reviewed by: Jana, Aldo, Chirag
\begin{table*}[t]
\centering
\footnotesize
\setlength{\tabcolsep}{4.5pt}
\begin{tabular*}{\linewidth}{
@{\extracolsep{\fill}}
l l r r r r r r
}
\toprule
\textbf{Target}
& \textbf{Model}
& $\downarrow$ \textbf{\shortstack{Normalized \\ MAE}}
& $\rightarrow 0$ \textbf{\shortstack{Normalized \\ Bias}}
& $\uparrow$ \textbf{Spearman}
& $\uparrow$ \textbf{CCC}
& $\uparrow$ \textbf{\shortstack{AUC IoU \\ Top-10\%}}
& $\uparrow$ \textbf{\shortstack{AUC IoU \\ Full}} \\
\midrule

\textbf{BP}
& Mean Baseline
& $5.18$ & $4.53$ & $-$ & $-$ & $0.10$ & $0.55$ \\

& XGBoost
& $1.19 \pm 0.02$
& $0.47 \pm 0.06$
& $0.70 \pm 0.01$
& $0.33 \pm 0.01$
& $0.05 \pm 0.01$
& $0.59 \pm 0.01$ \\

& U-Net Spatial
& $0.50 \pm 0.02$
& $-0.18 \pm 0.06$
& $0.88 \pm 0.00$
& $0.67 \pm 0.03$
& $0.24 \pm 0.01$
& $0.69 \pm 0.00$ \\

& + Weather
& $\mathbf{0.35 \pm 0.00}$
& $\mathbf{-0.14 \pm 0.03}$
& $\mathbf{0.91 \pm 0.00}$
& $\mathbf{0.85 \pm 0.00}$
& $0.39 \pm 0.04$
& $\mathbf{0.75 \pm 0.00}$ \\

& + Weather + Fire Size 
& $0.35 \pm 0.02$
& $\mathbf{-0.14 \pm 0.03}$
& $\mathbf{0.91 \pm 0.00}$
& $0.85 \pm 0.01$
& $\mathbf{0.40 \pm 0.04}$
& $\mathbf{0.75 \pm 0.00}$ \\

%old with fire size mean
%& + Weather + Fire Size
%& $\mathbf{0.34 \pm 0.01}$
%& $-0.15 \pm 0.01$
%& $0.91 \pm 0.00$
%& $\mathbf{0.85 \pm 0.01}$
%& $\mathbf{0.40 \pm 0.04}$
%& $0.75 \pm 0.00$ \\

\cmidrule{2-8}

\textbf{FI}
& Mean Baseline
& $0.64$ & $-0.32$ & $-$ & $-$ & $0.10$ & $0.50$ \\

& XGBoost
& $0.37 \pm 0.00$
& $-0.11 \pm 0.00$
& $0.73 \pm 0.00$
& $0.61 \pm 0.00$
& $0.14 \pm 0.00$
& $0.61 \pm 0.00$ \\

& U-Net Spatial
& $0.30 \pm 0.00$
& $-0.06 \pm 0.02$
& $0.85 \pm 0.00$
& $0.76 \pm 0.01$
& $0.23 \pm 0.00$
& $0.69 \pm 0.00$ \\

& + Weather
& $\mathbf{0.27 \pm 0.00}$
& $-0.04 \pm 0.01$
& $\mathbf{0.86 \pm 0.00}$
& $0.79 \pm 0.01$
& $0.24 \pm 0.01$
& $\mathbf{0.70 \pm 0.00}$ \\

& + Weather + Fire Size
& $\mathbf{0.27 \pm 0.00}$
& $\mathbf{-0.03 \pm 0.01}$
& $\mathbf{0.86 \pm 0.00}$
& $\mathbf{0.79 \pm 0.00}$
& $\mathbf{0.25 \pm 0.01}$
& $\mathbf{0.70 \pm 0.00}$ \\

%old with firesize mean
%& + Weather + Fire Size
%& $0.27 \pm 0.00$
%& $-0.06 \pm 0.02$
%& $\mathbf{0.86 \pm 0.00}$
%& $0.78 \pm 0.01$
%& $0.24 \pm 0.01$
%& $0.70 \pm 0.00$ \\

\cmidrule{2-8}

\textbf{ROS}
& Mean Baseline
& $0.44$ & $-0.16$ & $-$ & $-$ & $0.10$ & $0.50$ \\

& XGBoost
& $0.30 \pm 0.00$
& $-0.09 \pm 0.00$
& $0.62 \pm 0.00$
& $0.51 \pm 0.00$
& $0.11 \pm 0.00$
& $0.56 \pm 0.00$ \\

& U-Net Spatial
& $0.24 \pm 0.00$
& $-0.06 \pm 0.01$
& $0.77 \pm 0.00$
& $0.69 \pm 0.01$
& $0.21 \pm 0.00$
& $0.64 \pm 0.00$ \\

& + Weather
& $\mathbf{0.22 \pm 0.00}$
& $\mathbf{-0.04 \pm 0.00}$
& $0.78 \pm 0.01$
& $0.72 \pm 0.01$
& $0.22 \pm 0.01$
& $\mathbf{0.65 \pm 0.00}$ \\

& + Weather + Fire Size
& $\mathbf{0.22 \pm 0.00}$
& $-0.04 \pm 0.01$
& $\mathbf{0.78 \pm 0.00}$
& $\mathbf{0.72 \pm 0.00}$
& $\mathbf{0.23 \pm 0.00}$
& $\mathbf{0.65 \pm 0.00}$ \\

%old using fire size mean
%& + Weather + Fire Size
%& $0.22 \pm 0.01$
%& $-0.06 \pm 0.01$
%& $\mathbf{0.78 \pm 0.01}$
%& $0.71 \pm 0.01$
%& $0.22 \pm 0.01$
%& $\mathbf{0.65 \pm 0.00}$ \\

\bottomrule
\end{tabular*}
\caption{\textbf{Main results of AI surrogate model} comparing the predictive performance of the baselines and incremental multi-source U-Net models for \textbf{Burn Probability (BP)}, \textbf{Fire Intensity (FI)}, and \textbf{Rate of Spread (ROS)}. The baseline spatial architecture is further augmented with weather features and fire size features. 
Each value is reported as the mean and standard deviation of three runs with different random seeds (standard deviations less than $0.01$ are reported as $0.00$). Metrics are reported on the test set \textit{\textbf{at region level}}; patch-level results are reported in Table~\ref{tab:main_results_patch_level} of Appendix~\ref{appendix:additional_results}. Best performance for each specified metric is highlighted in bold. Note: Correlation metrics are undefined for the mean baseline.}
\label{tab:main_results_region_level}
\end{table*}

\begin{figure}[t]
    \centering
    \begin{subfigure}[b]{\linewidth}
        \centering
        \includegraphics[width=\linewidth]{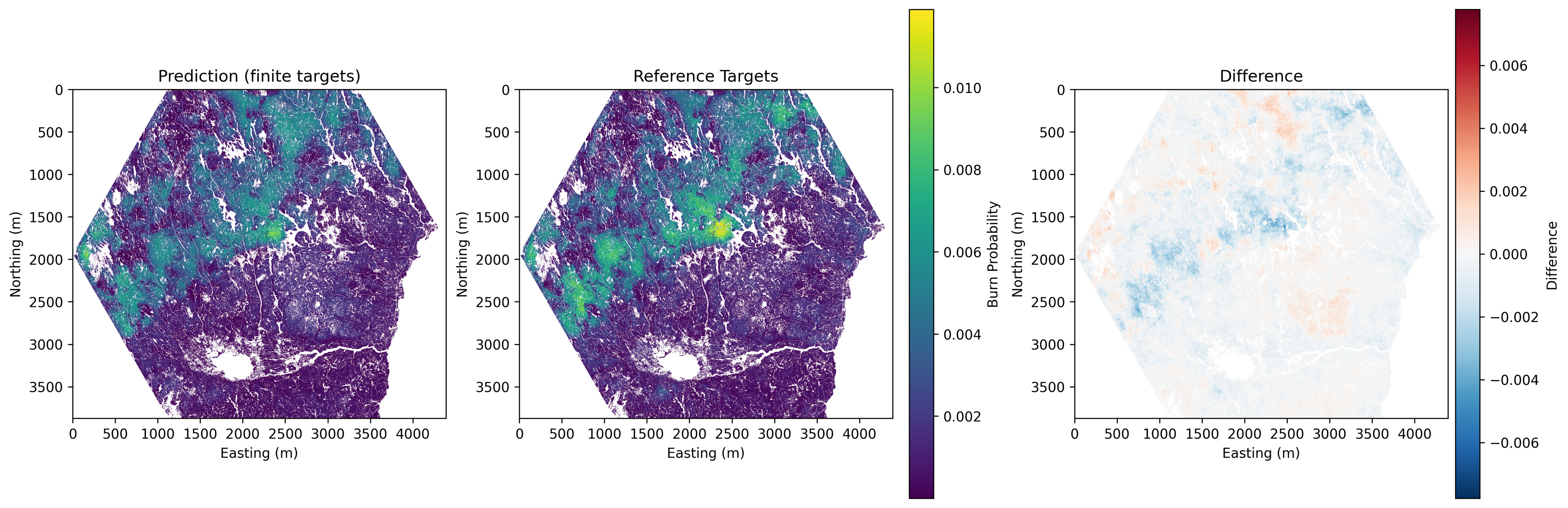}
        \caption{Burn Probability (BP)}
        \label{fig:bp}
    \end{subfigure}
    \vspace{0.5em}
    \begin{subfigure}[b]{\linewidth}
        \centering
        \includegraphics[width=\linewidth]{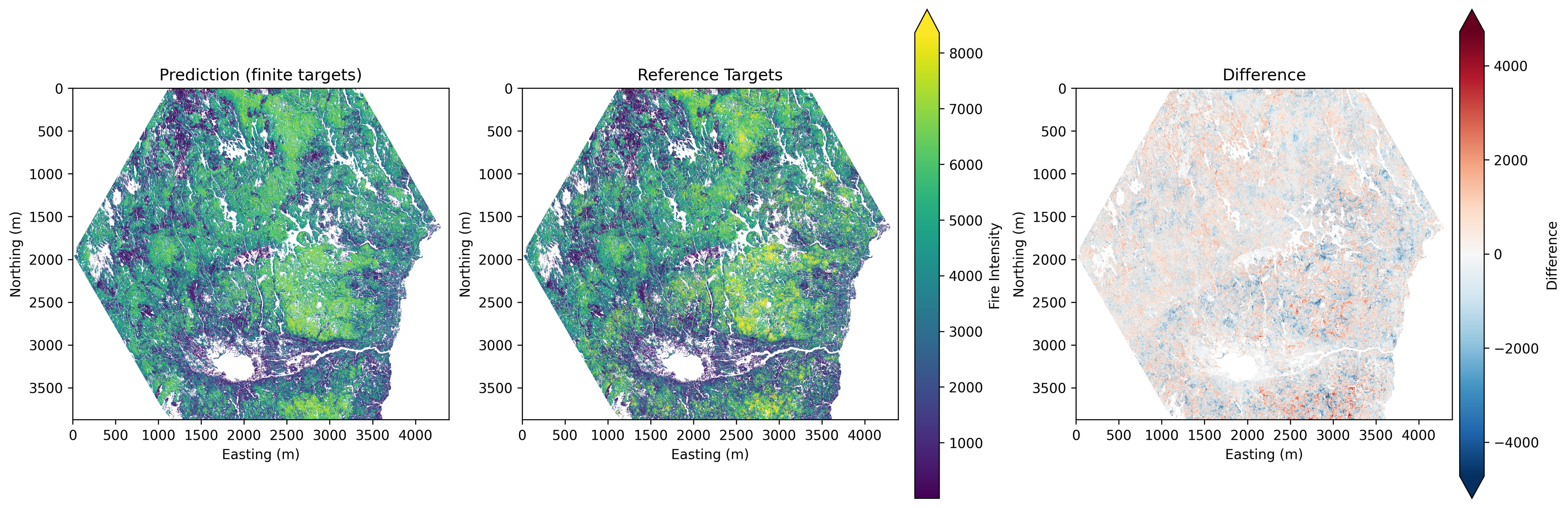}
        \caption{Fire Intensity (FI)}
        \label{fig:fi}
    \end{subfigure}
    \vspace{0.5em}
    \begin{subfigure}[b]{\linewidth}
        \centering
        \includegraphics[width=\linewidth]{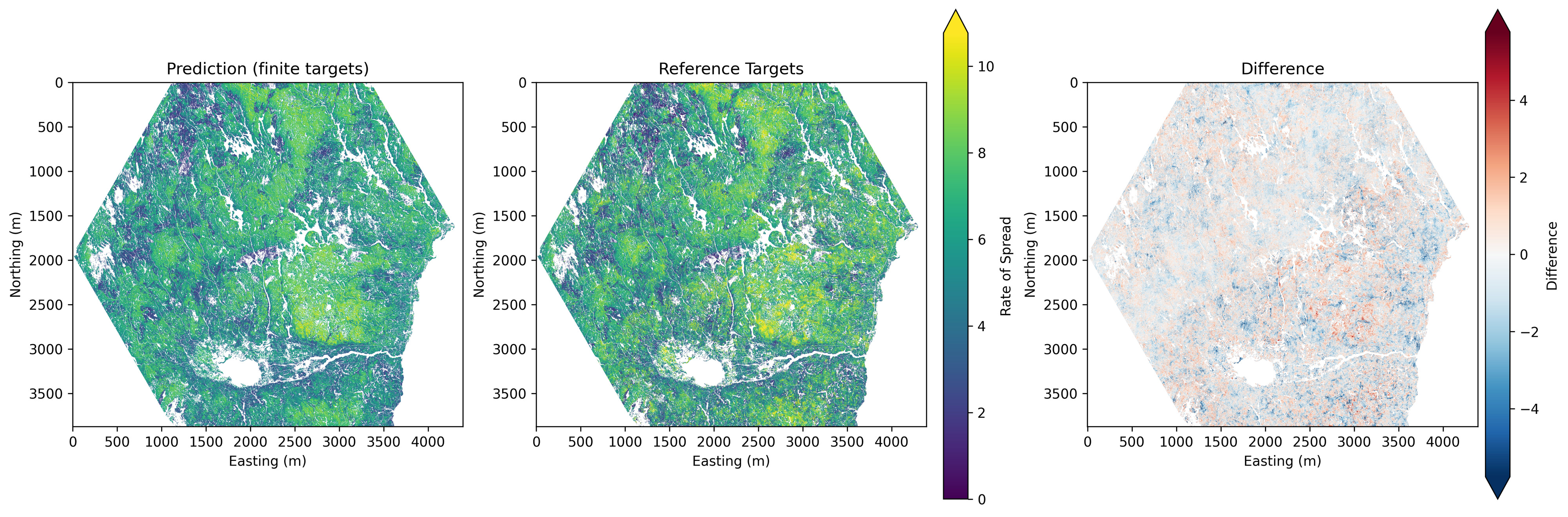}
        \caption{Fire Rate of Spread (ROS)}
        \label{fig:ros}
    \end{subfigure}
    \caption{Comparison of the AI surrogate model predictions against BurnP3+ reference outputs for a representative test region (region 16) across the three targets: BP, FI, and ROS. Masked-out regions represent water bodies or non-fuel areas.}
    \label{fig:main_results}
\end{figure}

We train our multi-task U-Net model under configurations incorporating spatial features, both with and without auxiliary weather data and fire size percentiles. 
For baselines, we report results for a constant mean predictor, which assigns every pixel the mean burn probability, fire intensity, or rate of spread computed from the corresponding training-set target. 
For a stronger learned baseline, we additionally employ XGBoost~\citep{chen2016xgboost} as a traditional machine learning baseline, trained using spatial features only. All U-Net models are trained using AdamW \citep{loshchilov2018_adamW} for $25$ epochs, after which validation loss had plateaued across all configurations. 
Final model evaluation relies on the checkpoint that achieves the highest CCC on the validation set across BP, FI, and ROS. Hyperparameter selection and architectural design are similarly guided by top-performing models on the validation set. 
Full experimental details and hyperparameter configurations are provided in Appendix~\ref{app:experimental_details}.
Table~\ref{tab:main_results_region_level} presents the test set evaluation across region-level metrics, reporting the mean and standard deviation over three random seeds. Patch-level results are reported in Table~\ref{tab:main_results_patch_level} of Appendix~\ref{appendix:additional_results}, as they yield the same model ranking as the region-level evaluation; we additionally report per-region metrics and further analysis on results in Appendix~\ref{appendix:additional_results}. Across all three prediction targets, our multi-task U-Net outperforms all evaluated baselines. Incorporating weather substantially improves BP and produces consistent, more modest gains for FI and ROS. 
Incorporating fire size percentiles yields slight improvements across certain metrics at the regional level. 
While these gains are modest, we retained this model because fire size is critical for BurnP3+ calibration and serves as a valuable input feature for the AI model. 
Consequently, for the remaining analyses, we use the full multi-task model integrating all spatial, weather and fire-size features as our best model. Figure~\ref{fig:main_results} compares the predictions of the best model against BurnP3+ reference outputs for BP, FI, and ROS. Appendix~\ref{app:additional_experiments} provides single-task experiments, alternative architecture evaluations, and ablations on our proposed modifications and fuel representation encodings.

% Hager - In review
% reviewed by: Jana, Aldo
\subsection{Time and Memory Performance Comparison}
To quantify computational efficiency, we perform runtime and memory profiling of the proposed AI surrogate and BurnP3+ \footnote{We use version 2.6.11 of BurnP3+, along with version 1.5.7 of FireSTARR.} under representative hardware configurations (Table~\ref{tab:run_time_performance}). 
The baseline BurnP3+ simulation for region 16 required $38.9$ hours of execution time and a peak memory footprint of $100.8$ GB across $32$ CPU cores to complete $30,000$ iterations, in line with typical operational settings for this region. 
Conversely, the AI surrogate's standalone inference and end-to-end pipeline execution times were benchmarked for the same region over $10$ runs to report mean and standard deviation statistics. The deep learning surrogate yielded substantial computational speedups over the traditional simulator: achieving acceleration factors of $\sim \mathbf{3,432}\times$ on Nvidia L40S GPU, $\sim \mathbf{1,491}\times$ on MacBook MPS, and $\sim \mathbf{806}\times$ on a standard CPU, while requiring only 13\% of BurnP3+'s host memory footprint on a 4-core CPU allocation. 
Accelerated configurations trade memory for speed, reaching 37\% (L40S: 8~GB host plus 29~GB device) and 30\% (MPS: 30~GB unified) of the same baseline.
Although BurnP3+ can exploit multiprocessing to execute simulations in parallel, its computational cost remains tied to the number of simulated iterations and scenarios. Once trained, the surrogate replaces these iterative simulations with batched inference, enabling rapid evaluation of multiple regional scenarios.
%Unlike BurnP3+, which scales linearly with the number of iterations and simulated regions, the AI surrogate utilizes batching to parallelize across multiple regions simultaneously, offering a highly scalable framework suitable for rapid regional scenario analysis and iterative hazard assessment.

\begin{table}[t]
\centering
\footnotesize
\begin{tabular}{@{} l c c r @{}}
\toprule
\multicolumn{4}{@{} p{\columnwidth} @{}}{%
  \textbf{BP3+ Baseline:} 38.9 h (140,040 s) \textbar{} Memory: 100.8 GB \textbar{} 32 CPUs%
} \\ \midrule
\textbf{Hardware Setup} & \textbf{Inference (s)} & \textbf{Total E2E (s)} & \textbf{RAM + Device}\\ \midrule
\textbf{GPU} (L40S) & $7.1 \pm 0.2$ & $40.8 \pm 0.2$ & 8 + 29 GB (37\%) \\
\textbf{MacBook} (MPS) & $78.9 \pm 2.4$ & $93.9 \pm 2.5$ & 30 GB (30\%) \\
\textbf{CPU} (4 Cores) & $138.6 \pm 0.7$ & $173.6 \pm 2.2$ & 13 GB (13\%)\\ \midrule
\textbf{Hardware Setup} & \multicolumn{3}{c}{\textbf{Speed-up Factor vs. BurnP3+}} \\ \midrule
\textbf{GPU} (L40S) & \multicolumn{3}{c}{$\sim\mathbf{3,432.35 \times}$} \\
\textbf{MacBook} (MPS) & \multicolumn{3}{c}{$\sim\mathbf{1,491.21 \times}$} \\
\textbf{CPU} (4 Cores) & \multicolumn{3}{c}{$\sim\mathbf{806.50 \times}$} \\ \bottomrule
\end{tabular}
\caption{Performance and memory comparison between the simulator baseline (BurnP3+) and the AI surrogate model across hardware variants. Values are shown for region 16, one of 54 distinct geographical zones used to partition the Canadian map. Percentages in parentheses indicate memory usage relative to the baseline's 100.8 GB.}
\label{tab:run_time_performance}
\end{table}

% Tolu: in review
% Reviewed by: Hager, Jana, Aldo, Chirag
\subsection{Hazard Mapping Evaluation}
Beyond individual metrics, we evaluate the surrogate model's ability to reproduce a derived wildfire hazard product used for downstream decision support. Following established risk frameworks that combine spatial fire likelihood with potential behaviour to characterize physical hazard \citep{finney2005challenge, scott2013wildfire} and Canadian BurnP3+ mapping protocols \citep{erni2024mapping}, we compute physical hazard as the joint product of BP and FI.

For each test region, using both BurnP3+ and surrogate outputs, hazard is computed pixelwise by: 1) capping average FI to mitigate extreme outliers; 2) multiplying BP by the capped FI to obtain raw hazard; 3) linearly rescaling the result to a $[0, 100]$ range; and 4) discretizing the continuous values into 13 ordinal hazard classes (from $<0.01$ to $\geq 50$) (More details in Appendix~\ref{app:hazard_mapping}).

Table~\ref{tab:hazard_metrics} summarizes the surrogate’s classification performance against BurnP3+. 
By accepting near-miss predictions, agreement increases from $\sim48\%$ (exact 13-class) to $\sim 87\%$ within 1 bin and $\sim 97\%$ within 2 bins. Crucially, the surrogate preserves decision-relevant spatial patterns in high-hazard areas, achieving an IoU of $\sim 85\%$ for classes 10--13.
While residual errors in extreme classes remain an operational priority, overall downstream hazard patterns are well maintained (see Appendix~\ref{app:hazard_mapping} for full confusion matrix).

\begin{table}[t] 
\centering 
% multi-output - weather
\begin{tabular}{lc} \toprule \textbf{Metric} & \textbf{Value} 
\\ \midrule Exact 13-Class Accuracy & $0.476 \pm 0.008$ \\
Within-1 Bin Accuracy & $0.873 \pm 0.004$\\
Within-2 Bin Accuracy & $0.970 \pm 0.001$\\
\midrule High Hazard ($\geq 10$) IoU & $0.850 \pm 0.002$ 
%\\
%Extreme Hazard ($=13$) IoU & $0.000\pm 0.025$
\\ 
\bottomrule 
% multi-output - weather + fire size
% \begin{tabular}{lc} \toprule \textbf{Metric} & \textbf{Value} \\ \midrule Exact 13-Class Accuracy & 0.4483 \\ Within-1 Bin Accuracy & 0.8306 \\ Within-2 Bin Accuracy & 0.9451\\ \midrule High Hazard ($\geq 10$) IoU & 0.7823 \\ Extreme Hazard ($=13$) IoU & 0.6512 \\ \bottomrule 
% \begin{tabular}{lc} \toprule \textbf{Metric} & \textbf{Value} \\ \midrule Exact 13-Class Accuracy & 0.4660 \\ Within-1 Bin Accuracy & 0.8495 \\ Within-2 Bin Accuracy & 0.9517 \\ \midrule High Hazard ($\geq 10$) IoU & 0.8244 \\ Maximal Hazard ($=13$) IoU & 0.7038 \\ \bottomrule 
\end{tabular} 
\caption{AI surrogate model performance on downstream hazard mapping metrics across test regions.} 
\label{tab:hazard_metrics} 
\end{table}

\section{Counterfactual Scenario Analysis and Validation}
\begin{figure*}[t]
    \centering
    \includegraphics[width=0.75\textwidth]{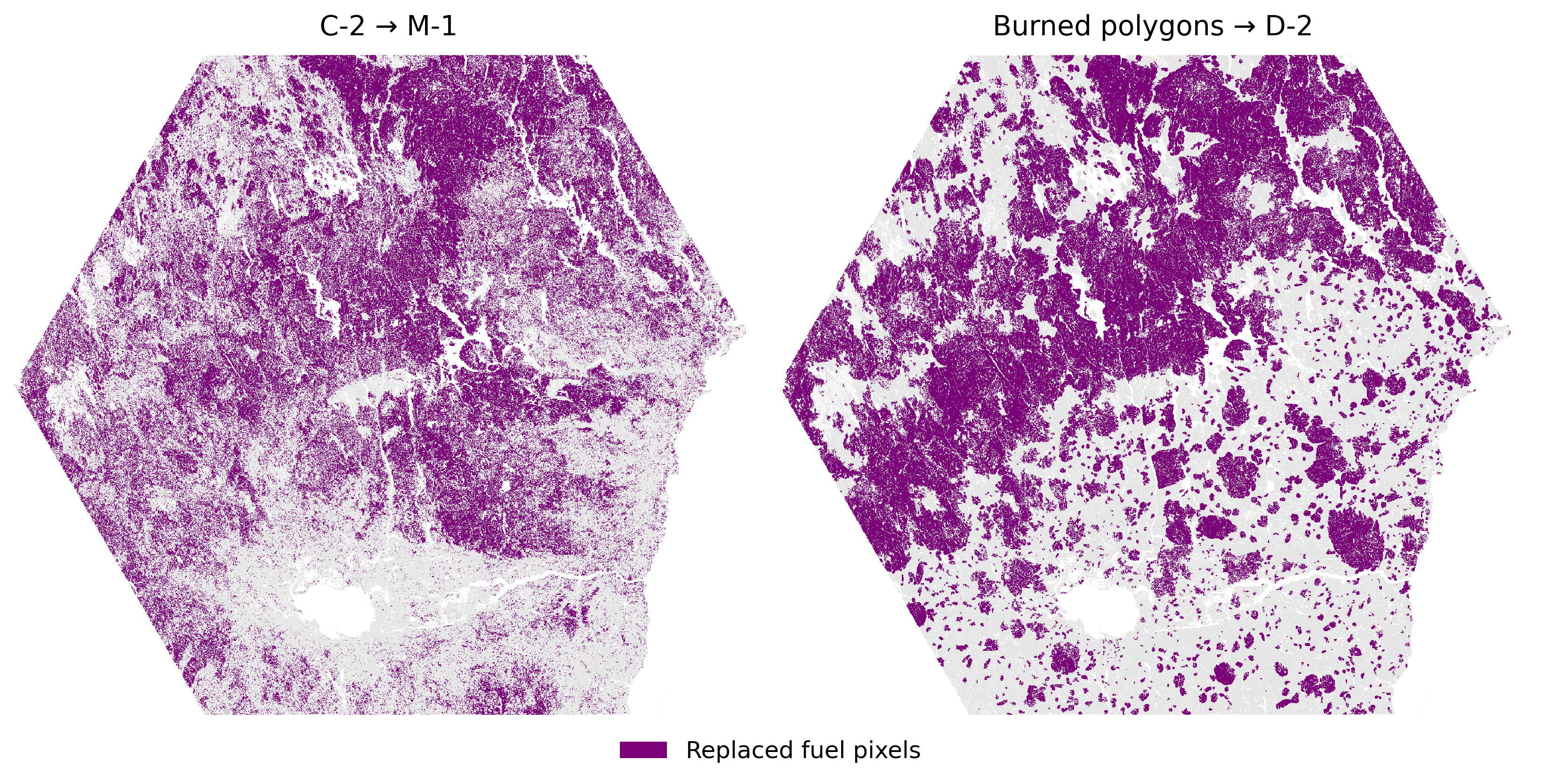}
    \caption{Fuel interventions in region 16: the left panel shows the first intervention: C2 pixels replaced with M1, while the right panel shows the second intervention: pixels within the selected BurnP3+ fire polygons reassigned to D2.}
    \label{fig:fuel_interventions}
\end{figure*}

A primary application of spatial wildfire simulators is \emph{scenario analysis}, which compares fire outcomes under varying landscape, fuel-treatment, weather, and climate conditions \citep{finney2011simulation, thompson2023avoided, zong2024role}. 
Such analyses inform critical decisions, including resource allocation across mitigation strategies like fuel management. 
Formulating these strategies as optimization problems typically demands extensive simulation; a computational burden our surrogate model substantially mitigates. 
Candidate solutions can then be validated against high-fidelity simulators such as BurnP3+. 
To validate the potential of this approach, we evaluate our model’s ability to generalize out-of-distribution by actively intervening on the data distribution.

We conduct this analysis on region 16 (central Québec), as this landscape is well characterized by the wildfire domain experts involved in the study. The region is typical of the transition between mixedwood and boreal forests, where the summer climate is humid and mild, and exhibits marked spatial variation in fire activity, with the most significant activity occurring in its northern half. This strong spatial contrast provides a useful setting for assessing whether the AI surrogate can reproduce variable fire patterns across a heterogeneous landscape.

We consider four counterfactual scenarios: two fuel interventions, one weather intervention, and one climate-change intervention.
For each intervention, we modify the relevant inputs for region 16 and compare the resulting surrogate predictions with the baseline predictions obtained from the original inputs.
Given the computational cost of BurnP3+ and the time available for validation, BurnP3+ validation was limited to the two fuel interventions, allowing us to compare the responses of the surrogate and the simulator\footnote{The training targets supplied by NRCan were generated with an earlier version of BurnP3+. The baseline and fuel-intervention reference simulations shown in this section were run with an updated version in September 2026. Differences between versions may contribute to discrepancies between the surrogate and simulator responses.}.
For these comparisons, each model's response is calculated as its counterfactual output minus its corresponding baseline output.
The weather and climate-change interventions are assessed through surrogate sensitivity analysis.
We examine the resulting changes in BP, FI, and ROS at both aggregate and spatial scales.

\subsection{Fuel Interventions}
\label{fuel_intervention}

We consider two idealized interventions to the landscape fuel layer to evaluate the surrogate's response to substantial changes in fuel composition. In the first scenario, pixels classified as C2 (highly flammable boreal spruce) are replaced with M1 (less flammable boreal mixedwood), based on the FBP system. 
This intervention represents a hypothetical large-scale fuel substitution, for which we expect BP, FI, and ROS to decrease.

\begin{figure*}[p]
    \centering
    \includegraphics[width=\textwidth]{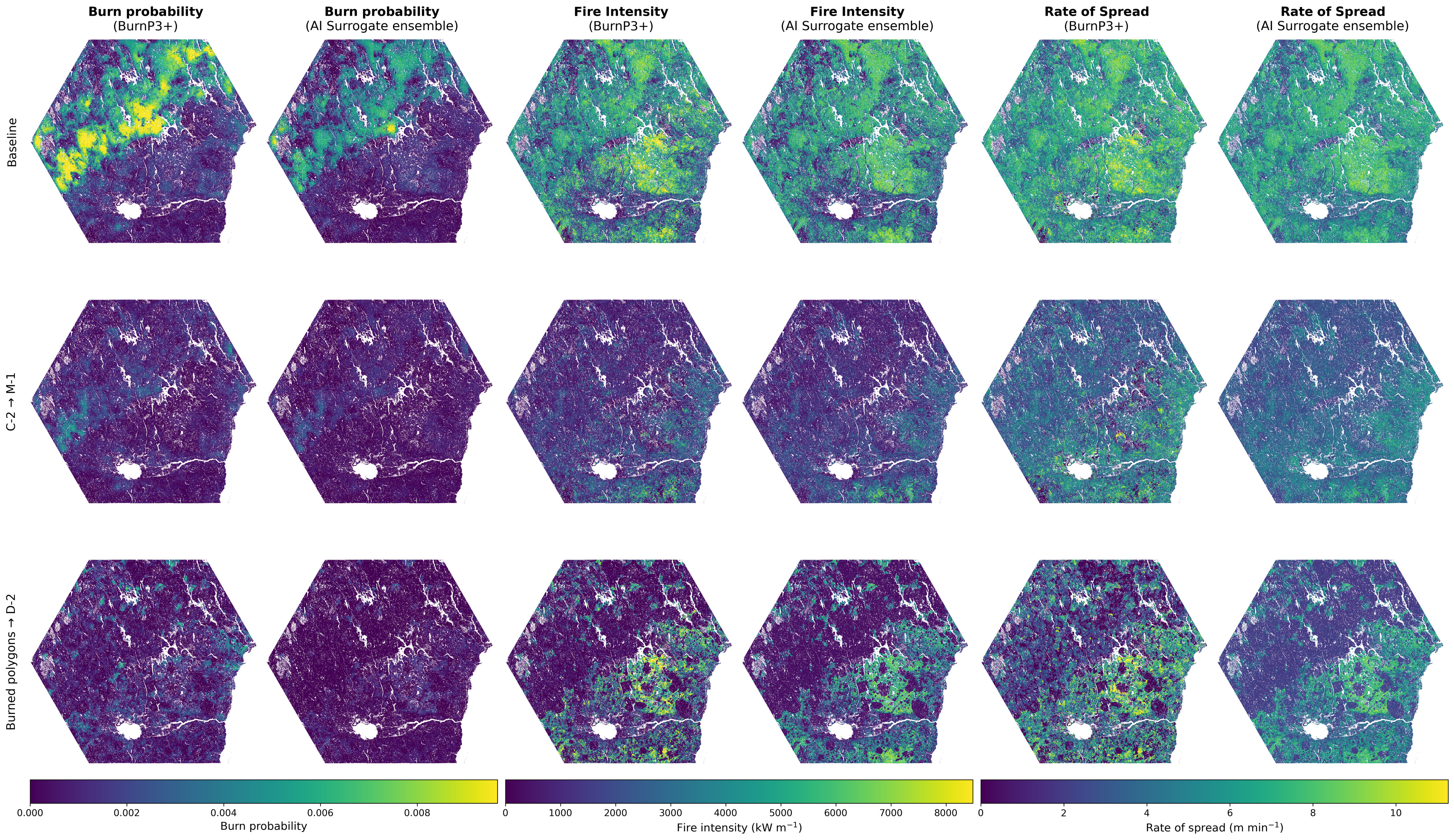}
    \caption{BurnP3+ and AI surrogate model responses for the two fuel interventions in region 16. Rows show the shared baseline (top), the C2-to-M1 intervention (middle), and the post-fire D2 intervention (bottom). Columns are grouped by output: BP, FI, and ROS. Within each pair, BurnP3+ predictions appear on the left and surrogate predictions on the right.}
    \label{fig:fuel_baseline_counterfactual}
\end{figure*}

\begin{figure*}[p]
    \centering
    \includegraphics[width=0.85\textwidth]{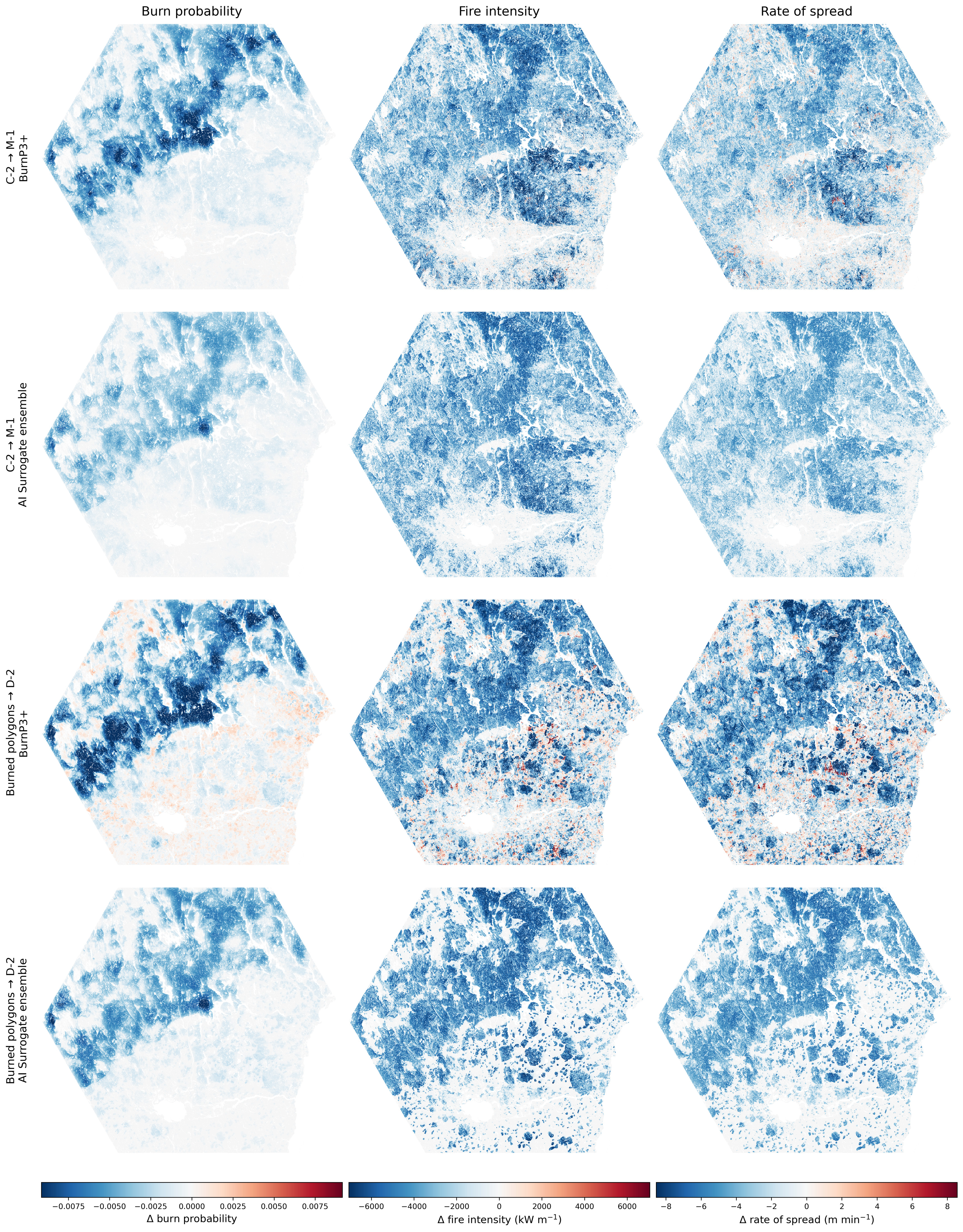}
    \caption{Counterfactual-minus-baseline changes produced by BurnP3+ and the surrogate for the two fuel interventions in region 16. Columns show changes in BP, FI, and ROS. The first two rows show the BurnP3+ and surrogate responses to the C2-to-M1 intervention, respectively. The last two rows show the corresponding responses to the post-fire D2 intervention.}
    \label{fig:fuel_bp3_validation}
\end{figure*}

In the second scenario, fire polygons generated by BurnP3+ are overlaid on the fuel layer for region 16, and the fuels within these polygons are replaced with D2 (Green Aspen), a deciduous FBP fuel type used here as a low-flammability proxy for previously burned areas. 
These areas remain burnable but support weaker fire behaviour. 
The resulting intervention represents an extensive post-fire modification of the fuel landscape. 
A more operationally representative post-fire scenario, based on a smaller number of simulated fire realizations, is evaluated separately in Appendix~\ref{app:realistic_d2}.
Figure~\ref{fig:fuel_interventions} shows the pixels modified by each intervention.

Figure~\ref{fig:fuel_baseline_counterfactual} compares the BP, FI, and ROS maps produced by BurnP3+ and the surrogate under the shared baseline inputs and the two fuel interventions.
We quantify agreement using the CCC, Spearman's rank correlation (\(\rho_s\)), and NMAE.
At baseline, the surrogate reproduces the broad spatial gradients represented by BurnP3+, although its predictions are smoother and exhibit less fine-scale contrast.
The baseline \((\mathrm{CCC}, \rho_s, \mathrm{NMAE})\) scores, reported as mean \(\pm\) standard deviation across three training seeds, are
\((0.948 \pm 0.007,\; 0.956 \pm 0.001,\; 0.233 \pm 0.011)\) for BP,
\((0.763 \pm 0.003,\; 0.813 \pm 0.001,\; 0.253 \pm 0.001)\) for FI, and
\((0.674 \pm 0.001,\; 0.732 \pm 0.002,\; 0.198 \pm 0.001)\) for ROS.

\begin{figure*}[t]
\centering
\includegraphics[width=0.95\textwidth]{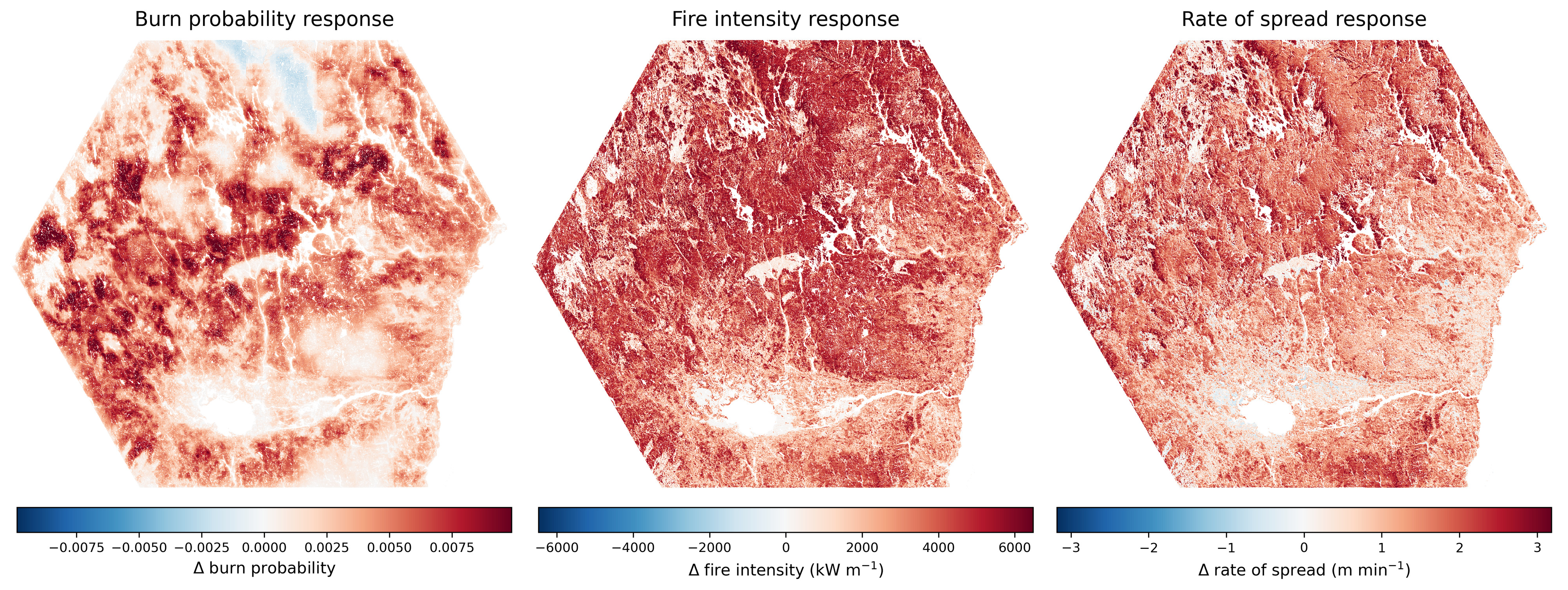}
\caption{Spatial response of the surrogate to the weather intervention in region 16, obtained by replacing the baseline weather inputs with the corresponding inputs from region 17 in British Columbia. The panels show the resulting counterfactual-minus-baseline changes for BP, FI, and ROS from left to right.}
\label{fig:counterfactual_weather}
\end{figure*}
\begin{figure*}[t]
\centering
\IfFileExists{Figures/climate_spread_duration_counterfactual.jpg}{
    \includegraphics[width=0.95\linewidth]{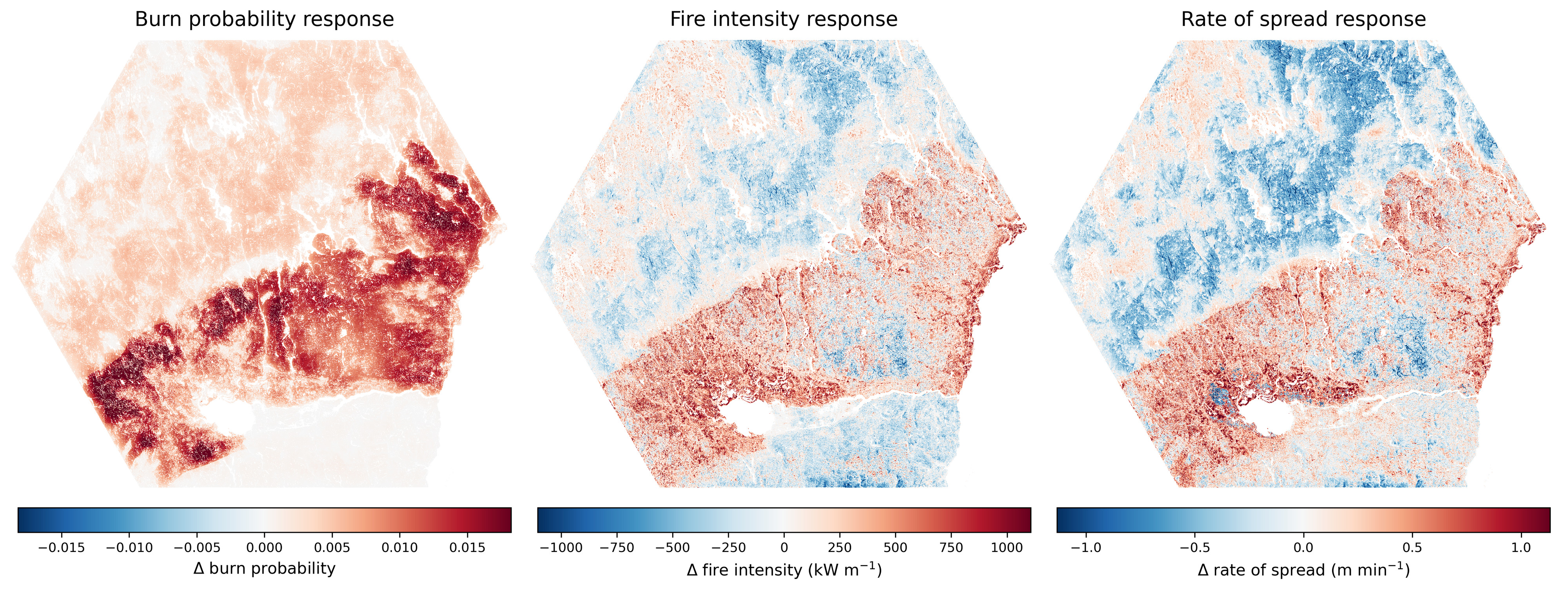}
}{
    \fbox{\parbox[c][0.22\textheight][c]{0.94\textwidth}{
        \centering
        \textbf{Climate-change counterfactual placeholder}\\[0.75em]
        Four panels: change in derived fire size, $\Delta$BP, $\Delta$FI, and $\Delta$ROS
    }}
}
\caption{Spatial response of the surrogate to the climate-change intervention in region 16. From left to right, the panels show the counterfactual-minus-baseline changes in BP, FI, and ROS.}
\label{fig:counterfactual_climate}
\end{figure*}

We assess the counterfactual response by subtracting each model's baseline prediction from its corresponding intervention prediction.
Figure~\ref{fig:fuel_bp3_validation} compares the resulting signed change maps.
As in the baseline comparison, we use CCC, Spearman's rank correlation (\(\rho_s\)), and NMAE to assess agreement between the surrogate and BurnP3+ responses, separately for each intervention and output.

\paragraph{C2-to-M1 substitution.}
The surrogate predicts reductions in mean BP, FI, and ROS of \(74.30\%\), \(62.36\%\), and \(38.35\%\), respectively.
The corresponding BurnP3+ reductions are \(70.64\%\), \(63.78\%\), and \(38.37\%\) (Table~\ref{tab:counterfactual_summary}).
The responses are concentrated around the modified fuel areas, with the surrogate reproducing the broad spatial pattern of the BurnP3+ changes.
The agreement scores \((\mathrm{CCC}, \rho_s, \mathrm{NMAE})\) are \((0.826, 0.924, 0.362)\) for BP, \((0.834, 0.861, 0.323)\) for FI, and \((0.725, 0.781, 0.424)\) for ROS.
BP has the highest agreement in spatial ranking, while FI has the highest CCC and lowest NMAE.

\paragraph{Post-fire D2 intervention.}
The surrogate predicts reductions in region-wide mean BP, FI, and ROS of \(87.59\%\), \(57.78\%\), and \(40.27\%\), respectively.
The corresponding BurnP3+ reductions are \(72.38\%\), \(60.53\%\), and \(48.29\%\) (Table~\ref{tab:counterfactual_summary}).
The surrogate therefore predicts a larger percentage reduction in BP, but smaller reductions in FI and ROS, than BurnP3+.
The changes are concentrated within and around the reassigned polygons, while the broad spatial patterns elsewhere remain similar to the baseline.
The surrogate captures this localization but produces smoother spatial variations than BurnP3+.
The agreement scores \((\mathrm{CCC}, \rho_s, \mathrm{NMAE})\) are \((0.848, 0.861, 0.406)\) for BP, \((0.812, 0.822, 0.354)\) for FI, and \((0.679, 0.740, 0.476)\) for ROS.
BP has the highest CCC and rank correlation, while FI has the lowest NMAE.

Across both interventions, the reductions in BP, FI, and ROS are consistent with the expected response to lower-flammability fuels.
The comparisons with BurnP3+ support the surrogate's ability to capture the principal spatial responses to the two fuel modifications tested here.
Agreement is consistently weaker for ROS than for BP and FI, indicating that reproducing changes in spread rate remains a limitation.

\begin{table*}[t]
\centering
\caption{Aggregate counterfactual responses for region 16. Each entry gives the counterfactual-minus-baseline percentage changes in the region-wide mean.}
\label{tab:counterfactual_summary}
\begin{tabular}{llrrr}
\toprule
\textbf{Intervention}
& \textbf{Model}
& \(\Delta\overline{\mathrm{BP}}\)
& \(\Delta\overline{\mathrm{FI}}\)
& \(\Delta\overline{\mathrm{ROS}}\) \\
\midrule
C2 \(\rightarrow\) M1
& BurnP3+
& $-70.64 \%$
& $-63.78 \%$
& $-38.37\% $ \\
C2 \(\rightarrow\) M1
&AI Surrogate
& \(-74.30\% \pm 3.24\%\)
& \(-62.36\% \pm 0.39\%\)
& \(-38.35\% \pm 0.83\% \) \\
\addlinespace
\midrule
Burned area \(\rightarrow\) D2
& BurnP3+
& $-72.38\%$
& $-60.53\%$
& $-48.29\%$ \\
Burned area \(\rightarrow\) D2
& AI Surrogate
& \(-87.59\% \pm 0.64\%\)
& \( -57.78 \% \pm 0.37\%\)
& \(-40.27\% \pm 0.67 \%\) \\
\addlinespace
\midrule
QC \(\rightarrow\) BC weather
& AI Surrogate
& \(206.18 \% \pm 59.22 \% \)
& \(97.79 \% \pm 0.71\%\) 
& \(23.67\% \pm 1.74\%\) \\
\addlinespace
\midrule
Climate-adjusted spread duration
& AI Surrogate
& \( 271.39 \% \pm 58.22 \%\) 
& \(1.27 \% \pm 0.71\%\) 
& \(-0.20 \% \pm 2.20\%\) \\
\bottomrule
\end{tabular}
\end{table*}

\subsection{Weather Intervention}
In the weather scenario, the baseline weather inputs for region 16 are replaced with the corresponding inputs from region 17, located in warmer and drier southern British Columbia, thereby imposing a more severe western Canadian weather regime.
This severity is reflected in the mean Fire Weather Index (FWI)\footnote{FWI is the final index of the Canadian Forest Fire Weather Index System and summarizes weather-driven fire potential based on fuel moisture and wind conditions \citep{vanwagner1987development}.}, which increases from 19.3 in central Québec to 29.1 in southern British Columbia.
Because the surrogate weather channels include internally related meteorological and FWI-system variables, we transfer the complete set of weather inputs rather than perturbing individual variables in isolation.
We expect the more severe weather regime to increase BP, FI, and ROS.

Imposing the BC weather regime increases the region-wide mean of all three outputs.
Mean BP across the region increases by $\sim 206\%$, mean FI by \(\sim98\%\), and mean ROS by \(\sim 24\%\) (Table~\ref{tab:counterfactual_summary}).
Figure~\ref{fig:counterfactual_weather} shows the spatial pattern of the surrogate response across the weather zones.
The increases in FI and ROS are widespread, whereas the BP response is more heterogeneous, with some areas showing decreases despite the positive region-wide mean change.

The intervention couples the landscape of region 16 with weather data from a geographically distinct region, producing an input configuration absent from the training data.
Under this configuration, the surrogate predicts the expected increases in region-wide mean BP, FI, and ROS, although BP decreases in some parts of the landscape.

\subsection{Climate-Change Intervention}

For the climate-change intervention, we modify the spread-event-day distributions using the projected changes for the 2050s under Representative Concentration Pathway 4.5 (RCP~4.5) reported by \citet{wang2017projected}.\footnote{RCP~4.5 is a stabilization pathway in which radiative forcing reaches approximately \(4.5~\mathrm{W\,m^{-2}}\) by 2100.}
Specifically, we adjust the season- and zone-specific distributions so that their median and 90th-percentile spread durations shift by the projected amounts, i.e., 0.7 days and 5 days, respectively.

In BurnP3+, spread-event duration directly controls the number of days over which a fire can grow.
For the surrogate, we use the regional relationships between the number of realized spread days and final fire size developed by \citet{wang2020projected} to obtain the corresponding fire-size distributions.
Fire size is therefore modified through its relationship with spread duration rather than prescribed independently.

The baseline weather lists are retained for this intervention.
These lists contain days selected as conducive to fire spread based on their FWI conditions.
Increasing the spread-event duration therefore represents fires experiencing longer sequences of fire-conducive weather.
Figure~\ref{fig:counterfactual_climate} shows the resulting counterfactual-minus-baseline changes in the surrogate predictions of BP, FI, and ROS. 

The climate-change intervention produces its largest and most spatially consistent response in BP.
Mean BP increases by \(\sim271\%\), while mean FI increases by \(\sim1\%\) and mean ROS decreases by \(\sim0.2\%\) (Table~\ref{tab:counterfactual_summary}).
The strong BP response is consistent with longer spread-event durations allowing fires to grow larger, thereby affecting a greater proportion of the landscape
The comparatively small changes in FI and ROS are consistent with retaining the baseline fuel and weather inputs, which govern local fire intensity and spread rate.

\subsection{Summary of Counterfactual Responses}
Table~\ref{tab:counterfactual_summary} summarizes the region-wide mean responses across the four counterfactual scenarios.
The two fuel interventions demonstrate the surrogate's robustness to the tested landscape modifications as it reproduces the direction and broad spatial distribution of the BurnP3+ responses under both the C2-to-M1 substitution and the post-fire D2 intervention.
The weather and climate-change interventions produce aggregate responses consistent with the expected effects of more severe fire weather and longer spread-event durations.
Together, these findings support the surrogate's use for rapid counterfactual assessment of wildfire outcomes under alternative fuel and environmental conditions.

\section{Cross-Study Model Transfer}
In addition to the national data used for training, we evaluate the AI surrogate model on an independent regional dataset\footnote{The dataset uses the same data sources and 100-m spatial resolution as the national dataset, with data preprocessing following the same pipeline described for the national dataset.} covering part of the Northwest Territories (NWT), a fire-prone boreal region in Canada, simulated in a separate BurnP3+ study different from the national one.
This out-of-distribution evaluation tests the model's ability to generalize spatial wildfire hazard mapping beyond the national study. Specifically, it subjects the surrogate to regional shifts in simulation parameters, landscape fuel, ignition patterns, weather and fire size distributions, reflecting the variations commonly found between different studies.

Unlike the national dataset, the regional study contains four distinct simulation scenarios representing different underlying input configurations of the same geographical region. Each scenario modifies the fuel input rasters and therefore requires a separate set of prepared input patches, and BurnP3+ reference outputs for burn probability, fire intensity, and rate of spread. Due to confidentiality constraints \footnote{Further details are withheld pending final approvals as this BurnP3+ study remains in progress.}, we refer to these configurations only as Scenarios A, B, C, and D throughout this paper.
Although a portion of the regional study area was included in the national study, specifically within training regions 10 and 13, none of the four fuel scenarios mirror the baseline fuel configuration used to train the AI surrogate model. Furthermore, differences in simulation parameterization and calibration between the two studies lead to distinct BurnP3+ outputs, even over the overlapping geographic area.
A portion of the regional study area geographically overlaps with areas represented in the national training dataset. However, none of the regional scenarios reproduce the input configuration used during national-scale model training. In addition, differences in simulation parameterization and calibration between the two studies produce distinct BurnP3+ outputs, particularly for burn probability, even within geographically overlapping areas. The regional dataset also contains fuel classes that were not represented in the national training data, introducing an additional source of distribution shift.

Evaluating across the four scenarios therefore provides a challenging test of surrogate-model robustness under changes in fuel representation, fire-size distributions, and simulation parameterization. This cross-scenario benchmark helps determine whether the surrogate captures transferable spatial relationships rather than simply reproducing the specific configurations encountered during training.

We designate Scenario A as the reference scenario for the primary zero-shot evaluation and regional fine-tuning experiments. The remaining scenarios are reserved for evaluating the surrogate model's generalization before and after fine-tuning.

\begin{table*}[t]
\centering
\footnotesize
\setlength{\tabcolsep}{4.5pt}
\begin{tabular*}{\linewidth}{
@{\extracolsep{\fill}}
l l r r r r r r
}
\toprule
\textbf{Target}
& \textbf{Model}
& $\downarrow$ \textbf{\shortstack{Normalized \\ MAE}}
& $\rightarrow 0$ \textbf{\shortstack{Normalized \\ Bias}}
& $\uparrow$ \textbf{Spearman}
& $\uparrow$ \textbf{CCC}
& $\uparrow$ \textbf{\shortstack{AUC IoU \\ Top-10\%}}
& $\uparrow$ \textbf{\shortstack{AUC IoU \\ Full}} \\
\midrule
\multicolumn{8}{c}{\textit{\textbf{Scenario A}}} \\
\midrule

\textbf{BP}
& U-Net (Base)
& $4.620 \pm 0.891$
& $4.595 \pm 0.895$
& $0.616 \pm 0.044$
& $0.076 \pm 0.021$
& $0.102 \pm 0.010$
& $0.550 \pm 0.015$ \\

& + Recalibration
& $0.500 \pm 0.057$
& $-0.435 \pm 0.090$
& $0.616 \pm 0.044$
& $0.366 \pm 0.064$
& $0.102 \pm 0.010$
& $0.550 \pm 0.015$ \\

& + Zone Alignment
& $\mathbf{0.335} \pm 0.071$
& $\mathbf{-0.260} \pm 0.113$
& $\mathbf{0.890} \pm 0.004$
& $\mathbf{0.708} \pm 0.093$
& $\mathbf{0.266} \pm 0.018$
& $\mathbf{0.718} \pm 0.006$ \\

\cmidrule{2-8}

\textbf{FI}
& U-Net (Base)
& $\mathbf{0.280} \pm 0.004$
& $-0.106 \pm 0.036$
& $\mathbf{0.812} \pm 0.002$
& $0.740 \pm 0.012$
& $\mathbf{0.142} \pm 0.003$
& $\mathbf{0.652} \pm 0.002$ \\

& + Recalibration
& $0.325 \pm 0.019$
& $0.144 \pm 0.043$
& $\mathbf{0.812} \pm 0.002$
& $\mathbf{0.757} \pm 0.011$
& $\mathbf{0.142} \pm 0.003$
& $\mathbf{0.652} \pm 0.002$ \\

& + Zone Alignment
& $0.320 \pm 0.020$
& $\mathbf{0.125} \pm 0.049$
& $0.809 \pm 0.003$
& $0.756 \pm 0.011$
& $0.130 \pm 0.005$
& $0.649 \pm 0.002$ \\

\cmidrule{2-8}

\textbf{ROS}
& U-Net (Base)
& $0.219 \pm 0.003$
& $-0.065 \pm 0.030$
& $0.795 \pm 0.003$
& $0.740 \pm 0.011$
& $\mathbf{0.136} \pm 0.009$
& $0.641 \pm 0.003$ \\

& + Recalibration
& $\mathbf{0.218} \pm 0.001$
& $\mathbf{-0.023} \pm 0.029$
& $0.795 \pm 0.003$
& $\mathbf{0.744} \pm 0.006$
& $\mathbf{0.136} \pm 0.009$
& $0.641 \pm 0.003$ \\

& + Zone Alignment
& $0.219 \pm 0.002$
& $-0.029 \pm 0.038$
& $\mathbf{0.796} \pm 0.002$
& $0.741 \pm 0.012$
& $0.134 \pm 0.001$
& $\mathbf{0.642} \pm 0.002$ \\

\midrule
\multicolumn{8}{c}{\textit{\textbf{Scenario B}}} \\
\midrule

% BP 
\textbf{BP} 
& U-Net (Base) 
& $5.115 \pm 0.982$ 
& $5.082 \pm 0.989$ 
& $0.584 \pm 0.048$ 
& $0.065 \pm 0.019$ 
& $0.052 \pm 0.005$ 
& $0.530 \pm 0.014$ \\

& + Fine-tuning 
& $\mathbf{0.278} \pm 0.011$
& $\mathbf{0.093} \pm 0.013$
& $\mathbf{0.901} \pm 0.002$
& $\mathbf{0.873} \pm 0.006$
& $\mathbf{0.326} \pm 0.013$
& $\mathbf{0.730} \pm 0.003$ \\

\cmidrule{2-8}

% FI 
\textbf{FI} 
& U-Net (Base) 
& $\mathbf{0.289} \pm 0.003$ 
& $\mathbf{-0.098} \pm 0.036$ 
& $\mathbf{0.809} \pm 0.003$ 
& $\mathbf{0.733} \pm 0.010$ 
& $0.151 \pm 0.001$ 
& $0.652 \pm 0.002$ \\

& + Fine-tuning
& $0.304 \pm 0.015$
& $-0.167 \pm 0.037$
& $0.808 \pm 0.003$
& $0.694 \pm 0.033$
& $\mathbf{0.163} \pm 0.004$
& $\mathbf{0.654} \pm 0.002$ \\

\cmidrule{2-8}

% ROS 
\textbf{ROS} 
& U-Net (Base) 
& $0.227 \pm 0.002$ 
& $-0.055 \pm 0.030$ 
& $0.792 \pm 0.004$ 
& $0.731 \pm 0.009$ 
& $0.143 \pm 0.011$ 
& $0.641 \pm 0.004$ \\

& + Fine-tuning
& $\mathbf{0.220} \pm 0.001$
& $\mathbf{-0.042} \pm 0.017$
& $\mathbf{0.792} \pm 0.002$
& $\mathbf{0.748} \pm 0.002$
& $\mathbf{0.162} \pm 0.008$
& $\mathbf{0.644} \pm 0.002$ \\

\midrule
\multicolumn{8}{c}{\textit{\textbf{Scenario C}}} \\
\midrule

\textbf{BP}
& U-Net (Base)
& $3.812 \pm 0.908$
& $3.741 \pm 0.911$
& $0.416 \pm 0.065$
& $0.062 \pm 0.031$
& $0.067 \pm 0.007$
& $0.486 \pm 0.016$ \\

& + Fine-tuning
& $\mathbf{0.302} \pm 0.010$
& $\mathbf{-0.191} \pm 0.036$
& $\mathbf{0.835} \pm 0.018$
& $\mathbf{0.784} \pm 0.017$
& $\mathbf{0.360} \pm 0.048$
& $\mathbf{0.688} \pm 0.016$ \\

\cmidrule{2-8}

\textbf{FI}
& U-Net (Base)
& $\mathbf{0.364} \pm 0.026$
& $\mathbf{-0.241} \pm 0.042$
& $\mathbf{0.792} \pm 0.017$
& $\mathbf{0.692} \pm 0.036$
& $\mathbf{0.220} \pm 0.010$
& $\mathbf{0.645} \pm 0.012$ \\

& + Fine-tuning
& $0.402 \pm 0.043$
& $-0.310 \pm 0.083$
& $0.778 \pm 0.018$
& $0.635 \pm 0.068$
& $0.209 \pm 0.005$
& $0.633 \pm 0.011$ \\

\cmidrule{2-8}

\textbf{ROS}
& U-Net (Base)
& $\mathbf{0.239} \pm 0.016$
& $\mathbf{-0.073} \pm 0.016$
& $\mathbf{0.768} \pm 0.037$
& $\mathbf{0.710} \pm 0.038$
& $\mathbf{0.191} \pm 0.015$
& $\mathbf{0.634} \pm 0.018$ \\

& + Fine-tuning
& $0.244 \pm 0.013$
& $-0.089 \pm 0.047$
& $0.756 \pm 0.020$
& $0.710 \pm 0.029$
& $0.187 \pm 0.007$
& $0.626 \pm 0.010$ \\

\midrule
\multicolumn{8}{c}{\textit{\textbf{Scenario D}}} \\
\midrule

\textbf{BP}
& U-Net (Base)
& $3.862 \pm 0.900$
& $3.810 \pm 0.904$
& $0.419 \pm 0.069$
& $0.060 \pm 0.029$
& $0.065 \pm 0.008$
& $0.485 \pm 0.017$ \\

& + Fine-tuning
& $\mathbf{0.284} \pm 0.009$
& $\mathbf{-0.181} \pm 0.037$
& $\mathbf{0.835} \pm 0.015$
& $\mathbf{0.786} \pm 0.018$
& $\mathbf{0.363} \pm 0.023$
& $\mathbf{0.688} \pm 0.010$ \\

\cmidrule{2-8}

\textbf{FI}
& U-Net (Base)
& $\mathbf{0.350} \pm 0.026$
& $\mathbf{-0.216} \pm 0.042$
& $\mathbf{0.789} \pm 0.018$
& $\mathbf{0.706} \pm 0.035$
& $\mathbf{0.223} \pm 0.009$
& $\mathbf{0.645} \pm 0.012$ \\

& + Fine-tuning
& $0.387 \pm 0.040$
& $-0.285 \pm 0.084$
& $0.773 \pm 0.020$
& $0.650 \pm 0.067$
& $0.213 \pm 0.005$
& $0.633 \pm 0.012$ \\

\cmidrule{2-8}

\textbf{ROS}
& U-Net (Base)
& $\mathbf{0.229} \pm 0.014$
& $\mathbf{-0.060} \pm 0.016$
& $\mathbf{0.767} \pm 0.035$
& $0.719 \pm 0.037$
& $\mathbf{0.197} \pm 0.013$
& $\mathbf{0.635} \pm 0.018$ \\

& + Fine-tuning
& $0.234 \pm 0.014$
& $-0.073 \pm 0.047$
& $0.753 \pm 0.024$
& $\mathbf{0.721} \pm 0.030$
& $0.196 \pm 0.007$
& $0.626 \pm 0.012$ \\

\bottomrule
\end{tabular*}
\caption{Main results comparing the predictive performance of our multi-source U-Net models for \textbf{Burn Probability (BP)}, \textbf{Fire Intensity (FI)}, and \textbf{Rate of Spread (ROS)} on a \textbf{separate regional study}. U-Net model is based on \textbf{Spatial + Weather} input data. Each value is reported as the mean and standard deviation of 3 different-seed runs. Metrics are reported on the test set \textit{\textbf{at region level}}. Best performance for each specified metric is highlighted in bold.}
\label{tab:NWT_data_weather_model}
\end{table*}

\subsubsection*{Results}
Table~\ref{tab:NWT_data_weather_model} summarizes the results across all four scenarios. Under Scenario A, the regional domain shift affects BP more strongly than FI or ROS. Although the spatial ranking of BP predictions remains moderately preserved, the predicted magnitudes are substantially miscalibrated. To address this, we re-calibrate the ground-truth targets by applying regional rather than national normalization statistics across all three ground-truth targets. \textit{Recalibration} with regional statistics reduces BP MAE from \(4.620\) to \(0.500\) and substantially corrects the bias, while Spearman correlation and IoU remain unchanged, indicating that recalibration primarily corrects prediction scale rather than spatial ranking. Following discussions with the data providers who executed the BurnP3+ simulations, we identified a mismatch between the regional fire-weather-zone \footnote{A fire weather zone is a homogeneous spatial partition used as the primary geographic unit for calibrating and parameterizing wildfire growth simulations} configuration and the configuration most consistent with the national training setup. We therefore apply \textit{Zone Alignment} by mapping the regional input features of the misaligned zone into neighboring zone features that align with the training data distribution. This adjustment provides an additional improvement, increasing Spearman from \(0.616\) to \(0.890\), CCC from \(0.366\) to \(0.708\), and full AUC IoU from \(0.550\) to \(0.718\).

These results indicate that part of the regional BP performance degradation arises from differences in fire-weather-zone feature representation between the national and regional studies. In contrast, FI and ROS remain relatively stable across configurations, with only minor changes that are often comparable to inter-seed variability. Overall, BP is more sensitive to regional recalibration and zone mismatch, whereas FI and ROS generalize more robustly from the nationally trained AI surrogate model.

Across the remaining scenarios, we assess the national model's zero-shot performance and determine how effectively lightweight fine-tuning improves its predictive performance. By leveraging fine-tuning rather than manual recalibration and alignment, a process that typically requires intensive expert intervention and iterative tuning, we demonstrate a practical strategy for adapting surrogate models to new regional settings. For fine-tuning, we use Scenario A as regional training data, comprising 116 data samples, and train for (20) epochs using the same configuration as the national model. The full fine-tuning process takes approximately (15) minutes. We evaluate the surrogate model before and after fine-tuning on the remaining 3 scenarios (B, C and D).

\begin{enumerate}
    \item Under Scenario B, BP again exhibits the largest domain shift, with the Base model showing high MAE and bias alongside moderate spatial rank correlation (spearman). Fine-tuning substantially improves all BP metrics, reducing MAE from \(5.115\) to \(0.278\), increasing Spearman from \(0.584\) to \(0.901\), and improving full AUC IoU from \(0.530\) to \(0.730\). These results indicate that BP is particularly sensitive to the distribution shift introduced by this scenario and benefits strongly from regional adaptation.
    FI remains comparatively robust without fine-tuning. Fine-tuning slightly improves the IoU metrics but degrades MAE, bias, and CCC, suggesting limited overall benefit for this target. For ROS, fine-tuning provides a more consistent improvement, with lower MAE and bias and higher CCC and IoU, while Spearman remains unchanged.
    \item Under Scenario C, BP again benefits substantially from fine-tuning. MAE decreases from \(3.812\) to \(0.302\), while Spearman improves from \(0.416\) to \(0.835\), CCC from \(0.062\) to \(0.784\), and full AUC IoU from \(0.486\) to \(0.688\). This indicates that BP remains highly sensitive to the distribution shift introduced by this scenario.
    In contrast, fine-tuning does not improve FI or ROS. FI performance decreases across all reported metrics, while ROS remains largely unchanged, with slight degradations in MAE, bias, Spearman, and IoU. These results further suggest that the benefits of regional adaptation are strongly target-dependent: BP requires adaptation to the regional setting, whereas FI and ROS transfer more robustly from the national model.
    \item Under Scenario D, BP again shows a strong benefit from regional fine-tuning. MAE decreases from \(3.862\) to \(0.284\), while Spearman improves from \(0.419\) to \(0.835\), CCC from \(0.060\) to \(0.786\), and full AUC IoU from \(0.485\) to \(0.688\). This indicates that BP remains highly sensitive to the distribution shift introduced by this scenario. By contrast, FI consistently performs better without fine-tuning across all reported metrics. ROS is also largely unchanged, with the Base model retaining slightly better MAE, bias, Spearman, and IoU, while fine-tuning provides only a negligible improvement in CCC.

\end{enumerate}

Overall, all scenarios exhibit the same pattern: fine-tuning is critical for BP, as cross-study variations in simulation parameters, specifically number of simulation iterations  and fire size distributions, heavily impact burn probability estimates, whereas FI and ROS transfer comparatively well from the national model.

We repeat all evaluations using the spatial + weather + fire size model and report the results in Appendix~\ref{app:nwt_data}, Table~\ref{app:tab:NWT_data_weather_firesize_model}. Overall, the model exhibits a similar target-dependent pattern to the spatial + weather model, with BP benefiting most from regional adaptation and FI remaining comparatively robust. However, its Base model generalizes less reliably because the regional fire-size distribution differs substantially from that used in the national study. Fine-tuning largely recovers this degradation. Under Scenario B, BP reaches \(0.901\ \) Spearman and \(0.732\ \) full AUC IoU, while under Scenarios C and D it substantially improves BP and ROS performance. FI remains broadly consistent with the previous model, with fine-tuning providing only limited or mixed gains.

These findings demonstrate the strong transferability and robustness of the AI surrogate model across distinct out-of-distribution scenarios, confirming it learns transferable spatial relationships rather than memorizing training data. The model shows strong zero-shot robustness by directly generalizing Fire Intensity (FI) and Rate of Spread (ROS) across unseen regional fuels and parameter shifts without adaptation. Meanwhile, for Burn Probability (BP), which is highly sensitive to regional calibration, the surrogate exhibits strong adaptability, as lightweight fine-tuning consistently resolves domain shifts across all scenarios, reducing MAE by over $90\%$ and raising Spearman correlation above $0.83$. This combination of out-of-the-box physical generalization and rapid adaptation highlights the model's reliability across diverse study settings.

\section{Conclusion}
We present an AI surrogate modelling framework and train a deep learning surrogate model for the Canadian BurnP3+ wildfire simulator. A single multi-task model approximates burn probability, fire intensity, and rate of spread from spatial, weather, and fire size input features, successfully approximating BurnP3+-derived wildfire hazard. Inference runs $\sim 806 \times$ faster than BurnP3+ on CPU hardware, shifting a regional scenario from days to seconds and making interactive scenario exploration practical for wildfire scientists. 
Counterfactual sensitivity analysis shows that the surrogate responds coherently to controlled changes in the input conditions, with prediction shifts that are directionally consistent with the corresponding BurnP3+ outputs in the verified scenarios. This provides evidence that the model is not only reproducing aggregate patterns, but is also sensitive to meaningful changes in the underlying simulation drivers. 
% Revert back if section 7 is added back
In addition, the regional data experiments demonstrate strong model transferability. Despite being trained on national-scale data, the surrogate retains useful predictive structure when applied to a separate regional study, and lightweight fine-tuning on a small amount of regional data substantially improves performance where domain shift is strongest. 
Together, these results suggest that the surrogate can generalize beyond its original training setting while requiring only limited adaptation to recover high-fidelity regional predictions.
Ultimately, as a simulator surrogate rather than a direct fire predictor, the model inherits BurnP3+'s limitations and is intended to complement expert judgment in high-stakes decisions.
% Several limitations bound these results. Scenario analysis covers one region and two interventions; magnitude agreement remains unverified against reference simulator runs; the burn probability target appears more sensitive to weather perturbations, lacking global robustness. Future work will use expanded BurnP3+ simulations to calibrate response magnitudes and test whether the model captures anisotropic, wind-aligned spread. Ultimately, as a simulator surrogate rather than a direct fire predictor, the model inherits BurnP3+'s limitations and is intended to complement, not replace, expert judgment in high-stakes decision-making.
\paragraph{Broader Impact.} Beyond computational speedups, this framework bridges complex simulation models and operational decision-making. By making rapid scenario testing feasible for scientists and emergency planners, it facilitates timely risk assessment and strengthens disaster resilience in the face of escalating wildfire threats.

% Counterfactual analysis suggests that the model goes beyond memorizing region-specific input–output relationships, even if it captures response direction rather than exact magnitude, while establishing a systematic framework for out-of-distribution testing.
% add a sentence about broader impactwhat 
% \subsection{Broader Impact}

% \subsubsection{Content Appendices.}
% Content appendices must appear after the main content. If your main sections are numbered, appendix sections must use letters instead of arabic numerals. In \LaTeX{} you can use the \texttt{\textbackslash appendix} command to achieve this effect and then use \texttt{\textbackslash section\{Heading\}} normally for your appendix sections.

% \subsubsection{Ethical Statement.}
% You can write a statement about the potential ethical impact of your work, including its broad societal implications, both positive and negative. If included, such statement must be written in an unnumbered section titled \emph{Ethical Statement}.

\section*{Acknowledgments}
This research was enabled in part by compute resources, software, and technical support provided by Mila (mila.quebec). The authors would also like to express their gratitude to Renee Larsen (Natural Resources Canada) for providing part of the data used in this study and for her valuable feedback on the manuscript.

\bibliography{aaai2027}

% Check whether the conference requires a reproducibility checklist to be included in the paper.
% \input{ReproducibilityChecklist.tex}

\appendix
\newpage
\section{Codebase and Trained Models}

The codebase and trained models are available in a public repository {
\mbox{\url{https://github.com/milatechtransfer/AISurrogate_wildfireriskmapping}}
}.
Model checkpoints are provided in the \texttt{model\_checkpoints} directory. 
All materials are available through this repository to facilitate access and navigation.

% The codebase and models are available \url{https://github.com/milatechtransfer/AISurrogate_wildfireriskmapping},
% with model checkpoints provided in the \nolinkurl{model_checkpoints} directory.

The dataset is subject to restricted access as advised by our wildfire expert collaborators to prevent misuse and misinterpretation, but the data custodians may consider access upon reasonable request for research purposes. Unrestricted public release poses risks, as raw wildfire hazard maps can be easily misinterpreted by commercial entities (e.g., insurance companies), potentially leading to adverse socio-economic impacts on local communities and public safety. 

\section{Dataset} \label{app:data_prep}
In this section, we provide additional details on the preprocessing and preparation of the spatial data used as BurnP3+ inputs and outputs. The raw per-region spatial datasets, including the DEM, fuel/FBP layers, fire zones, ignition grids, and masks, are provided at 100-m spatial resolution in region-specific, locally centered coordinate reference systems. Each region uses a custom Transverse Mercator projection based on the NAD83 geographic datum (EPSG:4269), with the central meridian adjusted to the longitude of the corresponding region.
%a scale factor of 0.9996, and a false easting of 500,000 m.

Because each region has a distinct projection, spatial extent, and raster grid, the native rasters cannot be directly aligned or stacked across regions. To enable consistent national-scale processing, the data preparation pipeline reprojects all rasters to a common target coordinate reference system, ESRI:102002 (Canada Albers Equal Area Conic), and aligns them to a shared spatial grid and transform. The reprojected rasters are then clipped, organized into model-ready patches, and used for national-scale mosaicking and model training. The same reprojection and preprocessing pipeline is applied consistently to the additional datasets incorporated in the study.

To ensure numerical stability and consistent optimization across multi-modal inputs, different normalization strategies are applied to the input grids, tabular inputs, and target variables based on their data characteristics. 

\begin{figure}[t]
    \centering
    \includegraphics[width=\columnwidth]
        {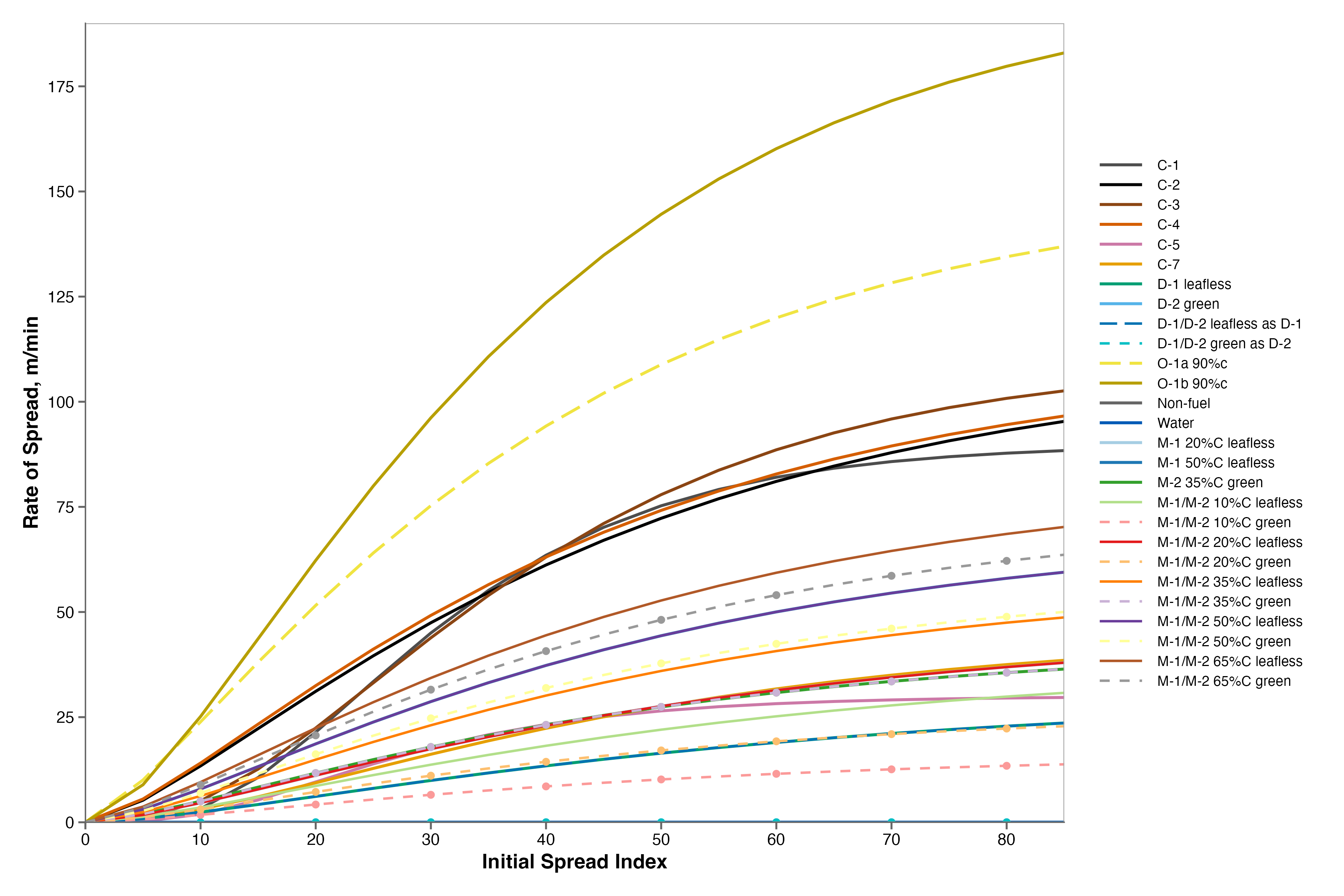}
    \caption{FBP-generated initial Rate of Spread (iROS) curves across fuel classes as a function of the Initial Spread Index (ISI). These curves serve as a continuous functional representation of fuel characteristics, replacing discrete fuel codes.}
    \label{fig:fuel_iROS_curves}
\end{figure}

\paragraph{Fuel Grid.} Fuel data are derived from the Spatialized Canadian Forest Inventory following the Canadian Forest Fire Behaviour Prediction (FBP) system \citep{guindon2024new, wotton2009updates}. The underlying map originally includes approximately $30$ distinct categorical fuel classes, which are summarized in Table~\ref{tab:fbp_restricted_fuel_mapping}. Rather than utilizing high-dimensional one-hot encodings, which treat fuel types as mutually exclusive categories lacking physical continuity, we employ a domain-specific continuous parameterization based on expert-verified FBP equations \citep{fbpSystem1992, cffdrPackage}.
For each fuel type, we calculate its initial Rate of Spread ($iROS$) curve across a range of Initial Spread Index ($ISI$) values under fixed baseline environmental conditions (see Figure~\ref{fig:fuel_iROS_curves}). Then, categorical ordinal fuel codes are mapped through a non-linear $iROS$-curve lookup to yield an 18-dimensional feature vector representing initial spread dynamics (from the curves shown). To mitigate skewness and stabilize variance across disparate fuel response profiles, each 18-value feature vector undergoes a logarithmic transformation:
\begin{equation}
    \mathbf{x}_{\text{log}} = \log(1 + \mathbf{x}_{\text{iROS}})
\end{equation}

The transformed vectors are subsequently standardized using training-set-wide statistics:
\begin{equation}
    \mathbf{x}_{\text{norm}} = \frac{\mathbf{x}_{\text{log}} - \mu_{\text{curve}}}{\sigma_{\text{curve}}}
\end{equation}
where $\mu_{\text{curve}}$ and $\sigma_{\text{curve}}$ denote the mean and standard deviation of the log-transformed curves calculated across the training dataset, respectively.

Next, the standardized 18-dimensional feature vector is passed through a small, dedicated neural network that learns a compressed 4-dimensional embedding, trained as part of our multi-task U-Net model. This low-dimensional representation forms the continuous spatial representation ($\text{Fuel Grid}$) for downstream modelling. By mapping discrete categories to physically grounded response curves and a continuous parameter space, this framework provides a continuous representation that may facilitate transfer to related fuel classes not present in training.

Finally, fuel features are used to represent fuel characteristics through 4 spatial input channels.

\begin{table*}[t]
\centering
\small
\begin{tabular}{r l l c c l}
\toprule
\textbf{ID} & \textbf{Name} & \textbf{Description} & \textbf{fbp\_code} & \textbf{\textit{FuelType}} & \textbf{\textit{SeasonState}} \\
\midrule
1   & Spruce-Lichen Woodland              & C-1     & 1   & C-1  & direct   \\
2   & Boreal Spruce                       & C-2     & 2   & C-2  & direct   \\
3   & Mature Jack or Lodgepole Pine       & C-3     & 3   & C-3  & direct   \\
4   & Immature Jack or Lodgepole Pine     & C-4     & 4   & C-4  & direct   \\
5   & Red and White Pine                  & C-5     & 5   & C-5  & direct   \\
6   & Ponderosa Pine - Douglas-Fir        & C-7     & 7   & C-7  & direct   \\
\addlinespace
7   & Leafless Aspen                      & D-1     & 11  & D-1  & leafless \\
8   & Green Aspen (with BUI Thresholding) & D-2     & 12  & D-2  & green    \\
9   & Aspen                               & D-1/D-2 & 13  & D-1  & leafless \\
10  & Aspen                               & D-1/D-2 & 13  & D-2  & green    \\
\addlinespace
11  & Matted Grass                        & O-1a    & 31  & O-1a & direct   \\
12  & Standing Grass                      & O-1b    & 32  & O-1b & direct   \\
\addlinespace
13  & Non-fuel                            & Non-fuel& 101 & NF   & nonfuel  \\
14  & Water                               & Non-fuel& 102 & WA   & water    \\
\addlinespace
15  & Boreal Mixedwood - Leafless (20\% Conifer) & M-1 (20 PC)     & 420 & M-1 & leafless \\
16  & Boreal Mixedwood - Leafless (50\% Conifer) & M-1 (50 PC)     & 450 & M-1 & leafless \\
17  & Boreal Mixedwood - Green (35\% Conifer)    & M-2 (35 PC)     & 535 & M-2 & green    \\
\addlinespace
18  & Boreal Mixedwood (10\% Conifer)            & M-1/M-2 (10 PC) & 610 & M-1 & leafless \\
19  & Boreal Mixedwood (10\% Conifer)            & M-1/M-2 (10 PC) & 610 & M-2 & green    \\
20  & Boreal Mixedwood (20\% Conifer)            & M-1/M-2 (20 PC) & 620 & M-1 & leafless \\
21  & Boreal Mixedwood (20\% Conifer)            & M-1/M-2 (20 PC) & 620 & M-2 & green    \\
22  & Boreal Mixedwood (35\% Conifer)            & M-1/M-2 (35 PC) & 635 & M-1 & leafless \\
23  & Boreal Mixedwood (35\% Conifer)            & M-1/M-2 (35 PC) & 635 & M-2 & green    \\
24  & Boreal Mixedwood (50\% Conifer)            & M-1/M-2 (50 PC) & 650 & M-1 & leafless \\
25  & Boreal Mixedwood (50\% Conifer)            & M-1/M-2 (50 PC) & 650 & M-2 & green    \\
26  & Boreal Mixedwood (65\% Conifer)            & M-1/M-2 (65 PC) & 665 & M-1 & leafless \\
27  & Boreal Mixedwood (65\% Conifer)            & M-1/M-2 (65 PC) & 665 & M-2 & green    \\
\bottomrule
\end{tabular}
\caption{Canadian fuel types present in our dataset, according to the FBP system.}
\label{tab:fbp_restricted_fuel_mapping}
\end{table*}

\paragraph{Elevation and Terrain Derivatives.} Raw elevation $z_{\text{raw}}$ is first min-max normalized to $z_{\text{norm}} \in [0, 1]$ using global scene bounds ($E_{\text{min}}$ and $E_{\text{max}}$) computed exclusively over the training split:
\begin{equation}
    z_{\text{norm}} = \frac{z_{\text{raw}} - E_{\text{min}}}{E_{\text{max}} - E_{\text{min}}}
\end{equation}

Slope and aspect channels do not undergo standard feature-wide scaling. Instead, they are derived post-hoc from physical elevation in meters $z_{\text{m}}$, recovered via:
\begin{equation}
    z_{\text{m}} = z_{\text{norm}} \times (E_{\text{max}} - E_{\text{min}}) + E_{\text{min}}
\end{equation}

Spatial derivatives along the raster grid dimensions ($\text{row}$ and $\text{col}$) are calculated via finite differences to estimate the gradient magnitude $G$:
\begin{equation}
    G = \sqrt{\left(\frac{\partial z_{\text{m}}}{\partial \text{row}}\right)^2 + \left(\frac{\partial z_{\text{m}}}{\partial \text{col}}\right)^2}
\end{equation}

Slope is calculated in radians using $\arctan(G)$ and normalized by the maximum possible slope ($\pi/2$), yielding an intrinsically bounded channel $\text{slope} \in [0, 1]$:
\begin{equation}
    \text{slope} = \frac{\arctan(G)}{\pi / 2}
\end{equation}

Aspect is parameterized as a two-dimensional downslope unit vector $(\text{aspect\_sin}, \text{aspect\_cos}) \in [-1, 1]^2$ in raster coordinates. For flat terrain ($G \le 10^{-12}$), both components are set to zero to avoid numerical instability; otherwise, they are defined as:
\begin{equation}
    \text{aspect\_sin} = \frac{-\partial z_{\text{m}} / \partial \text{col}}{G}, \quad \text{aspect\_cos} = \frac{\partial z_{\text{m}} / \partial \text{row}}{G}
\end{equation}

Finally, elevation, slope, and aspect (two channels) are used as 4 spatial input channels. 

\paragraph{Ignition Grids.} The raw ignition grids, originally partitioned by season and cause, are aggregated into two spatial layers. Specifically, we compute two seasonal-weighted grids, one for natural lightning and one for human-caused ignition, based on the seasonal ignition distribution used in the simulation framework to produce an ignition frequency map for each region. These channels are passed into the model as raw continuous values. No explicit min-max scaling or standardization is applied (they are in the range 0-1 by default), though missing values are handled via mean imputation using training-split statistics. These make two spatial input channels. 

\paragraph{Tabular Weather Inputs.} We utilize the historical weather data provided by the data provider, which are used directly as inputs to BurnP3+. The weather features include Temperature, Relative Humidity, Precipitation, $\text{wind}_x$, $\text{wind}_y$, Fine Fuel Moisture Code (FFMC), Duff Moisture Code (DMC), Drought Code (DC), Initial Spread Index (ISI), Buildup Index (BUI), and the Fire Weather Index (FWI). Preprocessing involves log-transforming Precipitation to correct for severe skewness, min-max normalizing Relative Humidity and the FFMC, and applying $z$-score standardization to Temperature, the wind vector components ($\text{wind}_x$, $\text{wind}_y$), and the remaining indices (DMC, DC, ISI, BUI, FWI) to manage outliers and stabilize variance across diverse numerical ranges.

\subsubsection*{Targets.}
Target variables, specifically burn probability (BP), fire intensity (FI), and rate of spread (ROS), are normalized independently from the input features based on the training split. Burn probability is scaled using min-max normalization to the range $[0, 1]$. For fire intensity and rate of spread, log-standardization is applied; the raw target values are transformed using a $\log(1+x)$ operation clipped at zero, and subsequently $z$-score standardized using the training-derived log mean ($\mu_{\log}$) and standard deviation ($\sigma_{\log}$):
    \begin{equation}
        y_{\text{norm}} = \frac{\log(1 + \max(0, y_{\text{raw}})) - \mu_{\log}}{\sigma_{\log}}
    \end{equation}

Finally, all feature inputs and outputs are summarized in Table~\ref{tab:feature_summary}.

\begin{table*}[!ht]
\centering
\begin{tabular}{@{} p{4.5cm} p{4.5cm} c p{6.5cm} @{}}
\toprule
\textbf{Feature Name} & \textbf{Source} & \textbf{Dimension} & \textbf{Processing} \\ \midrule

\multicolumn{4}{@{}l}{\textbf{Input: Spatial}} \\ \midrule
Fuel & Forest Fire Behaviour Prediction (FBP) System \citep{fbpSystem1992} & $H \times W \times 4$ & Neural network compression of 18-D iROS curves \\
Ignition & \citep{erni2024mapping} & $H \times W \times 2$ & Cause-aggregated layers with mean imputation \\
Elevation & Digital Elevation Model \citep{nrcan_mrdem} & $H \times W \times 1$ & Min-max normalization (train-split statistics) \\ 
Terrain Derivatives & Derived from Elevation & $H \times W \times 3$ & Slope via $\arctan(G)/(\pi/2)$; Aspect via unit vector ($\sin, \cos$) \\
\midrule

\multicolumn{4}{@{}l}{\textbf{Input: Tabular (Weather \& Environmental Context)}} \\ \midrule
Temperature & ECCC \citep{eccc_canada} \& CWFIS \citep{nrcan_cwfis} & $n \times 1$ & $z$-score standardization \\
Relative Humidity & ECCC \& CWFIS & $n \times 1$ & Min-max normalization \\
Precipitation & ECCC \& CWFIS & $n \times 1$ & Logarithm of plus 1 ($\log(1+x)$) \\
$\text{wind}_x$ & ECCC \& CWFIS & $n \times 1$ & Spatial vector component, $z$-score standardization \\
$\text{wind}_y$ & ECCC \& CWFIS & $n \times 1$ & Spatial vector component, $z$-score standardization \\
Fine Fuel Moisture Code (FFMC) & ECCC \& CWFIS & $n \times 1$ & Min-max normalization \\
Duff Moisture Code (DMC) & ECCC \& CWFIS & $n \times 1$ & $z$-score standardization \\
Drought Code (DC) & ECCC \& CWFIS & $n \times 1$ & $z$-score standardization \\
Initial Spread Index (ISI) & ECCC \& CWFIS & $n \times 1$ & $z$-score standardization \\
Buildup Index (BUI) & ECCC \& CWFIS & $n \times 1$ & $z$-score standardization \\
Fire Weather Index (FWI) & ECCC \& CWFIS & $n \times 1$ & $z$-score standardization \\
Fire Size Distribution & National Fire Database \citep{erni2024mapping} & $n \times 3$ & Logarithm of plus 1, min-max normalization, 10\%, 50\% and 90\% percentiles. \\ 
\midrule

\multicolumn{4}{@{}l}{\textbf{Output: Spatial Targets}} \\ \midrule
Burn Probability (BP) & BurnP3+ Simulation & $H \times W \times 1$ & Min-max normalization \\
Fire Intensity (FI) & BurnP3+ Simulation & $H \times W \times 1$ & $\log(1+x)$ clipped at 0, $z$-score standardization \\
Rate of Spread (ROS) & BurnP3+ Simulation & $H \times W \times 1$ & $\log(1+x)$ clipped at 0, $z$-score standardization \\ 
\bottomrule
\end{tabular}
\caption{Summary of input features and output targets, detailing spatial and tabular dimensions along with exact preprocessing configurations and sources.}
\label{tab:feature_summary}
\end{table*}

\section{Evaluation Metrics} \label{app:eval_metrics}

This section provides the mathematical definitions for the evaluation metrics used to quantify model performance. Each metric is selected to capture specific dimensions of predictive accuracy, including point-wise error, linear correlation, distributional alignment, and spatial overlap.

\subsection{Normalized Mean Absolute Error (Normalized MAE) and Bias (Normalized Bias)}
The Mean Absolute Error (MAE) computes the average magnitude of absolute point-wise differences, which we normalize per target to ensure comparability across different target scales. Bias computes the average \textbf{directional} offset of the predictions.
% \begin{equation}
%     \label{eq:mae}
%     \text{MAE} = \frac{1}{N} \sum_{i=1}^{N} |\hat{y}_i - y_i|
% \end{equation}
\begin{equation}
    \label{eq:nmae}
    \text{Normalized MAE} = \frac{\sum_{i \in \mathcal{V}} |\hat{y}_i - y_i|}{\max\left(\sum{i \in \mathcal{V}} y_i,\ \epsilon\right)}
\end{equation}

% \begin{equation}
%     \label{eq:bias}
%     \text{Bias} = \frac{1}{N} \sum_{i=1}^{N} (\hat{y}_i - y_i)
% \end{equation}
\begin{equation}
    \label{eq:nbias}
    \text{Normalized Bias} = \frac{\sum_{i \in \mathcal{V}} (\hat{y}i - y_i)}{\max\left(\sum{i \in \mathcal{V}} y_i,\ \epsilon\right)}
\end{equation}

\subsection{Spearman Correlation}
Spearman's rank correlation coefficient evaluates the monotonic relationship by applying the Pearson correlation formula to the ranks of the data. Let $R(y_i)$ and $R(\hat{y}_i)$ be the assigned ranks.
\begin{equation}
    \label{eq:spearman}
    \rho = \frac{\text{cov}(R(\hat{y}), R(y))}{\sigma_{R(\hat{y})} \sigma_{R(y)}}
\end{equation}

\subsection{Concordance Correlation Coefficient (CCC)}
The Concordance Correlation Coefficient (CCC) measures agreement between two variables by accounting for both correlation and systematic bias (shifts in scale and location). It serves as a standard evaluation metric for surrogate models in industrial applications.
\begin{equation}
    \label{eq:ccc}
    \text{CCC} = \frac{2 \sigma_{\hat{y}y}}{\sigma_{\hat{y}}^2 + \sigma_y^2 + (\mu_{\hat{y}} - \mu_y)^2}
\end{equation}

% \subsection{Kullback-Leibler (KL) Divergence}
% KL Divergence measures how the predicted probability distribution $Q$ diverges from the target probability distribution $P$. Note that raw values are normalized such that $P_i = y_i / \sum y_i$ and $Q_i = \hat{y}_i / \sum \hat{y}_i$.
% \begin{equation}
%     \label{eq:kl-div}
%     \text{KL}(P \parallel Q) = \sum_{i=1}^{N} P_i \log\left(\frac{P_i}{Q_i}\right)
% \end{equation}

\subsection{Intersection over Union (IoU) and AUC-IoU}
Let $\tau_{y, K}$ and $\tau_{\hat{y}, K}$ be the threshold values at the $K$-th percentile for the targets and predictions, respectively. We binarize the maps using the indicator function $\mathbb{I}$, creating masks $B$ and $\hat{B}$.
\begin{equation}
    \label{eq:topK-IoU-masks}
    B_i = \mathbb{I}(y_i \ge \tau_{y, K}) \quad \text{and} \quad \hat{B}_i = \mathbb{I}(\hat{y}_i \ge \tau_{\hat{y}, K})
\end{equation}
\begin{equation}
    \label{eq:topK-IoU}
    \text{IoU}_K = \frac{\sum_{i=1}^{N} (\hat{B}_i \land B_i)}{\sum_{i=1}^{N} (\hat{B}_i \lor B_i)}
\end{equation}

To approximate the continuous integral of the Top-K IoU across a defined range of percentile thresholds $K \in [k_{\text{min}}, k_{\text{max}}]$, we use the discrete trapezoidal rule. Let $S$ be the number of evaluation steps, defining a discrete sequence $K_1, K_2, \dots, K_S$ where $K_1 = k_{\text{min}}$ and $K_S = k_{\text{max}}$. The normalized discrete AUC is computed as:
\begin{equation}
    \label{eq:auc-iou}
    \text{AUC-IoU} = \frac{1}{k_{\text{max}} - k_{\text{min}}} \sum_{s=1}^{S-1} \frac{\text{IoU}_{K_s} + \text{IoU}_{K_{s+1}}}{2} (K_{s+1} - K_s)
\end{equation}

Integrating over this range distills the model's spatial accuracy across a wide spectrum of risk thresholds into a single metric. For our evaluations, we compute two specific variants using a discrete step size of $1\%$: the Top-10\% AUC-IoU ($K \in [90, 100]$) to isolate the model's ability to localize extreme risk areas, and the Full AUC-IoU ($K \in [1, 100]$) to evaluate spatial accuracy across all risk thresholds.

\section{Experimental Details} \label{app:experimental_details}
\textbf{XGBoost:} The baseline is trained on the same spatial and fuel-derived inputs as our multi-task U-Net. We use $500$ estimators with a maximum depth of $8$, a learning rate of $0.05$ and row/column subsampling of $0.8$ each, using the histogram-based tree method. Training and evaluation together take less than one hour on 8 CPUs.

\textbf{Multi-task U-Net:} All U-Net models (Spatial, Spatial + Weather) are trained using Nvidia A100 GPUs, for $25$ epochs, using $64$-GB of RAM memory and $4$ CPUs. We evaluated training durations of 50 and 100 epochs; however, these did not yield performance improvements, as the validation loss plateaued around approximately 20-25 epochs. A single training cycle takes from 5 to 10 hours, using a batch size of $64$. For each model, we report the mean values and standard deviation of 3 different training runs. The seeds used are 42, 1337, and 2024.

We use a learning rate of $7.0e-4$, along with a sequential learning rate schedule consisting of a linear warmup for the first 5 epochs (starting from $1\%$ of the initial learning rate) followed by a cosine annealing decay strategy over the remaining training steps. For the objective function, we employ a multi-task weighted loss where burn probability, fire intensity, and rate of spread are weighted at 0.5, 0.25, and 0.25, respectively. A higher weight is assigned to burn probability as it represents the most challenging task to learn. For burn probability, we use a composite loss function that balances distribution matching and spatial ranking, linearly combining (weights in parentheses) Kullback-Leibler divergence (0.45), Concordance Correlation Coefficient (0.45), and spatial pairwise ranking losses calculated across both mean region values (0.05) and top-$10$ percentile intensities (0.05). For the fire intensity and rate-of-spread heads, we use a weighted objective function that combines regression error and linear correlation, weighting a Huber loss ($\beta = 1.0$) at 0.7 alongside a raw Pearson correlation coefficient loss at 0.3. We initially evaluated several alternative loss functions, including Binary Cross-Entropy (BCE), Mean Squared Error (MSE), Mean Absolute Error (MAE), Dice, and Focal loss, alongside various weighting strategies across tasks. The final loss functions and task weightings were empirically optimized based on the best validation set performance. 

The final set of parameters and configurations is available under \textit{configs/} in our codebase. All models were implemented in Python using PyTorch and trained in a Linux environment. Complete specifications for the development environment and package dependencies are provided in our code repository.
% Add all information about numbers and ranges of values tried.

\section{Additional Results} \label{appendix:additional_results}

\begin{figure*}[t]
    \centering
    % Row 1: Target and prediction distributions
    \begin{subfigure}[t]{0.32\textwidth}
        \centering
        \includegraphics[width=\linewidth,trim={0 0 0 20},clip]{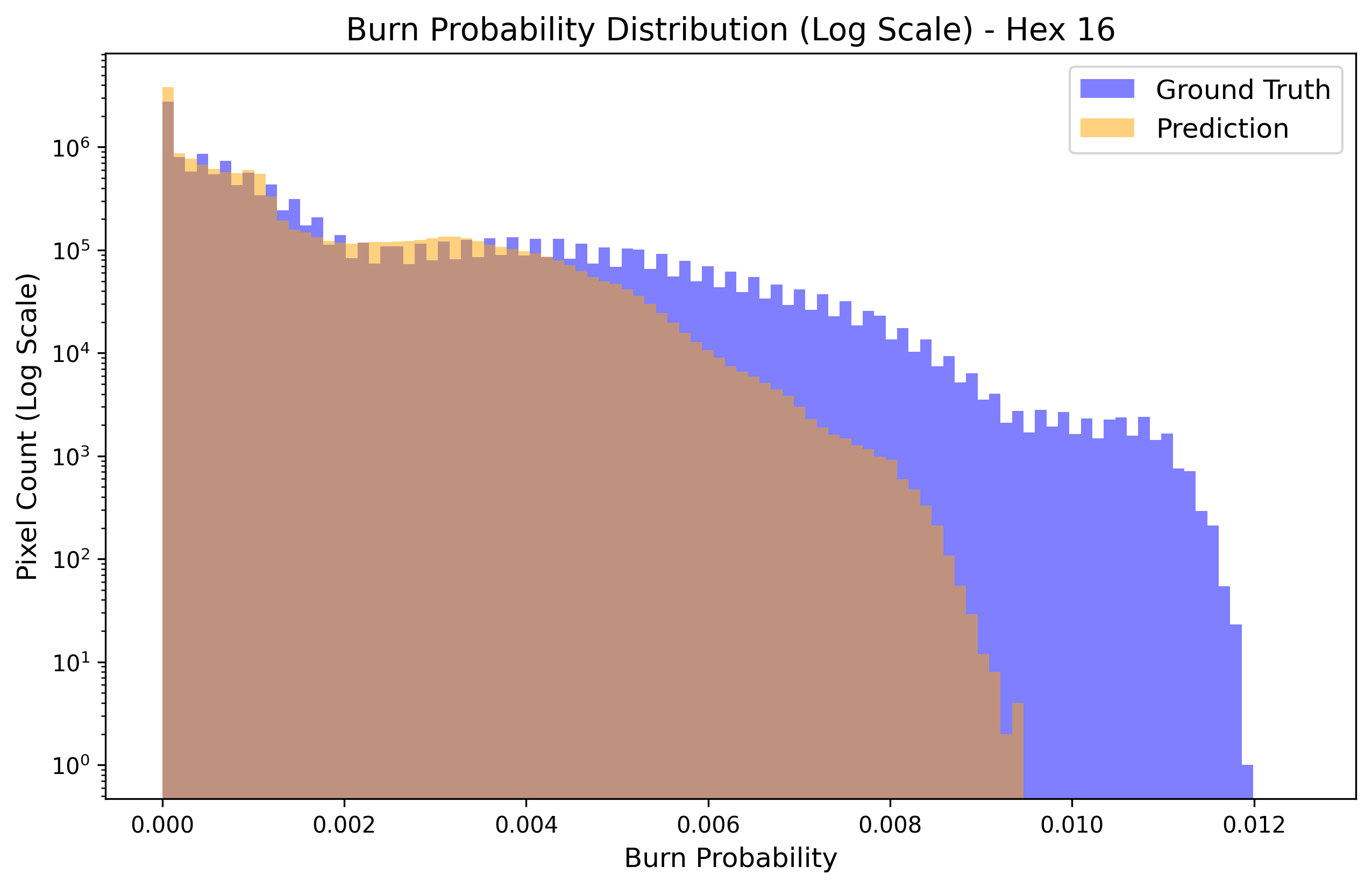}
        \caption{Probability density comparison plot: Burn Probability}
        \label{fig:hex16_bp_hist}
    \end{subfigure}
    \hfill
    \begin{subfigure}[t]{0.32\textwidth}
        \centering
        \includegraphics[width=\linewidth,trim={0 0 0 20},clip]{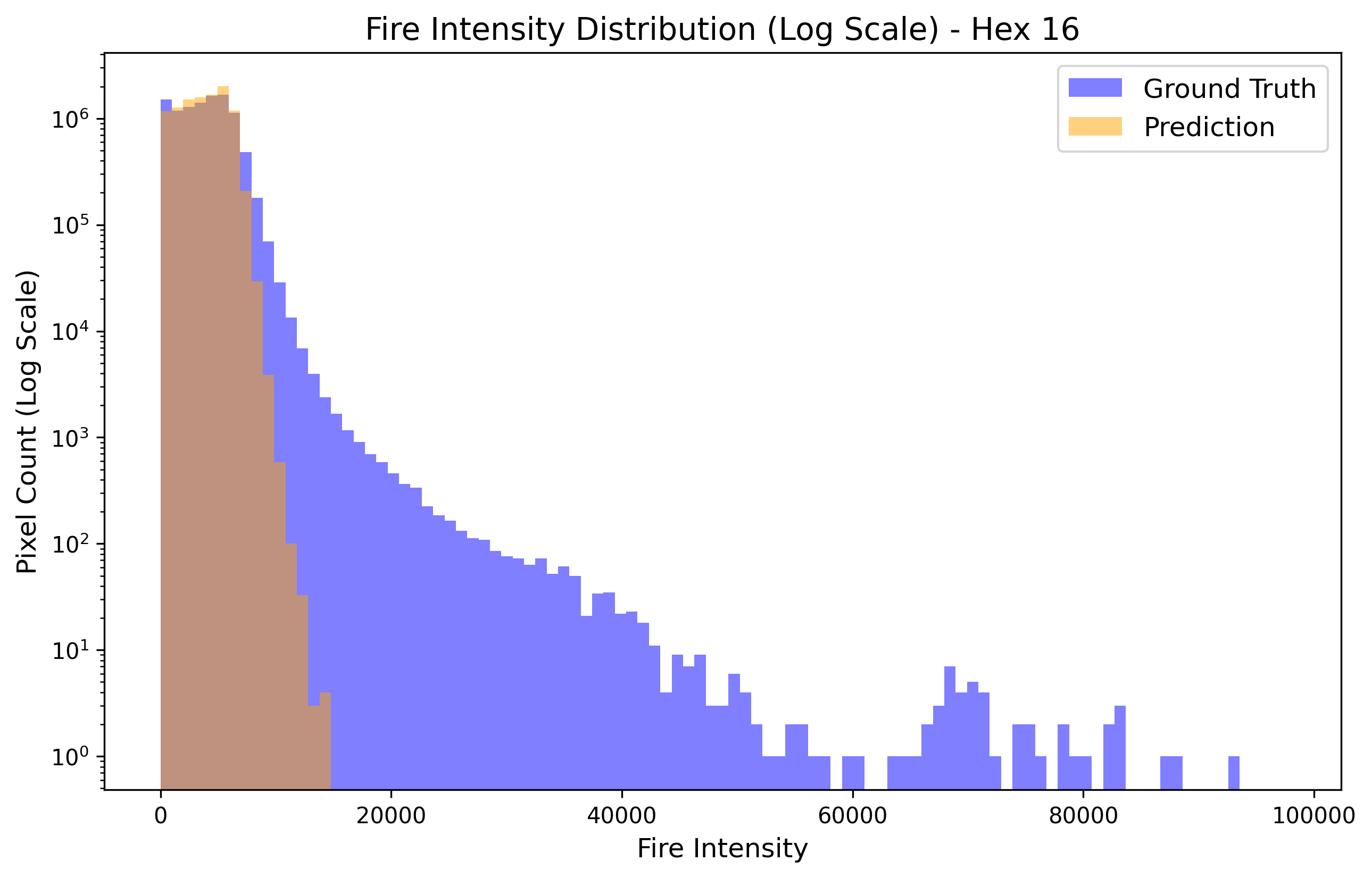}
        \caption{Probability density comparison plot: Fire Intensity}
        \label{fig:hex16_fi_hist}
    \end{subfigure}
    \hfill
    \begin{subfigure}[t]{0.32\textwidth}
        \centering
        \includegraphics[width=\linewidth,trim={0 0 0 20},clip]{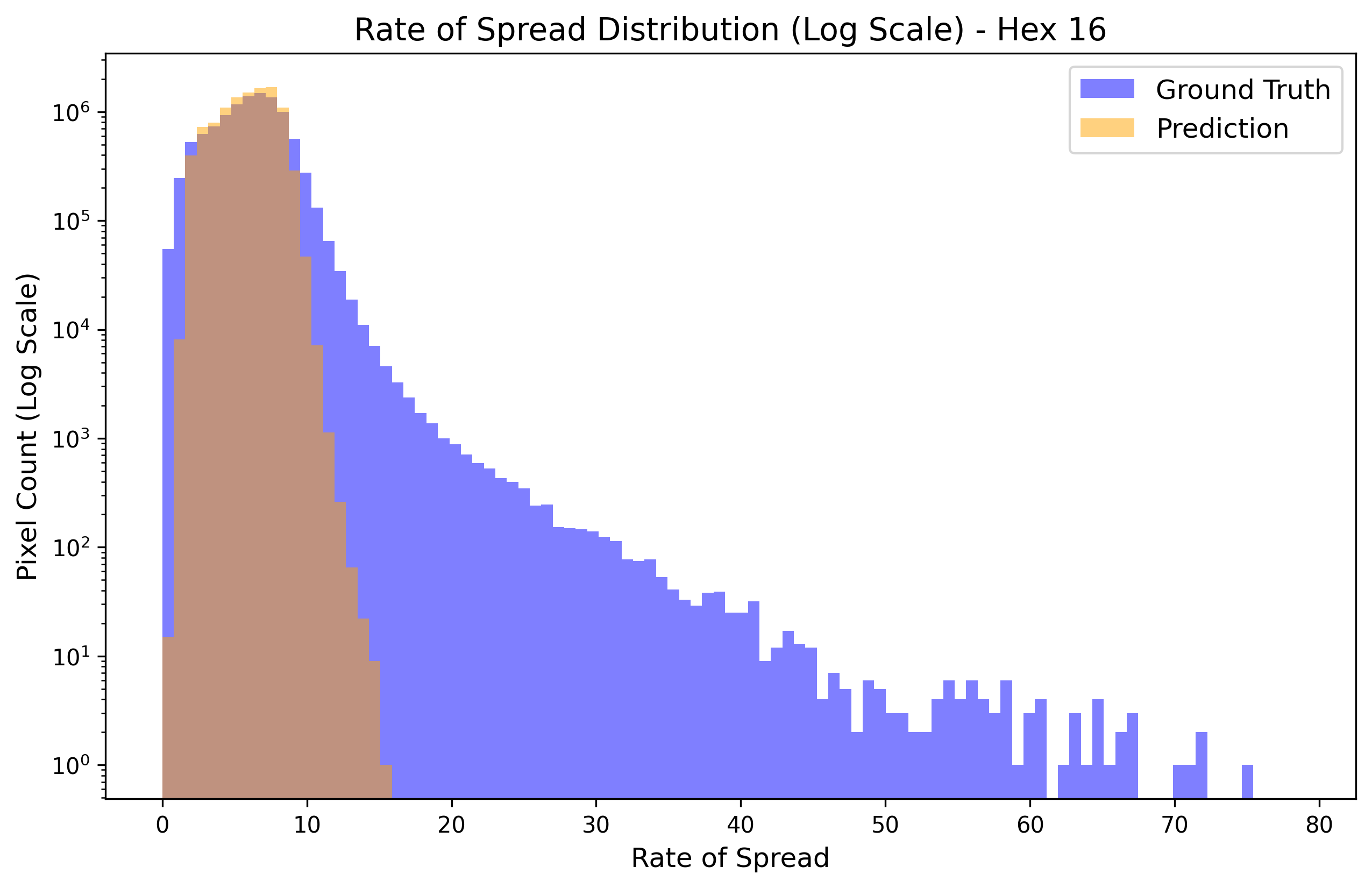}
        \caption{Probability density comparison plot: Rate of spread.}
        \label{fig:hex16_ros_hist}
    \end{subfigure}

    % Row 2: Prediction vs. ground truth
    \begin{subfigure}[t]{0.32\textwidth}
        \centering
        \includegraphics[width=\linewidth,trim={0 0 0 20},clip]{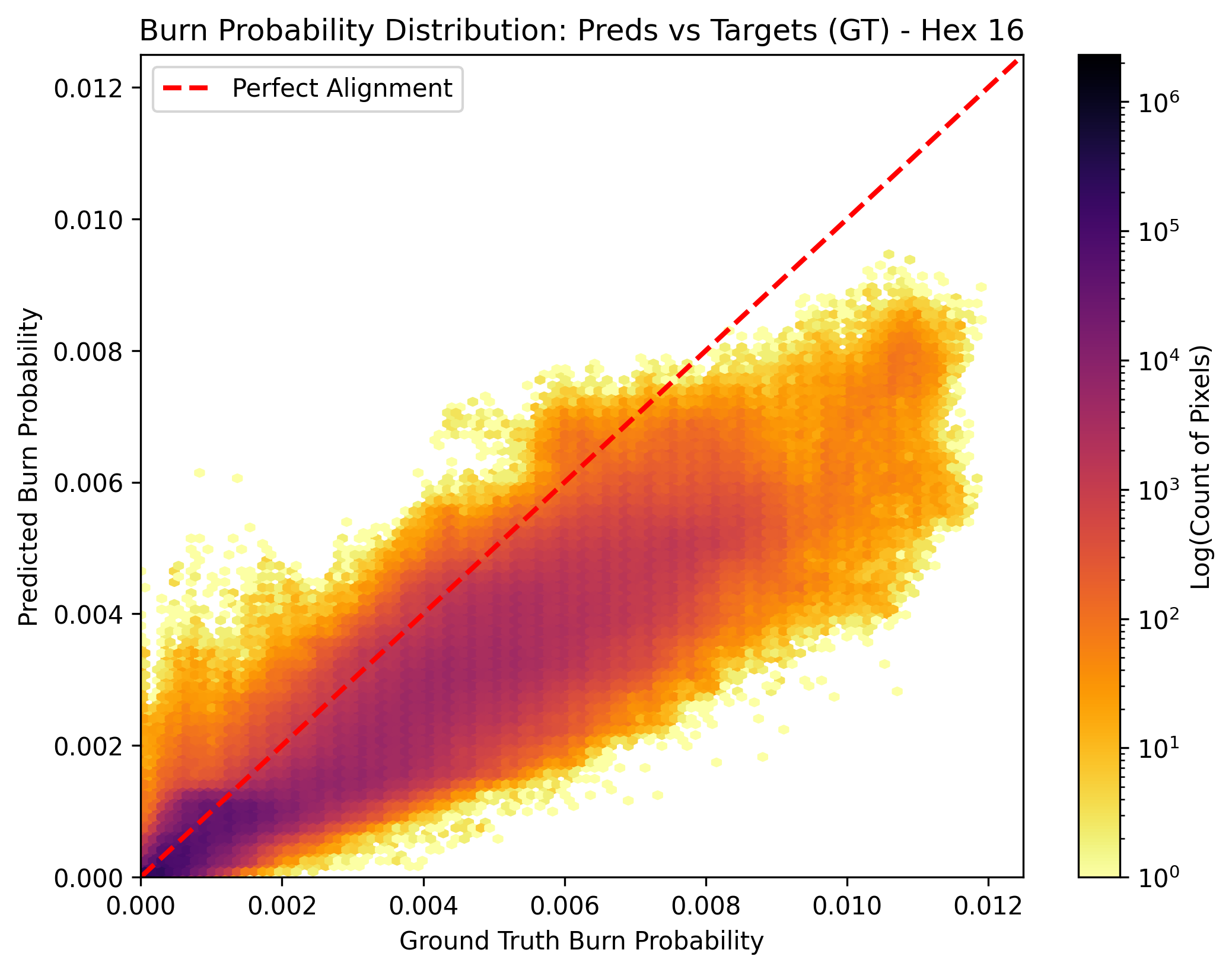}
        \caption{2D Hexbin Calibration Plot: Burn probability}
        \label{fig:hex16_bp_hexbin}
    \end{subfigure}
    \hfill
    \begin{subfigure}[t]{0.32\textwidth}
        \centering
        \includegraphics[width=\linewidth,trim={0 0 0 20},clip]{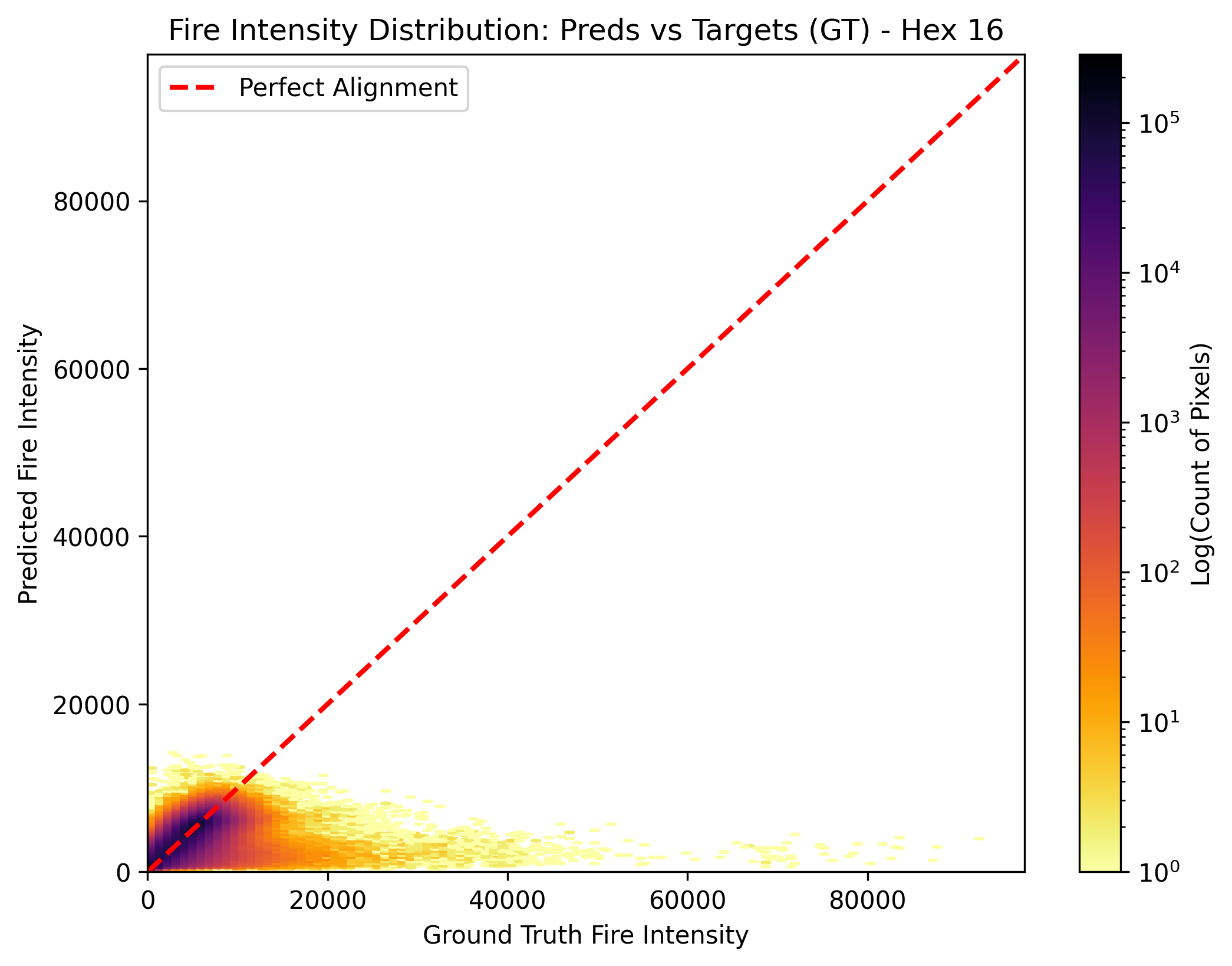}
        \caption{2D Hexbin Calibration Plot: Fire intensity}
        \label{fig:hex16_fi_hexbin}
    \end{subfigure}
    \hfill
    \begin{subfigure}[t]{0.32\textwidth}
        \centering
        \includegraphics[width=\linewidth,trim={0 0 0 20},clip]{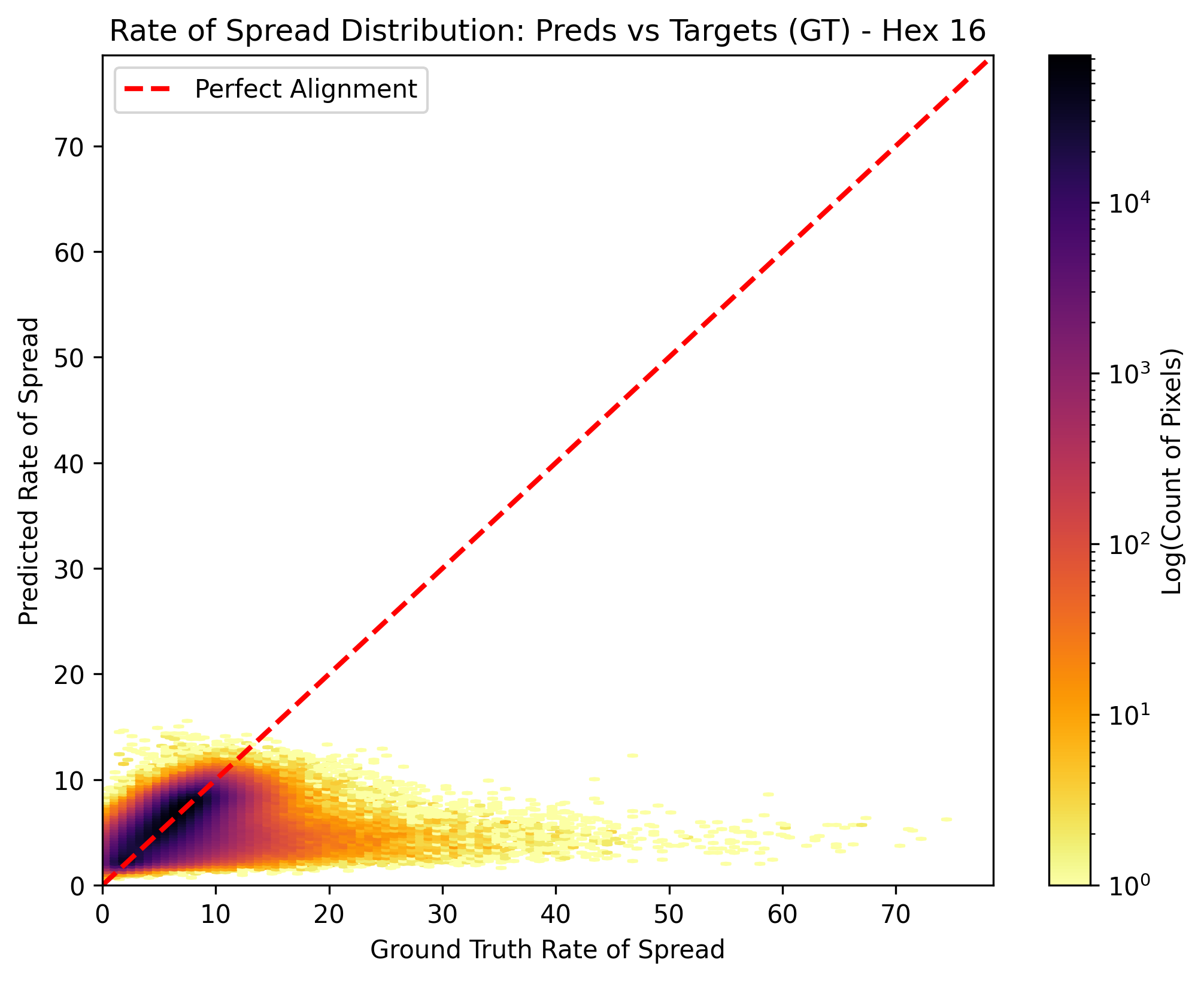}
        \caption{2D Hexbin Calibration Plot: Rate of spread}
        \label{fig:hex16_ros_hexbin}
    \end{subfigure}

    \vspace{0.5em}
    
    \caption{Prediction analysis for test Region 16 for the 3 targets: BP, FI, and ROS. Top row: probability-density comparisons using log-scaled histograms. Bottom row: 2D Hexbin Calibration Plot between AI-surrogate model predictions and BurnP3+ reference maps. }
    \label{fig:region_16_analysis}
\end{figure*}

In this section, we provide results in addition to Table~\ref{tab:main_results_region_level} in the main paper. Table ~\ref{tab:main_results_patch_level} reports patch-level results, which yield the same model ranking as the region-level results. Figure~\ref{fig:region_16_analysis} shows further analysis. For Region 16, burn probability shows the closest agreement with the reference simulation: predictions follow the overall trend and achieve a high CCC, although the highest burn-probability values are systematically underestimated. Fire intensity and rate of spread preserve the general ordering of low-to-moderate values but exhibit stronger regression toward the mean, with predictions concentrated within a much narrower range than the targets. The distribution plots confirm that the model reproduces the dominant low-value modes well but underrepresents the long upper tails, particularly for fire intensity and rate of spread. This explains why correlation remains reasonable while CCC is lower for these two targets: the model captures relative spatial variation better than the full magnitude and variability of extreme values.

Table~\ref{tab:test_regions_results_best_model} reports per-region metrics (of our best U-Net model: Spatial + Weather) for the 5 different test set regions across Canada (Regions 01, 12, 16, 39 and 49) as shown in Fig~\ref{fig:model_design} (a). CCC values above $0.8$ generally indicate strong agreement between predictions and reference simulations. Burn probability achieves CCC values above 0.8 in four of the five regions, exceeding 0.89 in Regions 01, 16, and 49. Fire intensity also shows strong agreement in Region 01, with a CCC of $0.84$, while the remaining regions exhibit moderate-to-strong agreement. Rate of spread exceeds 0.8 only in Region 01, likely because it is a more variable target whose prediction depends closely on the accurate representation of both burn probability and fire-intensity patterns.

\begin{table*}[t]
\centering
\footnotesize 
\setlength{\tabcolsep}{4.5pt} 
\begin{tabular*}{\linewidth}{@{\extracolsep{\fill}} l l r r r r r r }
\toprule
\textbf{Target} & \textbf{Model} & $\downarrow$ \textbf{\shortstack{Normalized \\ MAE}} & $\rightarrow 0$ \textbf{\shortstack{Normalized \\ Bias}} & $\uparrow$ \textbf{Spearman} & $\uparrow$ \textbf{CCC} & $\uparrow$ \textbf{\shortstack{AUC IoU \\ Top-10\%}} & $\uparrow$ \textbf{\shortstack{AUC IoU \\ Full}} \\
\midrule
\multicolumn{8}{c}{\textit{Patch-Level Evaluation}} \\
\midrule
\textbf{BP}  & Mean baseline                 & $9.33$ & $8.70$ & $-$ & $-$ & $0.16$ & $0.60$\\
             & XGBoost                       & $1.24 \pm 0.02$ & $0.55 \pm 0.05$ & $0.55 \pm 0.01$ & $0.28 \pm 0.00$ & $0.12 \pm 0.00$ & $0.58 \pm 0.01$ \\
             & U-Net Spatial                 & $0.52 \pm 0.01$ & $-0.16 \pm 0.07$ & $0.78 \pm 0.00$ & $0.57 \pm 0.02$ & $0.31 \pm 0.00$ & $0.67 \pm 0.00$ \\
             & + Weather                     & $\mathbf{0.36 \pm 0.01}$ & $\mathbf{-0.12 \pm 0.03}$ & $\mathbf{0.80 \pm 0.00}$ & $\mathbf{0.69 \pm 0.00}$ & $0.32 \pm 0.01$ & $0.68 \pm 0.00$ \\
             & + Weather + Fire Size
& $0.36 \pm 0.02$
& $-0.13 \pm 0.03$
& $\mathbf{0.80 \pm 0.00}$
& $0.69\pm 0.01$
& $\mathbf{0.33 \pm 0.00}$
& $\mathbf{0.69\pm 0.00}$ \\
\cmidrule{2-8}
\textbf{FI}  & Mean baseline                 & $0.64$ & $-0.32$ & $-$ & $-$ & $0.10$ & $0.50$ \\
             & XGBoost                       & $0.38 \pm 0.00$ & $-0.11 \pm 0.00$ & $0.58 \pm 0.00$ & $0.50 \pm 0.00$ & $0.13 \pm 0.00$ & $0.56 \pm 0.00$ \\
             & U-Net Spatial                 & $0.31 \pm 0.00$ & $-0.06 \pm 0.02$ & $0.74 \pm 0.00$ & $0.65 \pm 0.01$ & $\mathbf{0.21 \pm 0.00}$ & $0.63 \pm 0.00$ \\
             & + Weather                     & $\mathbf{0.28 \pm 0.00}$ & $-0.04 \pm 0.00$ & $0.74 \pm 0.01$ & $0.68 \pm 0.01$ & $\mathbf{0.21 \pm 0.00}$ & $0.63 \pm 0.00$ \\
             & + Weather + Fire Size
& $\mathbf{ 0.28 \pm 0.00}$
& $\mathbf{ -0.03 \pm 0.00}$
& $\mathbf{0.75 \pm 0.00 }$
& $\mathbf{0.68\pm 0.00 }$
& $\mathbf{0.21 \pm 0.00}$
& $\mathbf{0.64\pm 0.00}$ \\
\cmidrule{2-8}
\textbf{ROS} & Mean baseline                 & $0.44$ & $-0.16$ & $-$ & $-$ & $0.10$ & $0.50$ \\
             & XGBoost                       & $0.30 \pm 0.00$ & $-0.09 \pm 0.00$ & $0.52 \pm 0.00$ & $0.43 \pm 0.00$ & $0.09 \pm 0.00$ & $0.53 \pm 0.00$ \\
             & U-Net Spatial                 & $0.25 \pm 0.00$ & $-0.07 \pm 0.01$ & $0.69 \pm 0.00$ & $0.60 \pm 0.01$ & $0.18 \pm 0.00$ & $0.61 \pm 0.00$ \\
             & + Weather                     & $\mathbf{0.22 \pm 0.00}$ & $\mathbf{-0.04 \pm 0.00}$ & $0.70 \pm 0.01$ & $\mathbf{0.64 \pm 0.01}$ & $0.18 \pm 0.01$ & $\mathbf{0.61 \pm 0.00}$ \\
             & + Weather + Fire Size
& $\mathbf{0.22 \pm 0.00}$
& $-0.04 \pm 0.01$
& $\mathbf{0.70 \pm 0.00}$
& $\mathbf{0.64 \pm 0.01}$
& $\mathbf{0.19 \pm 0.00}$
& $\mathbf{0.61 \pm 0.00 }$ \\
\bottomrule
\end{tabular*}
\caption{Main results comparing the predictive performance of incremental multi-source U-Net models for \textbf{Burn Probability (BP)}, \textbf{Fire intensity (FI)}, and \textbf{Fire Rate of Spread (ROS)}. The baseline spatial architecture is progressively augmented with weather. Each value is reported as the mean and standard deviation of 3 different-seed runs (standard deviations less than $0.01$ are reported as $0.00$). Metrics are reported on the test set at the patch-level. Best performance for each specified metric is highlighted in bold. Note: Correlation metrics are undefined for the mean baseline.}
\label{tab:main_results_patch_level}
\end{table*}

\begin{table*}[t]
\centering
\footnotesize
\setlength{\tabcolsep}{4.5pt}

\begin{tabular*}{\linewidth}{
@{\extracolsep{\fill}}
l l r r r r r r
}
\toprule
\textbf{Target}
& \textbf{Hexel}
& $\downarrow$ \textbf{\shortstack{Normalized \\ MAE}}
& $\rightarrow 0$ \textbf{\shortstack{Normalized \\ Bias}}
& $\uparrow$ \textbf{Spearman}
& $\uparrow$ \textbf{CCC}
& $\uparrow$ \textbf{\shortstack{AUC IoU \\ Top-10\%}}
& $\uparrow$ \textbf{\shortstack{AUC IoU \\ Full}} \\
\midrule

\textbf{BP}
& Region 01
& $0.277 \pm 0.019$
& $-0.134 \pm 0.043$
& $0.951 \pm 0.003$
& $0.894 \pm 0.018$
& $0.417 \pm 0.046$
& $0.806 \pm 0.009$ \\

& Region 12
& $0.378 \pm 0.086$
& $-0.157 \pm 0.157$
& $0.887 \pm 0.001$
& $0.866 \pm 0.069$
& $0.429 \pm 0.145$
& $0.734 \pm 0.021$ \\

& Region 16
& $0.284 \pm 0.057$
& $-0.194 \pm 0.099$
& $0.953 \pm 0.001$
& $0.908 \pm 0.045$
& $0.450 \pm 0.027$
& $0.810 \pm 0.001$ \\

& Region 39
& $0.517 \pm 0.012$
& $-0.005 \pm 0.073$
& $0.819 \pm 0.009$
& $0.683 \pm 0.016$
& $0.165 \pm 0.008$
& $0.639 \pm 0.004$ \\

& Region 49
& $0.279 \pm 0.014$
& $-0.190 \pm 0.033$
& $0.944 \pm 0.005$
& $0.905 \pm 0.011$
& $0.487 \pm 0.008$
& $0.769 \pm 0.007$ \\

\cmidrule{2-8}

\textbf{FI}
& Region 01
& $0.213 \pm 0.003$
& $-0.095 \pm 0.014$
& $0.883 \pm 0.004$
& $0.840 \pm 0.006$
& $0.199 \pm 0.010$
& $0.710 \pm 0.004$ \\

& Region 12
& $0.321 \pm 0.008$
& $0.003 \pm 0.026$
& $0.852 \pm 0.001$
& $0.772 \pm 0.002$
& $0.256 \pm 0.011$
& $0.696 \pm 0.002$ \\

& Region 16
& $0.253 \pm 0.002$
& $-0.031 \pm 0.002$
& $0.812 \pm 0.003$
& $0.761 \pm 0.005$
& $0.180 \pm 0.010$
& $0.657 \pm 0.003$ \\

& Region 39
& $0.295 \pm 0.008$
& $-0.097 \pm 0.025$
& $0.897 \pm 0.004$
& $0.797 \pm 0.008$
& $0.354 \pm 0.009$
& $0.731 \pm 0.005$ \\

& Region 49
& $0.277 \pm 0.008$
& $0.013 \pm 0.030$
& $0.844 \pm 0.002$
& $0.776 \pm 0.008$
& $0.227 \pm 0.010$
& $0.686 \pm 0.002$ \\

\cmidrule{2-8}

\textbf{ROS}
& Region 01
& $0.166 \pm 0.005$
& $-0.069 \pm 0.012$
& $0.878 \pm 0.005$
& $0.856 \pm 0.007$
& $0.249 \pm 0.012$
& $0.712 \pm 0.005$ \\

& Region 12
& $0.274 \pm 0.005$
& $0.016 \pm 0.013$
& $0.810 \pm 0.002$
& $0.745 \pm 0.004$
& $0.293 \pm 0.011$
& $0.676 \pm 0.002$ \\

& Region 16
& $0.199 \pm 0.001$
& $-0.038 \pm 0.010$
& $0.729 \pm 0.005$
& $0.668 \pm 0.003$
& $0.157 \pm 0.007$
& $0.614 \pm 0.003$ \\

& Region 39
& $0.258 \pm 0.005$
& $-0.110 \pm 0.005$
& $0.734 \pm 0.013$
& $0.602 \pm 0.015$
& $0.238 \pm 0.007$
& $0.619 \pm 0.007$ \\

& Region 49
& $0.186 \pm 0.004$
& $0.004 \pm 0.009$
& $0.742 \pm 0.004$
& $0.706 \pm 0.003$
& $0.183 \pm 0.012$
& $0.626 \pm 0.003$ \\

\bottomrule
\end{tabular*}
\caption{Region-level performance across individual test hexels for our best model (Spatial + Weather). Values report the mean $\pm$ standard deviation across three random seeds.}
\label{tab:test_regions_results_best_model}
\end{table*}

\section{Hazard Evaluation Details}
\label{app:hazard_mapping}

In this section, we provide further details on how hazard is defined and computed, as well as additional analysis. For each test region, hazard is computed pixel-wise from the BP and FI rasters via the following four-step procedure:
\begin{enumerate}
    \item \textbf{Intensity Capping:} The average fire intensity is capped at $10{,}000~\mathrm{kW \cdot m^{-1}}$ to mitigate the influence of extreme outliers:
    \[
    \mathrm{FI}_{\mathrm{cap}}(x) = \min\left(\mathrm{FI}(x),\, 10{,}000\right),
    \]
    where $x$ indexes pixels.
    
    \item \textbf{Raw Hazard Calculation:} Raw hazard is computed as the product of likelihood and capped intensity:
    \[ 
    H_{\mathrm{raw}}(x) = \mathrm{BP}(x) \cdot \mathrm{FI}_{\mathrm{cap}}(x). 
    \]

    \item \textbf{Linear Rescaling:} The raw hazard map is linearly rescaled to a $[0, 100]$ range using the maximum value of reference BurnP3+ outputs:
    \[
    H_{\mathrm{scaled}}(x) = 100 \times \frac{H_{\mathrm{raw}}(x)}{\max_{x'} H_{\mathrm{raw}}(x')}. 
    \]

    \item \textbf{Discretization:} Continuous hazard values are mapped into 13 ordinal hazard classes with thresholds spanning from $<0.01$ to $\geq 50$. The complete bin intervals are provided below.
\end{enumerate}

Table~\ref{tab:hazard_bins} lists the 13 ordinal hazard classes used to discretize the scaled hazard index \(H_{\mathrm{scaled}}\). 

\begin{table}[htbp]
\centering
\begin{tabular}{cc}
\toprule
\textbf{Hazard class} & \(\boldsymbol{H_{\mathrm{scaled}}}\) \textbf{interval} \\
\midrule
1  & \(H_{\mathrm{scaled}} < 0.01\) \\
2  & \(0.01 \leq H_{\mathrm{scaled}} < 0.025\) \\
3  & \(0.025 \leq H_{\mathrm{scaled}} < 0.05\) \\
4  & \(0.05 \leq H_{\mathrm{scaled}} < 0.1\) \\
5  & \(0.1 \leq H_{\mathrm{scaled}} < 0.25\) \\
6  & \(0.25 \leq H_{\mathrm{scaled}} < 0.5\) \\
7  & \(0.5 \leq H_{\mathrm{scaled}} < 1\) \\
8  & \(1 \leq H_{\mathrm{scaled}} < 2.5\) \\
9  & \(2.5 \leq H_{\mathrm{scaled}} < 5\) \\
10 & \(5 \leq H_{\mathrm{scaled}} < 10\) \\
11 & \(10 \leq H_{\mathrm{scaled}} < 25\) \\
12 & \(25 \leq H_{\mathrm{scaled}} < 50\) \\
13 & \(H_{\mathrm{scaled}} \geq 50\) \\
\bottomrule
\end{tabular}
\caption{NRCan hazard bin definitions used to discretize the scaled hazard index \(H_{\mathrm{scaled}}\).}
\label{tab:hazard_bins}
\end{table}

\begin{figure}[t]
    \centering
    \includegraphics[width=0.85\columnwidth,trim={0 20pt 0 29pt},clip]
    {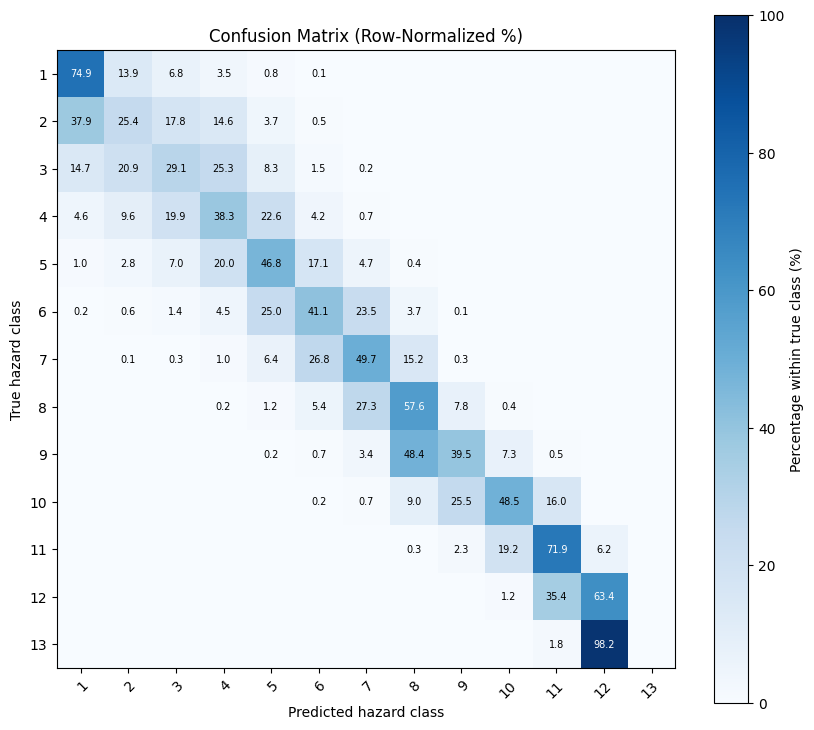}
    \caption{Row-normalized confusion matrix for BurnP3+ reference and surrogate-predicted hazard classes.}
    \label{fig:hazard_confusion_matrix}
\end{figure}

The confusion matrix in Figure~\ref{fig:hazard_confusion_matrix} shows that classification errors become increasingly concentrated near the diagonal at moderate-to-high hazard levels, indicating that errors in these classes generally correspond to small ordinal shifts. 
Agreement is weaker among the lower hazard classes, where the narrow bin intervals make the classifications more sensitive to small differences in the underlying continuous hazard values.

For the critical high-hazard regime (classes 10--13), the surrogate achieves an Intersection-over-Union (IoU) of approximately \(85\%\), indicating strong spatial agreement with BurnP3+ in delineating elevated-hazard areas. 
At the most extreme level, however, the IoU for class 13 is \(0\). 
Because the hazard index is scaled by the maximum BurnP3+ raw hazard, a small number of exceptionally high reference values raise the effective raw-hazard threshold for this class. 
The surrogate underestimates these extreme peaks, causing the corresponding pixels to fall into lower hazard classes after scaling. 
Thus, the zero IoU reflects a failure to reproduce the most extreme hazard magnitudes rather than a broader failure to identify high-hazard areas.

%For the maximal hazard class (class 13), it achieves an IoU of 71\%, indicating that the surrogate also delineates the most extreme hazard regions with substantial agreement.
%These results show that the emulator largely preserves downstream, decision-relevant hazard patterns, although residual errors in the upper hazard classes remain important given their operational significance.

%\section{Counterfactual Scenario Details}
%\label{app:counterfactuals}

%\subsection{Fuel Intervention Masks}
%\subsection{Full Counterfactual Maps}
%\subsection{Pixelwise Change Distributions}
%\subsection{Distance-to-Barrier Profiles}

\section{Additional Experiments} \label{app:additional_experiments}
In this section, we present additional experiments comparing different weather encoders and three single-task U-Net models with the multi-task U-Net.

\paragraph{Architecture Ablation.}
In addition to representing weather variables as spatially projected channels, as described in the main paper, we investigate an alternative tabular encoding and bottleneck-fusion strategy. This formulation more closely resembles the stochastic weather sampling procedure used in BurnP3+, as the encoder operates on a sampled subset of the available weather records. Specifically, the tabular encoder processes \(n=256\) sampled weather rows using a three-layer MLP composed of linear transformations, batch normalization, and ReLU activations. The resulting row-wise representations are aggregated through max pooling and projected to a 128-dimensional embedding. This embedding is then broadcast to the spatial resolution of the U-Net bottleneck and concatenated with the bottleneck feature map and spatial wind-grid embeddings. A \(1\times1\) convolution followed by batch normalization and ReLU fuses the combined representation while restoring the original bottleneck channel dimension.

Table~\ref{tab:results_region_level_tabular_spatial_ablation} compares the spatial and tabular weather encoders at the region level. Overall, spatial weather encoding achieves stronger performance across most metrics and all three prediction targets. The tabular encoder produces biases closer to zero for BP and FI and a slightly higher CCC for BP, but underperforms on the remaining correlation, error, and spatial-overlap metrics. These results suggest that explicitly preserving the spatial structure of weather variables is generally more effective than compressing sampled weather records into a single global embedding.

\begin{table*}[htp]
\centering
\footnotesize
\setlength{\tabcolsep}{4.5pt}
\begin{tabular*}{\linewidth}{
@{\extracolsep{\fill}}
l l r r r r r r
}
\toprule
\textbf{Target}
& \textbf{Model}
& $\downarrow$ \textbf{\shortstack{Normalized \\ MAE}}
& $\rightarrow 0$ \textbf{\shortstack{Normalized \\ Bias}}
& $\uparrow$ \textbf{Spearman}
& $\uparrow$ \textbf{CCC}
& $\uparrow$ \textbf{\shortstack{AUC IoU \\ Top-10\%}}
& $\uparrow$ \textbf{\shortstack{AUC IoU \\ Full}} \\
\midrule

\textbf{BP}
& U-Net Spatial
& $0.496$
& $-0.199$
& $0.878$
& $0.674$
& $0.239$
& $0.692$ \\

& + Weather (Spatial)
& $\mathbf{0.345}$
& $-0.176$
& $\mathbf{0.915}$
& $0.847$
& $\mathbf{0.375}$
& $\mathbf{0.754}$ \\

& + Weather (Tabular)
& $0.371$
& $\mathbf{-0.025}$
& $0.905$
& $\mathbf{0.851}$
& $0.362$
& $0.739$ \\

\cmidrule{2-8}

\textbf{FI}
& U-Net Spatial
& $0.301$
& $-0.057$
& $0.854$
& $0.768$
& $0.237$
& $0.691$ \\

& + Weather (Spatial)
& $\mathbf{0.272}$
& $-0.048$
& $\mathbf{0.859}$
& $\mathbf{0.788}$
& $\mathbf{0.246}$
& $\mathbf{0.697}$ \\

& + Weather (Tabular)
& $0.290$
& $\mathbf{-0.021}$
& $0.853$
& $0.779$
& $0.223$
& $0.688$ \\

\cmidrule{2-8}

\textbf{ROS}
& U-Net Spatial
& $0.242$
& $-0.073$
& $0.773$
& $0.686$
& $0.210$
& $0.643$ \\

& + Weather (Spatial)
& $\mathbf{0.216}$
& $\mathbf{-0.044}$
& $\mathbf{0.782}$
& $\mathbf{0.717}$
& $\mathbf{0.225}$
& $\mathbf{0.652}$ \\

& + Weather (Tabular)
& $0.231$
& $-0.050$
& $0.778$
& $0.702$
& $0.209$
& $0.647$ \\

\bottomrule
\end{tabular*}
\caption{Region-level comparison of multi-task U-Net models using spatial and tabular weather encoders for \textbf{Burn Probability (BP)}, \textbf{Fire Intensity (FI)}, and \textbf{Rate of Spread (ROS)}. The U-Net Spatial model does not include weather variables. Best performance for each target and metric is highlighted in bold, with normalized bias evaluated according to proximity to zero.}
\label{tab:results_region_level_tabular_spatial_ablation}
\end{table*}

\paragraph{Single-Task vs. Multi-Task Models}
We train three separate single-task models, each specialized for one target, and compare them with a single multi-task model that jointly predicts all three targets. Both configurations use the same backbone, inputs, and training settings. Results are shown in Table~\ref{tab:results_region_level_single_task_vs_multi_task_ablation}. For BP, multi-task learning outperforms the single-task models for most of the metrics. In contrast, the single-task models consistently outperform the multi-task model across all metrics for FI and ROS. These results suggest that joint learning benefits BP through shared representations, but introduces negative transfer for the two fire-behaviour targets. Thus, the multi-task formulation provides a more computationally efficient unified model, while separate task-specific models offer stronger predictive performance for FI and ROS. Given the computational and deployment advantages of shared representations, future work will further investigate improved multi-task learning strategies to reduce negative transfer and match or exceed the performance of task-specific models.

\begin{table*}[htp]
\centering
\footnotesize
\setlength{\tabcolsep}{4.5pt}
\begin{tabular*}{\linewidth}{
@{\extracolsep{\fill}}
l l r r r r r r
}
\toprule
\textbf{Target}
& \textbf{Model}
& $\downarrow$ \textbf{\shortstack{Normalized \\ MAE}}
& $\rightarrow 0$ \textbf{\shortstack{Normalized \\ Bias}}
& $\uparrow$ \textbf{Spearman}
& $\uparrow$ \textbf{CCC}
& $\uparrow$ \textbf{\shortstack{AUC IoU \\ Top-10\%}}
& $\uparrow$ \textbf{\shortstack{AUC IoU \\ Full}} \\
\midrule

\textbf{BP}
& Single-task
& $0.404$
& $\mathbf{-0.137}$
& $0.893$
& $0.783$
& $\mathbf{0.427}$
& $0.744$ \\

& Multi-task
& $\mathbf{0.345}$
& $-0.176$
& $\mathbf{0.915}$
& $\mathbf{0.847}$
& $0.375$
& $\mathbf{0.754}$ \\

\cmidrule{2-8}

\textbf{FI}
& Single-task
& $\mathbf{0.259}$
& $\mathbf{0.009}$
& $\mathbf{0.873}$
& $\mathbf{0.814}$
& $\mathbf{0.277}$
& $\mathbf{0.712}$ \\

& Multi-task
& $0.272$
& $-0.048$
& $0.859$
& $0.788$
& $0.246$
& $0.697$ \\

\cmidrule{2-8}

\textbf{ROS}
& Single-task
& $\mathbf{0.204}$
& $\mathbf{-0.026}$
& $\mathbf{0.799}$
& $\mathbf{0.747}$
& $\mathbf{0.248}$
& $\mathbf{0.665}$ \\

& Multi-task
& $0.216$
& $-0.044$
& $0.782$
& $0.717$
& $0.225$
& $0.652$ \\

\bottomrule
\end{tabular*}
\caption{Region-level comparison between a multi-task U-Net that jointly predicts all three targets and three single-task U-Nets trained separately for \textbf{Burn Probability (BP)}, \textbf{Fire Intensity (FI)}, and \textbf{Rate of Spread (ROS)}. Both configurations use identical inputs, architecture, and training settings. Best performance for each target and metric is highlighted in bold.}
\label{tab:results_region_level_single_task_vs_multi_task_ablation}
\end{table*}

\section{Operationally Representative Post-Fire Fuel Intervention} \label{app:realistic_d2}
\begin{figure}[t]
\centering
\includegraphics[width=0.75\linewidth]{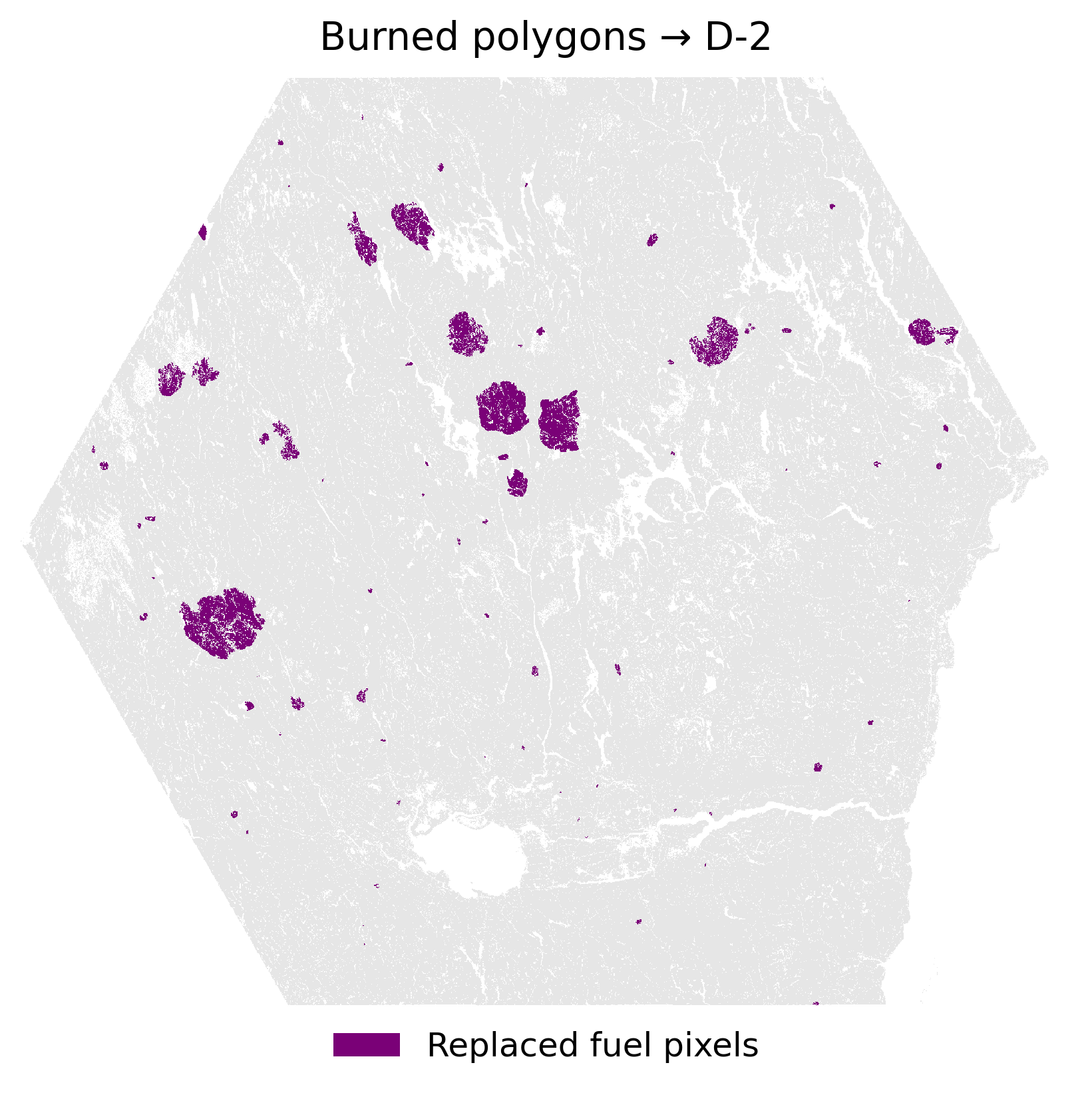}
\caption{Post-fire D2 intervention in region 16, showing fuels replaced within selected BurnP3+ fire polygons.}
\label{fig:realistic_d2_mask}
\end{figure}

\begin{figure*}[p]
\centering
\includegraphics[width=\textwidth]{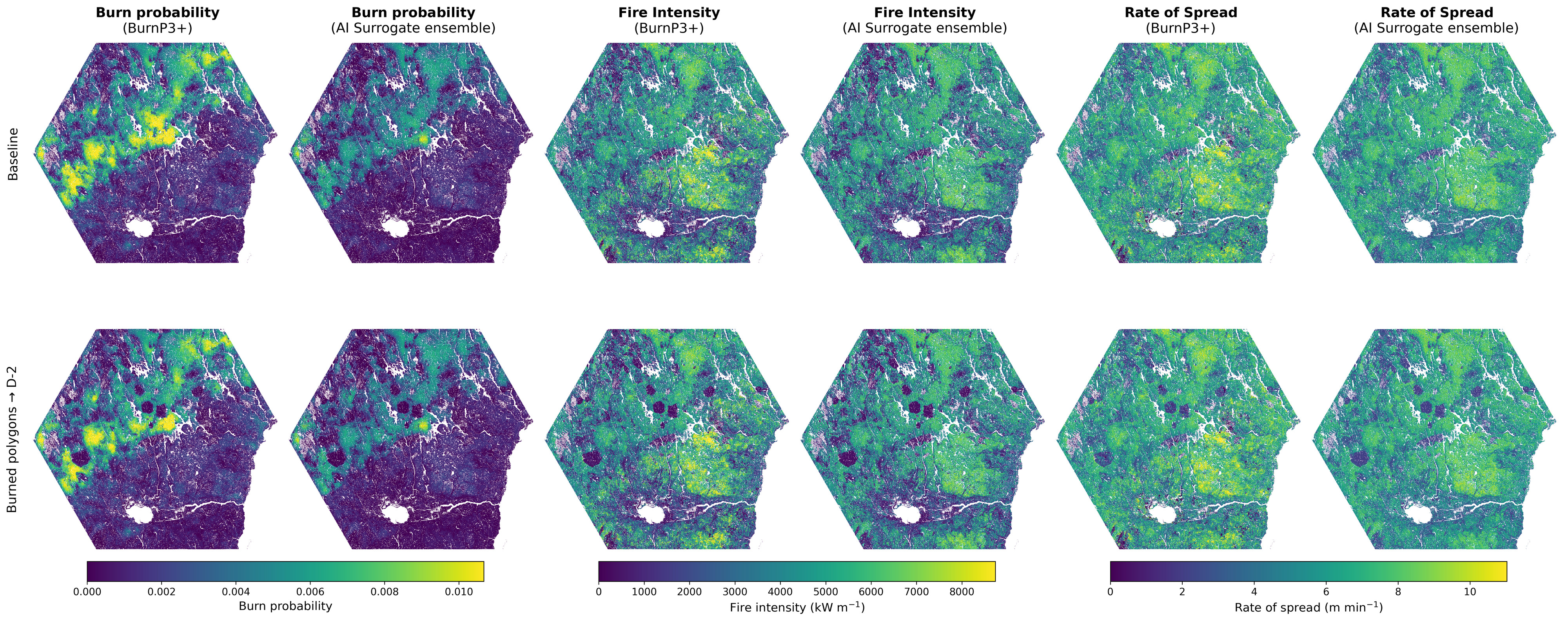}
\caption{BurnP3+ and surrogate outputs under the baseline and post-fire D2 intervention in region 16.}
\label{fig:realistic_d2_fuel_baseline_counterfactual}
\end{figure*}

\begin{figure*}[p]
\centering
\includegraphics[width=0.85\textwidth]{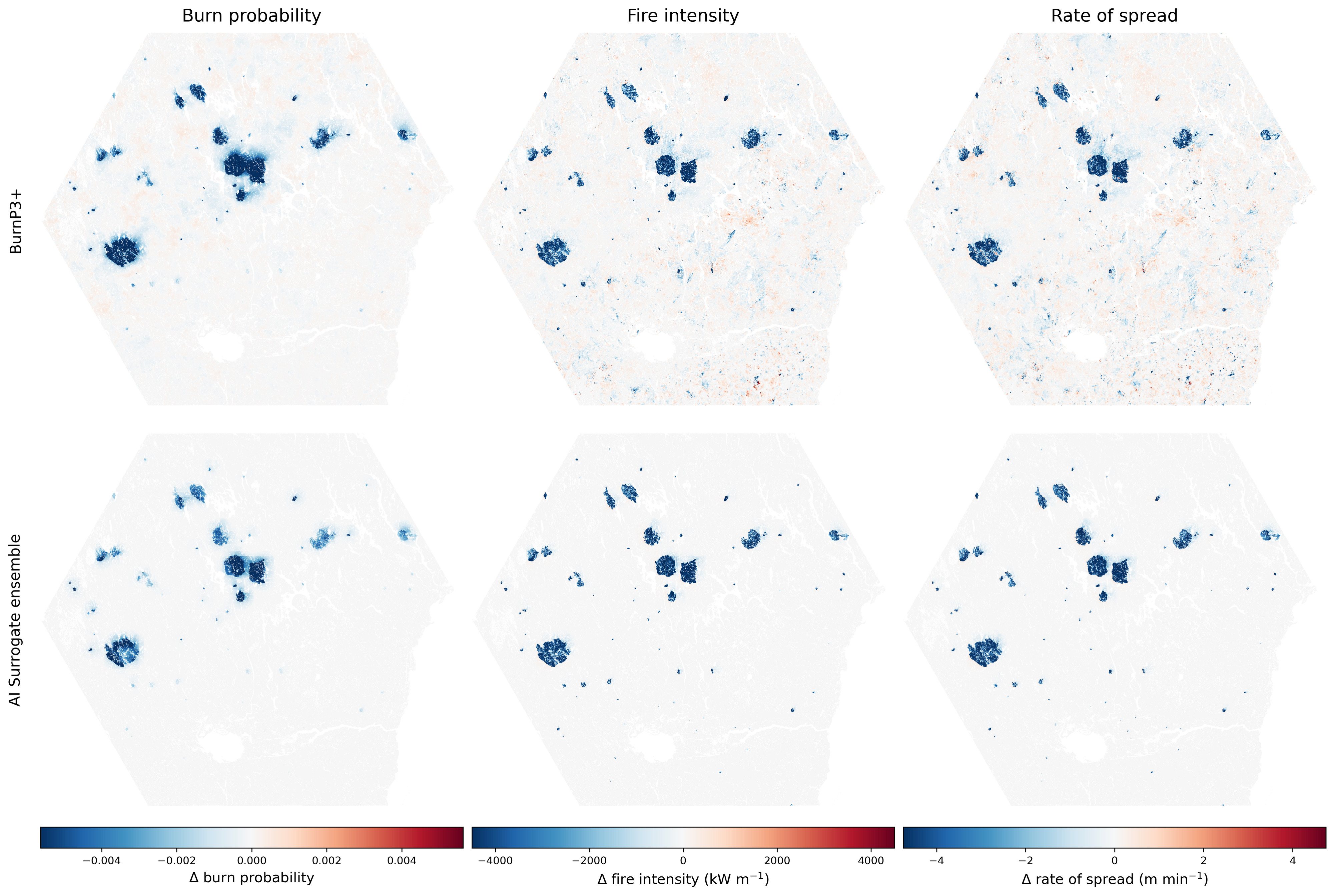}
\caption{Counterfactual-minus-baseline changes in BP, FI, and ROS for BurnP3+ and the surrogate under the post-fire D2 intervention.}
\label{fig:realistic_d2_fuel_bp3_validation}
\end{figure*}

To complement the large-scale fuel interventions considered in Section~\ref{fuel_intervention}, we evaluate a more operationally representative post-fire modification of the region 16 fuel landscape. 
Rather than combining fire polygons across a large number of BurnP3+ realizations, we select the fire polygons from a single relatively severe simulated fire-season realization, corresponding to approximately 84 fires and affecting 243,100~ha, or 1.45\% of the region. This produces a localized landscape modification more representative of the fuel changes that may occur following one to two fire seasons. 
Figure~\ref{fig:realistic_d2_mask} shows the spatial extent of the modified areas.

As in the main post-fire intervention, fuels within the selected fire polygons are replaced with D2 (Green Aspen). 
All other inputs are held at their baseline values, and the modified fuel layer is supplied independently to BurnP3+ and the surrogate.

Figure~\ref{fig:realistic_d2_fuel_baseline_counterfactual} compares the BP, FI, and ROS outputs produced by BurnP3+ and the surrogate under the baseline and modified fuel conditions. 
We assess the counterfactual response by subtracting each model's baseline output from its corresponding intervention output. 
Figure~\ref{fig:realistic_d2_fuel_bp3_validation} compares the resulting signed change maps.

The surrogate predicts region-wide mean changes in BP, FI, and ROS of $-7.745\%\pm0.078$, $-2.842\%\pm0.037$, and $-1.979\%\pm0.048$, respectively. 
The corresponding BurnP3+ changes are $-7.803\%$, $-3.058\%$, and $-2.076\%$. 
The close agreement in these region-wide mean responses indicates that the surrogate reproduces the overall magnitude and direction of the intervention effect. 
Spatially, the changes are concentrated within and around the modified fuel areas, with comparatively limited response elsewhere in the landscape.

As in Section~\ref{fuel_intervention}, we quantify agreement between the surrogate and BurnP3+ signed change maps using CCC, Spearman's rank correlation \((\rho_s)\), and NMAE. The agreement scores \((\mathrm{CCC}, \rho_s, \mathrm{NMAE})\) are \((0.865, 0.263, 0.586)\) for BP, \((0.871, 0.125, 0.691)\) for FI, and \((0.811, 0.133, 0.749)\) for ROS.

The comparatively low region-wide Spearman correlations reflect the localized nature of the intervention: because only 1.45\% of the landscape is directly modified, most locations exhibit very small counterfactual changes, making their relative ranking sensitive to small differences between the surrogate and BurnP3+. 
Agreement is substantially stronger in the vicinity of the intervention. Within the edited areas, Spearman correlations are 0.87, 0.82, and 0.67 for BP, FI, and ROS, respectively, and remain 0.78, 0.57, and 0.57 within 2~km of the modified fuels. 
The correlations decrease with distance as the intervention signal becomes progressively weaker.

Overall, the surrogate closely reproduces the region-wide mean response of BurnP3+ and captures the localized spatial response around the modified fuels. 
These results complement the larger-scale fuel perturbations presented in the main analysis by showing that the surrogate remains responsive when fuel changes are confined to a small and operationally plausible fraction of the landscape.

\paragraph{Alternative Architectures}
We further compare U-Net with ResNet-based and Swin-UNet architectures using the same inputs and model configuration, with results reported in Table~\ref{tab:architecture_comparison}. All three models achieve comparable performance for BP, while U-Net consistently outperforms the other architectures across all metrics for FI and ROS. These results support our choice of U-Net, as its simpler architecture provides competitive or better performance without additional complexity.

\begin{table*}[htp]
\centering
\footnotesize
\setlength{\tabcolsep}{4.5pt}
\begin{tabular*}{\linewidth}{
@{\extracolsep{\fill}}
l l r r r r r r
}
\toprule
\textbf{Target}
& \textbf{Model}
& $\downarrow$ \textbf{\shortstack{Normalized \\ MAE}}
& $\rightarrow 0$ \textbf{\shortstack{Normalized \\ Bias}}
& $\uparrow$ \textbf{Spearman}
& $\uparrow$ \textbf{CCC}
& $\uparrow$ \textbf{\shortstack{AUC IoU \\ Top-10\%}}
& $\uparrow$ \textbf{\shortstack{AUC IoU \\ Full}} \\
\midrule

\textbf{BP}
& U-Net
& $0.3445$
& $-0.1762$
& $\mathbf{0.9150}$
& $0.8468$
& $0.3753$
& $\mathbf{0.7544}$ \\

& ResNet
& $\mathbf{0.3284}$
& $-0.0969$
& $0.9036$
& $0.8519$
& $\mathbf{0.4286}$
& $0.7522$ \\

& Swin-UNet
& $0.3598$
& $\mathbf{-0.0174}$
& $0.8943$
& $\mathbf{0.8596}$
& $0.4230$
& $0.7454$ \\

\cmidrule{2-8}

\textbf{FI}
& U-Net
& $\mathbf{0.2717}$
& $\mathbf{-0.0485}$
& $\mathbf{0.8589}$
& $\mathbf{0.7881}$
& $\mathbf{0.2461}$
& $\mathbf{0.6969}$ \\

& ResNet
& $0.2840$
& $-0.0550$
& $0.8406$
& $0.7670$
& $0.2215$
& $0.6815$ \\

& Swin-UNet
& $0.3083$
& $-0.1041$
& $0.8180$
& $0.7053$
& $0.1904$
& $0.6638$ \\

\cmidrule{2-8}

\textbf{ROS}
& U-Net
& $\mathbf{0.2160}$
& $\mathbf{-0.0439}$
& $\mathbf{0.7821}$
& $\mathbf{0.7169}$
& $\mathbf{0.2253}$
& $\mathbf{0.6515}$ \\

& ResNet
& $0.2251$
& $-0.0503$
& $0.7570$
& $0.6910$
& $0.2088$
& $0.6359$ \\

& Swin-UNet
& $0.2335$
& $-0.0585$
& $0.7411$
& $0.6494$
& $0.1691$
& $0.6226$ \\

\bottomrule
\end{tabular*}
\caption{Region-level comparison of network architectures for the multi-task model predicting \textbf{Burn Probability (BP)}, \textbf{Fire Intensity (FI)}, and \textbf{Rate of Spread (ROS)}, using spatial and weather inputs (seed 42). We compare the standard U-Net with ResNet-based and Swin-based U-shaped architectures that share the same fuel embedding, auxiliary fusion, and output heads. Best performance for each target and metric is highlighted in bold.}
\label{tab:architecture_comparison}
\end{table*}

\paragraph{Fuel Representation Ablation.} 
We compare the proposed fuel representation with one-hot encoding, with results reported in Table~\ref{tab:fuel_representation_ablation}. Both approaches achieve comparable performance for burn probability (BP), while the proposed representation achieves slightly better results for fire intensity (FI) and rate of spread (ROS). More importantly, our representation encodes fuel behavior using physically informed iROS curves rather than fixed categorical identifiers, allowing the model to handle previously unseen fuel classes. We further demonstrate this flexibility in our regional transfer experiments, which include four fuel classes not present during national training.

\begin{table*}[htp]
\centering
\footnotesize
\setlength{\tabcolsep}{4.5pt}
\begin{tabular*}{\linewidth}{
@{\extracolsep{\fill}}
l l r r r r r r
}
\toprule
\textbf{Target}
& \textbf{Model}
& $\downarrow$ \textbf{\shortstack{Normalized \\ MAE}}
& $\rightarrow 0$ \textbf{\shortstack{Normalized \\ Bias}}
& $\uparrow$ \textbf{Spearman}
& $\uparrow$ \textbf{CCC}
& $\uparrow$ \textbf{\shortstack{AUC IoU \\ Top-10\%}}
& $\uparrow$ \textbf{\shortstack{AUC IoU \\ Full}} \\
\midrule

\textbf{BP}
& iROS-features
& $0.3445$
& $-0.1762$
& $\mathbf{0.9150}$
& $0.8468$
& $0.3753$
& $\mathbf{0.7544}$ \\

& One-Hot Encoding
& $\mathbf{0.3385}$
& $\mathbf{-0.0267}$
& $0.9023$
& $\mathbf{0.8584}$
& $\mathbf{0.4134}$
& $0.7465$ \\

\cmidrule{2-8}

\textbf{FI}
& iROS-features
& $\mathbf{0.2717}$
& $-0.0485$
& $\mathbf{0.8589}$
& $0.7881$
& $0.2461$
& $\mathbf{0.6969}$ \\

& One-Hot Encoding
& $0.2801$
& $\mathbf{-0.0160}$
& $0.8558$
& $\mathbf{0.7890}$
& $\mathbf{0.2469}$
& $0.6945$ \\

\cmidrule{2-8}

\textbf{ROS}
& iROS-features
& $\mathbf{0.2160}$
& $-0.0439$
& $\mathbf{0.7821}$
& $\mathbf{0.7169}$
& $\mathbf{0.2253}$
& $\mathbf{0.6515}$ \\

& One-Hot Encoding
& $0.2254$
& $\mathbf{-0.0334}$
& $0.7679$
& $0.7034$
& $0.2123$
& $0.6416$ \\

\bottomrule
\end{tabular*}
\caption{Region-level ablation study of input feature encoding for the multi-task U-Net predicting \textbf{Burn Probability (BP)}, \textbf{Fire Intensity (FI)}, and \textbf{Rate of Spread (ROS)}. We compare spatial and weather inputs with two different fuel representations: FBP-derived features and one-hot encoding. Best performance for each target and metric is highlighted in bold, including ties.}
\label{tab:fuel_representation_ablation}
\end{table*}

\section{Cross-Study Model Transfer Additional Results} \label{app:nwt_data}
We present the cross-study evaluation across all scenarios for the Spatial + Weather + Fire Size model in Table~\ref{app:tab:NWT_data_weather_firesize_model}. As described in the main text, the performance trends align closely between the two model configurations.

\begin{table*}[t]
\centering
\footnotesize
\setlength{\tabcolsep}{4.5pt}
\begin{tabular*}{\linewidth}{
@{\extracolsep{\fill}}
l l r r r r r r
}
\toprule
\textbf{Target}
& \textbf{Model}
& $\downarrow$ \textbf{\shortstack{Normalized \\ MAE}}
& $\rightarrow 0$ \textbf{\shortstack{Normalized \\ Bias}}
& $\uparrow$ \textbf{Spearman}
& $\uparrow$ \textbf{CCC}
& $\uparrow$ \textbf{\shortstack{AUC IoU \\ Top-10\%}}
& $\uparrow$ \textbf{\shortstack{AUC IoU \\ Full}} \\
\midrule
\multicolumn{8}{c}{\textit{\textbf{Scenario A}}} \\
\midrule

\textbf{BP}
& U-Net (Base)
& $2.605 \pm 1.108$
& $2.517 \pm 1.161$
& $0.543 \pm 0.077$
& $0.121 \pm 0.026$
& $0.085 \pm 0.018$
& $0.524 \pm 0.028$ \\

& + Recalibration
& $0.666 \pm 0.111$
& $-0.645 \pm 0.117$
& $0.544 \pm 0.076$
& $0.182 \pm 0.122$
& $0.085 \pm 0.018$
& $0.524 \pm 0.028$ \\

& + Zone Alignment
& $\mathbf{0.652} \pm 0.122$
& $\mathbf{-0.631} \pm 0.128$
& $\mathbf{0.588} \pm 0.094$
& $\mathbf{0.209} \pm 0.148$
& $\mathbf{0.085} \pm 0.017$
& $\mathbf{0.538} \pm 0.034$ \\

\cmidrule{2-8}

\textbf{FI}
& U-Net (Base)
& $\mathbf{0.285} \pm 0.005$
& $\mathbf{-0.095} \pm 0.041$
& $\mathbf{0.809} \pm 0.005$
& $0.730 \pm 0.011$
& $\mathbf{0.143} \pm 0.011$
& $\mathbf{0.649} \pm 0.003$ \\

& + Recalibration
& $0.330 \pm 0.023$
& $0.158 \pm 0.050$
& $\mathbf{0.809} \pm 0.005$
& $\mathbf{0.747} \pm 0.019$
& $\mathbf{0.143} \pm 0.011$
& $\mathbf{0.649} \pm 0.003$ \\

& + Zone Alignment
& $0.330 \pm 0.025$
& $0.155 \pm 0.056$
& $\mathbf{0.809} \pm 0.004$
& $0.745 \pm 0.020$
& $0.133 \pm 0.012$
& $0.648 \pm 0.001$ \\

\cmidrule{2-8}

\textbf{ROS}
& U-Net (Base)
& $\mathbf{0.231} \pm 0.008$
& $-0.038 \pm 0.062$
& $\mathbf{0.784} \pm 0.008$
& $\mathbf{0.710} \pm 0.020$
& $\mathbf{0.133} \pm 0.014$
& $\mathbf{0.634} \pm 0.006$ \\

& + Recalibration
& $0.232 \pm 0.013$
& $0.004 \pm 0.060$
& $\mathbf{0.784} \pm 0.008$
& $0.708 \pm 0.023$
& $\mathbf{0.133} \pm 0.014$
& $\mathbf{0.634} \pm 0.006$ \\

& + Zone Alignment
& $0.234 \pm 0.011$
& $\mathbf{-0.001} \pm 0.069$
& $\mathbf{0.784} \pm 0.004$
& $0.703 \pm 0.018$
& $0.119 \pm 0.003$
& $0.633 \pm 0.002$ \\

\midrule
\multicolumn{8}{c}{\textit{\textbf{Scenario B}}} \\
\midrule

% BP
\textbf{BP}
& U-Net (Base)
& $2.967 \pm 1.189$
& $2.863 \pm 1.250$
& $0.522 \pm 0.074$
& $0.101 \pm 0.020$
& $0.050 \pm 0.008$
& $0.515 \pm 0.022$ \\

& + Fine-tuning
& $\mathbf{0.273} \pm 0.023$
& $\mathbf{0.076} \pm 0.014$
& $\mathbf{0.901} \pm 0.009$
& $\mathbf{0.876} \pm 0.014$
& $\mathbf{0.332} \pm 0.007$
& $\mathbf{0.732} \pm 0.005$ \\

\cmidrule{2-8}

% FI
\textbf{FI}
& U-Net (Base)
& $0.297 \pm 0.006$
& $\mathbf{-0.086} \pm 0.042$
& $0.803 \pm 0.007$
& $0.719 \pm 0.012$
& $0.142 \pm 0.015$
& $0.645 \pm 0.004$ \\

& + Fine-tuning
& $\mathbf{0.290} \pm 0.007$
& $-0.138 \pm 0.030$
& $\mathbf{0.813} \pm 0.001$
& $\mathbf{0.723} \pm 0.013$
& $\mathbf{0.167} \pm 0.007$
& $\mathbf{0.657} \pm 0.001$ \\

\cmidrule{2-8}

% ROS
\textbf{ROS}
& U-Net (Base)
& $0.241 \pm 0.010$
& $\mathbf{-0.027} \pm 0.062$
& $0.777 \pm 0.011$
& $0.697 \pm 0.021$
& $0.130 \pm 0.019$
& $0.630 \pm 0.009$ \\

& + Fine-tuning
& $\mathbf{0.218} \pm 0.001$
& $-0.039 \pm 0.016$
& $\mathbf{0.796} \pm 0.001$
& $\mathbf{0.750} \pm 0.006$
& $\mathbf{0.161} \pm 0.003$
& $\mathbf{0.646} \pm 0.001$ \\

\midrule
\multicolumn{8}{c}{\textit{\textbf{Scenario C}}} \\
\midrule

\textbf{BP}
& U-Net (Base)
& $3.567 \pm 1.734$
& $3.448 \pm 1.800$
& $0.193 \pm 0.003$
& $0.023 \pm 0.009$
& $0.011 \pm 0.009$
& $0.421 \pm 0.004$ \\

& + Fine-tuning
& $\mathbf{0.306} \pm 0.011$
& $\mathbf{-0.129} \pm 0.101$
& $\mathbf{0.812} \pm 0.024$
& $\mathbf{0.767} \pm 0.015$
& $\mathbf{0.350} \pm 0.035$
& $\mathbf{0.676} \pm 0.019$ \\

\cmidrule{2-8}

\textbf{FI}
& U-Net (Base)
& $\mathbf{0.345} \pm 0.015$
& $\mathbf{-0.041} \pm 0.045$
& $0.783 \pm 0.018$
& $\mathbf{0.721} \pm 0.030$
& $0.202 \pm 0.014$
& $\mathbf{0.640} \pm 0.011$ \\

& + Fine-tuning
& $0.379 \pm 0.032$
& $-0.276 \pm 0.064$
& $\mathbf{0.788} \pm 0.018$
& $0.668 \pm 0.045$
& $\mathbf{0.209} \pm 0.010$
& $0.640 \pm 0.012$ \\

\cmidrule{2-8}

\textbf{ROS}
& U-Net (Base)
& $0.306 \pm 0.030$
& $0.120 \pm 0.070$
& $0.616 \pm 0.113$
& $0.530 \pm 0.076$
& $0.151 \pm 0.016$
& $0.565 \pm 0.045$ \\

& + Fine-tuning
& $\mathbf{0.233} \pm 0.003$
& $\mathbf{-0.065} \pm 0.056$
& $\mathbf{0.779} \pm 0.011$
& $\mathbf{0.727} \pm 0.006$
& $\mathbf{0.193} \pm 0.009$
& $\mathbf{0.638} \pm 0.006$ \\

\midrule
\multicolumn{8}{c}{\textit{\textbf{Scenario D}}} \\
\midrule

\textbf{BP}
& U-Net (Base)
& $3.503 \pm 1.732$
& $3.406 \pm 1.791$
& $0.174 \pm 0.007$
& $0.023 \pm 0.009$
& $0.018 \pm 0.010$
& $0.420 \pm 0.005$ \\

& + Fine-tuning
& $\mathbf{0.283} \pm 0.015$
& $\mathbf{-0.122} \pm 0.097$
& $\mathbf{0.813} \pm 0.023$
& $\mathbf{0.772} \pm 0.014$
& $\mathbf{0.326} \pm 0.020$
& $\mathbf{0.673} \pm 0.016$ \\

\cmidrule{2-8}

\textbf{FI}
& U-Net (Base)
& $\mathbf{0.332} \pm 0.014$
& $\mathbf{-0.020} \pm 0.045$
& $\mathbf{0.787} \pm 0.015$
& $\mathbf{0.734} \pm 0.029$
& $0.207 \pm 0.015$
& $\mathbf{0.644} \pm 0.010$ \\

& + Fine-tuning
& $0.364 \pm 0.030$
& $-0.251 \pm 0.064$
& $0.784 \pm 0.018$
& $0.683 \pm 0.044$
& $\mathbf{0.214} \pm 0.011$
& $0.640 \pm 0.012$ \\

\cmidrule{2-8}

\textbf{ROS}
& U-Net (Base)
& $0.296 \pm 0.031$
& $0.128 \pm 0.071$
& $0.624 \pm 0.113$
& $0.530 \pm 0.081$
& $0.161 \pm 0.020$
& $0.570 \pm 0.046$ \\

& + Fine-tuning
& $\mathbf{0.224} \pm 0.002$
& $\mathbf{-0.051} \pm 0.056$
& $\mathbf{0.777} \pm 0.010$
& $\mathbf{0.736} \pm 0.004$
& $\mathbf{0.203} \pm 0.008$
& $\mathbf{0.638} \pm 0.006$ \\

\bottomrule
\end{tabular*}
\caption{Main results comparing the predictive performance of our multi-source U-Net models for \textbf{Burn Probability (BP)}, \textbf{Fire Intensity (FI)}, and \textbf{Rate of Spread (ROS)} on a \textbf{separate regional study}. U-Net model is based on \textbf{Spatial + Weather + Fire Size} input data. Each value is reported as the mean and standard deviation of 3 different-seed runs. Metrics are reported on the test set \textit{\textbf{at region level}}. Best performance for each specified metric is highlighted in bold.}
\label{app:tab:NWT_data_weather_firesize_model}
\end{table*}

\section{How to Use Our AI Framework}
\label{app:framework_use}

Users with an existing BurnP3 or BurnP3+ simulation can use our AI framework to evaluate the provided models or train new models on their own simulation data. The framework requires the following inputs:

\begin{itemize}
    \item \textbf{Spatial rasters:} fuel, elevation, ignition, and fire-zone rasters.
    \item \textbf{Tabular data:} daily weather records, as well as the fire size distribution file used for calibration.
    \item \textbf{Simulation outputs:} average burn probability, mean fire intensity, and mean rate-of-spread rasters aggregated across all seasons.
    \item \textbf{Auxiliary data:} season/cause ignition distributions and season green-up specifications.
\end{itemize}

The project repository provides detailed instructions in the README file for preparing these inputs from raw simulation data, training and evaluating models, and running inference with pretrained or newly trained models.

% \section{Data Ethics} \label{app:data_ethics}

\end{document}